\documentclass[11pt]{article}
\PassOptionsToPackage{table,xcdraw}{xcolor}

\usepackage[preprint]{acl}
\usepackage{times}
\usepackage{latexsym}
\usepackage[T1]{fontenc}
\usepackage[utf8]{inputenc}
\usepackage{microtype}
\usepackage{inconsolata}

\usepackage{graphicx}
\usepackage{float}
\usepackage{subcaption}
\usepackage{wrapfig}
\usepackage{booktabs}
\usepackage{multirow}
\usepackage{makecell}
\usepackage{minibox}
\usepackage{arydshln}
\usepackage{threeparttable}
\usepackage{amsmath}
\usepackage{amssymb}
\usepackage{amsfonts}
\usepackage{mathtools}
\usepackage{amsthm}
\usepackage{bbm}
\usepackage{algorithm}
\usepackage{algorithmic}
\usepackage{numprint}

\usepackage{xspace}
\usepackage{hyphenat}
\usepackage{url}
\usepackage{enumitem}
\usepackage{verbatim}
\usepackage{marvosym}
\usepackage{ifthen}
\usepackage{textcomp}
\usepackage{todonotes}

\usepackage{tikz}
\usetikzlibrary{positioning, fit, arrows.meta}
\usepackage[capitalize,noabbrev]{cleveref}

\newcommand{\chatoDisplayMode}[1]{#1}

\definecolor{MyRed}{rgb}{0.6,0.0,0.0} 
\definecolor{MyBlack}{rgb}{0.1,0.1,0.1} 
\newcommand{\inred}[1]{{\color{MyRed}\sf\textbf{\textsc{#1}}}}
\newcommand{\frameit}[2]{
  \begin{center}
  {\color{MyRed}
  \framebox[.9\columnwidth][l]{
    \begin{minipage}{.85\columnwidth}
    \inred{#1}: {\sf\color{MyBlack}#2}
    \end{minipage}
  }\\
  }
  \end{center}
}

\newcommand{\note}[2][]{\chatoDisplayMode{\def\@tmpsig{#1}\frameit{{\Pointinghand} Note}{#2\ifx \@tmpsig \@empty \else \mbox{ --\em #1}\fi}}}

\newcommand{\abbrevStyle}[1]{#1}

\newcommand{\eg}{\abbrevStyle{e.g.}\xspace}

\newcommand{\Secref}[1]{Sec.~\ref{#1}}

\newcommand{\Tabref}[1]{Table~\ref{#1}}
\newcommand{\Figref}[1]{Fig.~\ref{#1}}

\newcommand{\xhdr}[1]{\vspace{1.7mm}\noindent{{\bf #1.}}}

\newcommand{\textcite}[1]{\citeauthor{#1} \shortcite{#1}}

\newcommand{\RB}{\textsc{ReasonBENCH}\xspace}
\newcommand{\githuburl}{\url{https://github.com/au-clan/ReasonBench}}

\newcommand{\moveup}{\vspace*{-2mm}}
\newcommand{\moveups}{\vspace*{-1mm}}
\newcommand{\ignore}[1]{}
\newcommand{\highlightblue}[1]{\colorbox[HTML]{bae6fb}{\textbf{#1}}}
\newcommand{\highlightorange}[1]{\colorbox[HTML]{fdd55b}{\textbf{#1}}}

\definecolor{VanshVTwoGreen}{HTML}{1B7F3A}
\long\def\vanshv2#1{{\color{black}#1}}

\definecolor{vdBenchSys}{HTML}{2E6CA4}
\definecolor{vdBenchXSys}{HTML}{D4812A}
\definecolor{vdItem}{HTML}{5B9E5F}
\definecolor{vdSysItem}{HTML}{7B5EA7}
\definecolor{vdRepeat}{HTML}{B0B0B0}
\definecolor{vdUsage}{HTML}{C0392B}
\definecolor{vdResidual}{HTML}{D8CFC4}
\newcommand{\swatch}[1]{{\color{#1}\rule[0.05ex]{0.85ex}{0.85ex}}}

\newif\ifshownotes
\shownotesfalse
\renewcommand{\chatoDisplayMode}[1]{\ifshownotes#1\fi}

\newcommand{\io}{\textsc{IO}\xspace}
\renewcommand{\cot}{\textsc{CoT}\xspace}
\newcommand{\cotsc}{\textsc{CoT-SC}\xspace}
\newcommand{\tot}{\textsc{ToT}\xspace}
\newcommand{\got}{\textsc{GoT}\xspace}
\newcommand{\foa}{\textsc{FoA}\xspace}
\newcommand{\react}{\textsc{ReAct}\xspace}
\newcommand{\rap}{\textsc{RAP}\xspace}
\newcommand{\reflexion}{\textsc{Reflexion}\xspace}

\newcommand{\totdfs}{\textsc{ToT-DFS}\xspace}
\newcommand{\totbfs}{\textsc{ToT-BFS}\xspace}
\newcommand{\chainofthought}{Chain-of-Thought\xspace}
\newcommand{\selfconsistency}{Self-Consistency\xspace}

\title{ReasonBENCH: Benchmarking the (In)Stability of LLM Reasoning}

\DeclareSymbolFont{extraup}{U}{zavm}{m}{n}
\DeclareMathSymbol{\epfl}{\mathalpha}{extraup}{83}
\DeclareMathSymbol{\iitd}{\mathalpha}{extraup}{85}
\DeclareMathSymbol{\au}{\mathalpha}{extraup}{86}

\newcommand{\authornote}[1]{\textbf{#1}}

\author{
Nearchos Potatmitis,\thanks{Equal contribution. Correspondence to: \texttt{nearchos.potamitis@cs.au.dk}, \texttt{akhil.arora@cs.au.dk}}~$^{\au}$ Vansh Ramani,\footnotemark[1]~$^{\iitd}$ \\[1ex]
\authornote{Har Ashish Arora},\thanks{Work done at Aarhus University; equal contribution}~$^{\iitd}$
\authornote{Dhairya Kuchhal},\footnotemark[2]~$^{\iitd}$ 
\authornote{Lars Klein},\thanks{Equal supervision.}~$^{\epfl}$
\authornote{Akhil Arora}\footnotemark[3]~$^{\au}$\\[1ex]
$^{\au}$Aarhus University \quad
$^{\iitd}$Indian Institute of Technology Delhi \quad 
$^{\epfl}$EPFL \quad 
}

\begin{document}
\maketitle

\begin{abstract}

Benchmark scores for LLM reasoning systems are reported as single numbers, yet the same model, strategy, and task can produce meaningfully different answers and costs across repeated executions — even under greedy decoding ($T = 0$). This variance is not a statistical nuisance: the highest-performing strategy wins only 77\% of head-to-head runs against its nearest competitor, meaning a single observed score can silently misrank systems. We introduce \RB, a benchmark suite recording 30 independent trials across 10 reasoning strategies, 12 models, and 6 tasks, treating quality and cost as distributions rather than point estimates.
We find that this variance is structured rather than random: a two-component taxonomy—Global Noise, capturing cross-benchmark unevenness, and Run Noise, capturing within-benchmark stochasticity—reveals that strategy architecture predicts stability profiles, while models and strategies shift orthogonal aspects of the distribution. A hierarchical decomposition attributes three-quarters of score variance to benchmark, system, and item structure, with a persistent residual that single-run evaluation silently absorbs. Finally, cost and quality decouple asymmetrically: cheap methods are structurally immune to joint cost-quality failure, while expensive methods remain exposed regardless of their accuracy.
These findings establish instability as an inherent property of reasoning systems and motivate distribution-aware evaluation as standard practice.

\end{abstract}

\section{Introduction}

Progress in large language model (LLM) reasoning is increasingly distilled into single-number benchmark scores. Prompting frameworks~\citep{cot, cot_sc, tot, react, got, foa}, reasoning-trained models~\citep{openai_o1, deepseek_r1}, and tree- or graph-based search procedures~\citep{rap_reasoner, lats} are each reduced to one accuracy figure per task, and that figure is treated as a definitive proxy for capability: it anchors leaderboards, steers model selection, and underwrites deployment decisions.

\begin{figure}[t!]
\centering
\vspace{2mm}
\includegraphics[width=0.95\linewidth]{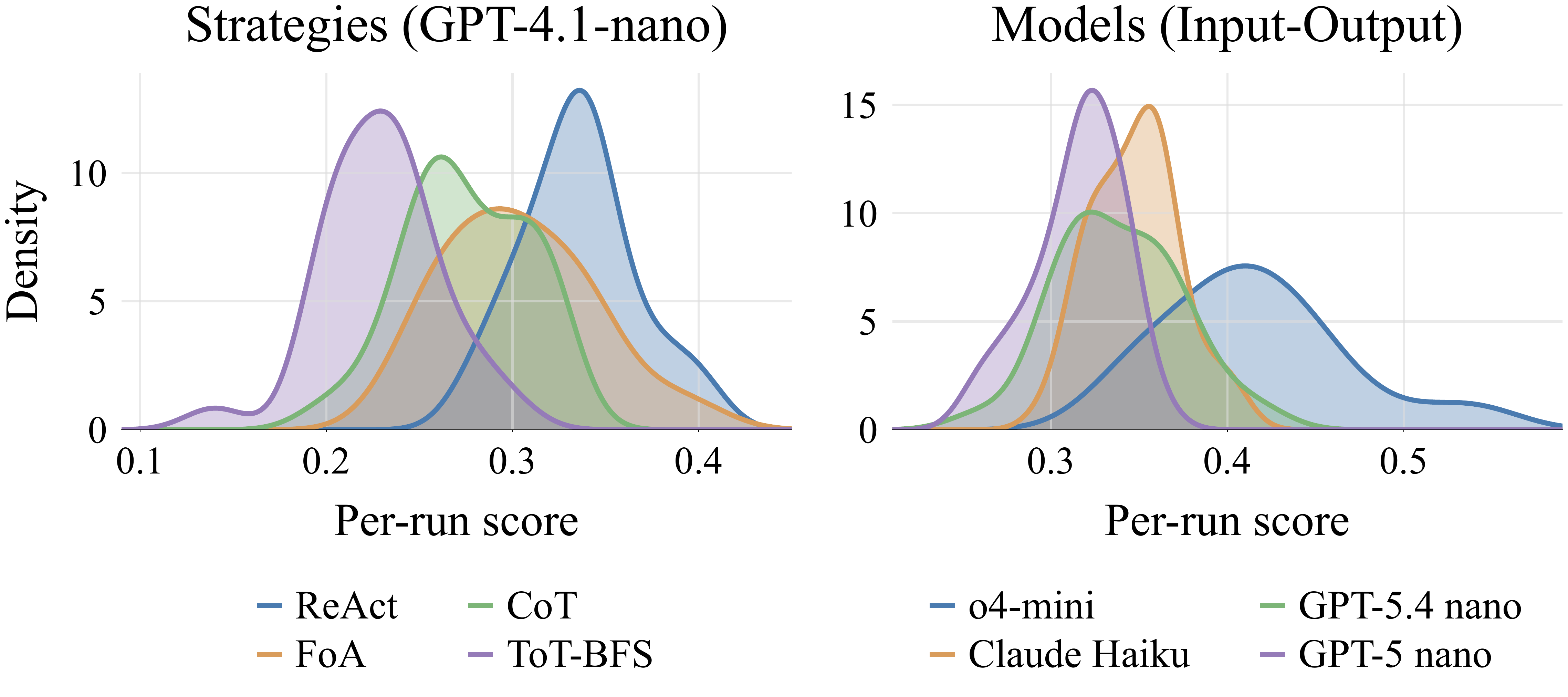}
\moveup
\moveups
\caption{{\textbf{Run-level distributions overlap.} (Left) Four reasoning strategies on a shared GPT-4.1 Nano backbone. (Right) Four reasoning models on shared base strategy, on HotpotQA.}}
\moveup
\moveup
\moveup
\label{fig:distributions}
\end{figure}

\begin{figure}[t!]
\centering
\includegraphics[width=0.99\linewidth]{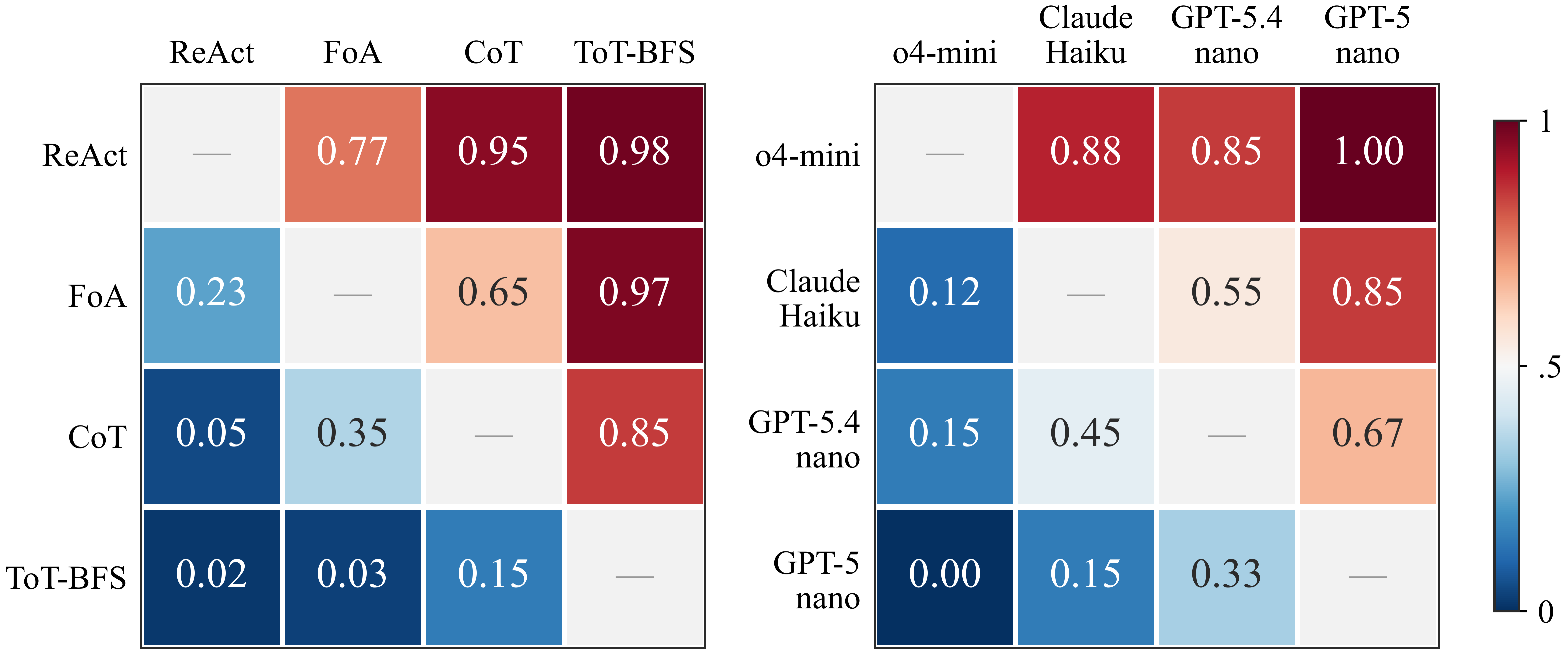}
\moveup
\moveup
\moveup
\caption{{\textbf{Paired per-run win rates.} Cell $(i,j)$ gives P(strategy $i$ outscores strategy $j$) across 30 runs for the HotpotQA benchmark. (Left) shows ReAct vs FoA at 0.77, and (Right) shows o4-mini vs Claude-Haiku 4.5 at 0.88.}}
\moveup
\moveup
\moveup
\moveup
\moveups
\label{fig:winrate_heatmap}
\end{figure}

Yet a benchmark score is only a point estimate of an inherently stochastic process. Because LLM generation is probabilistic and commercial endpoints seldom guarantee deterministic seeds, the same model--strategy--task configuration can trace different reasoning paths, arrive at different answers, and incur different costs across repeated executions. Collapsing these run-level outcomes into a single mean discards the variance that separates a reliably correct system from one that is correct only in expectation. Classical statistical learning has long treated bias and variance as co-equal properties of an estimator~\citep{nn_bias_variance_dilemma, elements_statistical_learning}; LLM reasoning evaluation has largely reported only the former. \Figref{fig:distributions} shows that 30-run distributions overlap substantially for both reasoning strategies and models, so a single observation from each can flip the ranking; \Figref{fig:winrate_heatmap} quantifies this---the top strategy and model beat their nearest competitor on only $77\%$ and $88\%$ of head-to-head runs, far from the near-certainty a mean gap alone would imply. Crucially, this variance does not vanish under greedy decoding ($T{=}0$), ruling out token-level sampling as the dominant explanation.

We instantiate this agenda with \RB, a benchmark suite and open-source library for controlled multi-run evaluation of LLM reasoning. \RB standardizes 10 reasoning strategies under a shared backbone and evaluates 12 contemporary reasoning models zero-shot. Across 6 benchmarks, we execute 30 independent trials per configuration---justified by a two-stage power analysis (see \S\ref{subsec:experimental_setup})---and report quality and cost as empirical distributions. The protocol mirrors operational reality: most provider APIs omit seed control or do not preserve determinism across infrastructure changes, so repeated calls are the variance source practitioners actually encounter. We position \RB against prior reasoning, robustness, and variance-aware evaluation work in \S\ref{sec:related_work}.

\xhdr{Contributions}

\noindent $\bullet$ \textbf{Instability is structured, not noise.} Across 120 model--strategy--task configurations and 30 runs each, variance is substantial, benchmark-dependent, and axis-dependent. A taxonomy along two orthogonal metrics---Global Noise and Run Noise---shows that strategy architecture predicts the stability profile, and a hierarchical decomposition attributes three quarters of variance to benchmark, system, item, and usage structure (\S\ref{sec:results}).

\noindent $\bullet$ \textbf{Models and strategies are non-substitutable axes.} Models move the score ceiling; strategies move execution variance and task-level sensitivity. Cost and quality decouple along both axes, producing a structural asymmetry: cheap methods avoid joint cost--quality failure while expensive ones remain exposed regardless of accuracy (\S\ref{sec:cross_axis}).

\noindent $\bullet$ \textbf{Controlled source elimination with actionable negatives.} Variance persists at $T{=}0$, ruling out token-level sampling; prompt and parser fixes correct bias but not variance; reasoning effort scales cost without reliable quality gains. Only evaluator fidelity and model scale move variance (\S\ref{sec:analysis}).

\noindent $\bullet$ \textbf{Open, reproducible infrastructure.} A modular library with decoupled Method, Agent, Model, State, and Environment abstractions (\S\ref{subsec:library_design}).
\section{Related Work}
\label{sec:related_work}
\xhdr{Reasoning methods and models reported as point estimates}
Prompting and search-based strategies---\chainofthought{}~\citep{cot}, self-consistency~\citep{cot_sc}, \tot{}~\citep{tot}, \react{}~\citep{react}, \got{}~\citep{got}, \rap{}~\citep{rap_reasoner}, \reflexion{}~\citep{reflexion}, LATS~\citep{lats}, \foa{}~\citep{foa}---and reasoning-trained models such as the o1- and DeepSeek-R1-class systems~\citep{openai_o1, deepseek_r1} are compared through a single number per task. Some of these methods do sample multiple chains, votes, or branches internally (notably self-consistency and tree-search family), but that internal multiplicity is not the same as rerunning the released system under identical conditions. The distinction matters: \Figref{fig:distributions} shows that 30 reruns of nominally identical configurations span a wide quality band, so a single observed score can rank a method above or below its neighbours by accident. We do not claim these works are mistaken; their reporting estimates expected performance under one draw, while a practitioner who reruns the same configuration encounters a distribution---and the two estimands can disagree on which system is better, cheaper, or safer to deploy.

\xhdr{Robustness and evaluation methodology}
Two adjacent literatures address evaluation reliability by varying a different variable. Input-side robustness benchmarks perturb prompts, numerical values, or semantic structure to probe sensitivity to the stimulus itself~\citep{matcha, rupbench, investigating_deductive_reasoning, numerical_variations, thinkbench}; methodological critiques argue for error bars, paired significance tests, and multi-prompt aggregation when reporting results~\citep{adding_error_bars, state_of_what_art, towards_reproducible_llm_evaluation}; and general-purpose harnesses such as HELM and the LM Evaluation Harness standardize execution at scale, but treat evaluation as a single-run protocol~\citep{helm, lm_eval_harness}. The first thread asks whether reasoning survives changes in the input; the second asks how to attach uncertainty to a model score. Neither isolates the variance that remains when input, model, strategy, and benchmark are all held fixed---the regime practitioners face on every repeated production call.


\xhdr{Variance-aware benchmarks}
The closest precedents already treat variability as a measurement target, yet each isolates a different facet of it. \citet{sober_look} share our concern that reasoning rankings can shift with decoding, seeds, prompt formatting, and hardware, but their primary question is reinforcement learning versus supervised finetuning on mathematical reasoning; repeated runs---typically ten, ``unless stated otherwise''---serve to denoise that comparison rather than to characterize variance itself. \citet{within_model_between_prompt_variability} partition creative-generation variability into model, prompt, interaction, and within-model components, but on a single domain and across models alone. \citet{llms_stable_reasoning} extend pass@$k$ to G-Pass@$k_\tau$ to capture whether sampled solutions are \emph{consistently} correct on math; further work quantifies variance across benchmarks, seeds, and checkpoints~\citep{quantifying_variance}, or under uncertainty quantification~\citep{benchmarking_llms, cer-eval}. \RB instead makes variance the first-class object of study: we fix the configuration and vary only the execution; treat reasoning \emph{strategies}, not only base models, as comparable systems; make cost a random variable alongside quality; and decompose instability into \emph{within-benchmark} stochasticity (Run Noise) and \emph{across-benchmark} unevenness (Global Noise), leaving \emph{within-problem} variance to future work. The combination exposes effects that accuracy-only stability cannot---joint cost--quality failure, architecture-predicted noise quadrants, and axis-dependent residual variance---and Appendix~\ref{appendix:detailed-related-work} gives a fuller treatment of each thread.

\section{\RB: Framework Design and Experimental Setup}
\label{sec:reasonbench}
\RB separates the reasoning \emph{method}, task \emph{environment}, model interface, and evaluation state, allowing a strategy to be rerun across tasks and models without changes to the experimental harness. The library logs responses, token usage, and costs, and repeated execution provides the variance estimates.

\subsection{Library design}
\label{subsec:library_design}
Each reasoning strategy implements the same interface: a method proposes or searches over intermediate states; an environment validates actions and computes task outcomes; and a model wrapper records provider metadata. Prompt handling, parsing, state transitions, and cost accounting are implemented once per strategy, so the strategy comparison in this paper does not confound the method with the harness. The architecture diagram and per-abstraction details are in Appendix~\ref{appendix:library-design}.

\subsection{Experimental Setup}
\label{subsec:experimental_setup}

\xhdr{Repeated runs}
We repeat each configuration 30 times and report stratified-bootstrap confidence intervals over runs. The choice of $n{=}30$ is justified by a two-stage power analysis (Appendix~\ref{appendix:power-analysis}): $n{\geq}9$ suffices to detect that empirical variance exceeds the threshold at which a 5-point mean difference would be reliable at 90\% power; 30 runs provide $3{\times}$ this minimum, with median 95\% CI width falling below two percentage points by $n{=}30$.

\xhdr{Prompts} We reuse prompts from the original methods when available, and share prompts across strategies whenever the task format permits. This keeps comparisons focused on the reasoning procedure rather than on prompt engineering.

\xhdr{Tasks and data}
We evaluate on six tasks that stress different reasoning regimes: Game of 24~\citep{tot}, HumanEval~\citep{humaneval}, HotpotQA~\citep{hotpotqa}, Humanity's Last Exam~\citep{hle}, SciBench~\citep{scibench}, and Shakespearean Sonnet Writing~\citep{meta_prompting}. The suite intentionally mixes mathematical reasoning, code generation, multi-hop QA, scientific problem-solving, and creative writing with different scoring ranges and output formats, because instability can be hidden by any single benchmark or domain.

\xhdr{Panel selection rationale}
The 10 strategies span five architecture classes (direct, adaptive, structured, planning, and evolutionary), and the 12 models span seven providers across three capability tiers. This is a deliberate spread across architectures, providers, and scales. We do not claim exhaustive coverage of all reasoning methods or all model families; instead, the panel is designed to test whether instability patterns generalize across structural variation. We exclude methods whose released artifacts are incomplete or whose compute requirements make the repeated-run protocol impractical~\citep{tout, recmind, bot, lats}.

\xhdr{Reasoning strategies}
We evaluate 10 representative strategies: \io{}, \cot{}~\citep{cot}, \cotsc{}~\citep{cot_sc}, \react{}~\citep{react}, \reflexion{}~\citep{reflexion}, \totdfs{}~\citep{tot}, \totbfs{}~\citep{tot}, \got{}~\citep{got}, \rap{}~\citep{rap_reasoner}, and \foa{}~\citep{foa}. Each is reimplemented in \RB with shared interfaces for prompt handling, state transitions, parsing, evaluation, and cost accounting.

\xhdr{Reasoning models}
We evaluate contemporary reasoning models spanning OpenAI (GPT-OSS-120B~\citep{gpt_oss}, GPT-4.1 Mini, GPT-4.1 Nano, GPT-5 Mini, GPT-5 Nano, GPT-5.4 Mini, GPT-5.4 Nano, o4-mini), DeepSeek (DeepSeek R1~\citep{deepseek_r1}, DeepSeek-V3.2), Meta (Llama 4 Maverick~\citep{llama4}), Alibaba Cloud (Qwen3-235B~\citep{qwen3}), Google (Gemini-3 Flash~\citep{gemini3}), Anthropic (Claude-Haiku 4.5) and Z.ai (GLM-4.5 Air). Models are evaluated with direct input-output with identical benchmark prompts and harmonized decoding settings where provider APIs permit.

\xhdr{Evaluation metrics}
We report quality and cost with three main statistics. \emph{Average} gives the stratified-bootstrap mean and confidence interval. \emph{Global Noise} measures cross-benchmark unevenness after z-score normalization---it captures how inconsistent a system's performance profile is across tasks, independently of absolute score. \emph{Run Noise} measures within-benchmark z-score variance, isolating repeated-run stochasticity from benchmark difficulty. We also use relative run deviation in targeted diagnostics, but keep the main tables focused on the two noise axes that separate benchmark dependence from repeated-run instability. Formal definitions are in Appendix~\ref{appendix:metrics}.

\begin{table*}[t]
\centering
\caption{\textbf{Quality \& cost variability of reasoning frameworks (GPT-4.1 Nano).} Best in \highlightblue{blue}, worst in \highlightorange{orange}.}
\moveup
\label{tab:strategies_quality_cost_ordered_type}
\small
\begin{minipage}{0.95\textwidth}
\centering
\begin{threeparttable}
    \resizebox{\textwidth}{!}{%
        \begin{tabular}{l ccc ccc}
        \toprule
        \multirow{2}{*}{\textbf{Strategy}} &
        \multicolumn{3}{c}{\textbf{Quality}} &
        \multicolumn{3}{c}{\textbf{Cost}} \\
        \cmidrule(lr){2-4}\cmidrule(lr){5-7}
        &
        \textbf{Average*} &
        \textbf{Noise (Global)} &
        \textbf{Noise (Run)} &
        \textbf{Average*} &
        \textbf{Noise (Global)} &
        \textbf{Noise (Run)} \\
        \midrule

        \io{} &
        0.1341 [0.13, 0.14] &
        1.4016 & 3.9102 &
        \highlightblue{0.0001 [0.00, 0.00]} &
        \highlightorange{1.2333} & \highlightorange{0.9067} \\
        \cot{}~\cite{cot} &
        0.2792 [0.27, 0.29] &
        0.9595 & 2.7089 &
        0.0003 [0.00, 0.00] &
        0.7264 & 0.5993 \\
        \cotsc{}~\cite{cot_sc} &
        0.2265 [0.22, 0.23] &
        1.1607 & \highlightorange{8.1146} &
        0.0014 [0.00, 0.00] &
        \highlightblue{0.5132} & \highlightblue{0.3537} \\

        \hdashline
        \react{}~\cite{react} &
        0.3040 [0.29, 0.31] &
        1.0286 & 4.1126 &
        0.0012 [0.00, 0.00] &
        0.5979 & 0.7647 \\
        \reflexion{}~\cite{reflexion} &
        0.3124 [0.30, 0.32] &
        0.9898 & 4.4020 &
        0.0015 [0.00, 0.00] &
        0.5949 & 0.7739 \\

        \hdashline
        \totdfs{}~\cite{tot} &
        \highlightorange{0.0949 [0.09, 0.10]} &
        \highlightorange{1.5690} & 1.7566 &
        0.0021 [0.00, 0.00] &
        0.7310 & 0.5189 \\
        \totbfs{}~\cite{tot} &
        0.4145 [0.40, 0.42] &
        0.7509 & 1.4098 &
        0.0087 [0.01, 0.01] &
        0.5870 & 0.7059 \\
        \got{}~\cite{got} &
        0.3364 [0.33, 0.35] &
        0.9272 & 1.8614 &
        0.0099 [0.01, 0.01] &
        0.8489 & 0.4444 \\

        \hdashline
        \rap{}~\cite{rap_reasoner} &
        0.3713 [0.36, 0.38] &
        0.9724 & 2.5624 &
        \highlightorange{0.0107 [0.01, 0.01]} &
        0.9114 & 0.8639 \\

        \hdashline
        \foa{}~\cite{foa} &
        \highlightblue{0.4549 [0.45, 0.46]} &
        \highlightblue{0.7020} & \highlightblue{1.2046} &
        0.0065 [0.01, 0.01] &
        0.6310 & 0.6720 \\

        \bottomrule
        \end{tabular}
    } 
    \begin{tablenotes}[flushleft]
        \small \item[*] Reports average value and 95\% confidence intervals in brackets.
    \end{tablenotes}
\end{threeparttable}
\end{minipage}
\moveup
\moveup
\end{table*}

\section{Results}
\label{sec:results}

We evaluate two axes of reasoning-system design. On the \emph{strategy axis}, the underlying model is fixed to GPT-4.1 Nano and the reasoning procedure varies. On the \emph{model axis}, the prompting setup is fixed and the model varies. This separation distinguishes instability due to model capability from instability induced by the reasoning procedure. Reproduction resources are available at \githuburl.

\subsection{Suite-level comparison}
\label{subsec:suite_results}


\xhdr{Quality}
Quality varies sharply on both axes. Suite-level score ranges from $0.23$ (DeepSeek R1) to $0.76$ (Gemini-3 Flash) across models (\Tabref{tab:models_results}), and from $0.09$ (\totdfs{}) to $0.45$ (\foa{}) across strategies on the GPT-4.1 Nano backbone (\Tabref{tab:strategies_quality_cost_ordered_type}). Confidence intervals change the interpretation: Gemini-3 Flash separates cleanly from the rest, whereas \foa{}'s strategy-level lead overlaps with \totbfs{}. Repeated runs therefore distinguish robust rankings from those induced by a single point estimate.


\xhdr{Cost}
Cost spreads exceed quality spreads. Mean cost varies by more than two orders of magnitude across models and by two orders across strategies, and higher cost does not imply higher quality: DeepSeek R1 is the most expensive model yet the weakest, while DeepSeek V3.2 reaches competitive quality at a fraction of the price (\Figref{fig:appx_cross_axis_cost}, Appendix~\ref{appendix:cross_axis}). Strategies couple cost and quality more tightly---search uses additional calls by construction---but the relationship is still non-monotonic: \totdfs{} costs more than direct prompting yet performs worse.


\xhdr{Noise}
Instability is not a scalar. Among strategies, \cotsc{} has the highest quality Run Noise and \totdfs{} the highest quality Global Noise, while \io{} leads on both cost dimensions. Among models, DeepSeek R1 has the highest quality Global Noise and Qwen3-235B the highest quality Run Noise; GPT-5.4-mini and GLM-4.5 Air lead respectively on cost Global and Run Noise. The two metrics flag different systems by design.

\begin{table*}[t]
\centering
\vspace{2mm}
\caption{\textbf{Quality and cost variability of \textit{reasoning} models across benchmarks.} Best in \highlightblue{blue}, worst in \highlightorange{orange}.}
\moveup
\label{tab:models_results}
\begin{threeparttable}
    \resizebox{0.95\textwidth}{!}{%
        \begin{tabular}{l ccc ccc}
        \toprule
        \textbf{Reasoning Model} &
        \multicolumn{3}{c}{\textbf{Quality}} &
        \multicolumn{3}{c}{\textbf{Cost}} \\
        \cmidrule(lr){2-4}\cmidrule(lr){5-7}
        & \textbf{Average*} & \textbf{Noise (Global)} & \textbf{Noise (Run)} &
        \textbf{Average*} & \textbf{Noise (Global)} & \textbf{Noise (Run)} \\
        \midrule

        Gemini 3 Flash &
        \highlightblue{0.7615 [0.75, 0.77]}\tnote{$\dagger$} &
        \highlightblue{0.3631} & \highlightblue{0.5809} &
        0.0195 [0.02, 0.02] &
        \highlightblue{0.3127} & \highlightblue{0.6354} \\

        DeepSeek V3.2 &
        0.6479 [0.64, 0.66] &
        0.4607 & 0.8525 &
        0.0015 [0.00, 0.00] &
        0.7989 & 0.8406 \\

        GPT-5.4 mini &
        0.6220 [0.61, 0.63] &
        0.4242 & 0.8895 &
        0.0036 [0.00, 0.00] &
        \highlightorange{0.8163} & 1.2042 \\

        GPT-5 mini &
        0.5660 [0.56, 0.58] &
        0.5937 & 1.0892 &
        0.0034 [0.00, 0.00] &
        0.6351 & 0.8731 \\

        o4-mini &
        0.5564 [0.55, 0.57] &
        0.5023 & 1.0611 &
        0.0084 [0.01, 0.01] &
        0.6481 & 0.8534 \\

        GPT-5 nano &
        0.5085 [0.50, 0.52] &
        0.7817 & 1.7309 &
        0.0012 [0.00, 0.00] &
        0.5206 & 0.6684 \\

        GPT-OSS 120B &
        0.5012 [0.49, 0.51] &
        0.5735 & 1.2225 &
        0.0006 [0.00, 0.00] &
        0.4390 & 0.8269 \\

        GPT-5.4 nano &
        0.4528 [0.44, 0.46] &
        0.4833 & 1.1781 &
        0.0007 [0.00, 0.00] &
        0.7057 & 1.0182 \\

        GLM-4.5 Air &
        0.4007 [0.39, 0.41] &
        0.8248 & 1.6563 &
        0.0047 [0.00, 0.00] &
        0.6895 & \highlightorange{1.2460} \\

        Claude Haiku 4.5 &
        0.3777 [0.37, 0.39] &
        0.8065 & 1.4370 &
        0.0022 [0.00, 0.00] &
        0.5624 & 0.7297 \\

        Qwen3 235B Thinking &
        0.3525 [0.34, 0.36] &
        0.9818 & \highlightorange{3.9442} &
        0.0130 [0.01, 0.01] &
        0.8031 & 0.8382 \\

        DeepSeek R1 &
        \highlightorange{0.2326 [0.22, 0.24]} &
        \highlightorange{1.2217} & 3.3963 &
        \highlightorange{0.0281 [0.03, 0.03]} &
        0.7430 & 0.7528 \\
        \bottomrule
        \end{tabular}
    } 
    
    \begin{tablenotes}[flushleft]
        \small 
        \item[$\dagger$] Indicates statistical significance ($p<0.05$) between the best and the second-best scores among reasoning models.
        \item[*] Reports average value and 95\% confidence intervals in brackets. Run Noise and Global Noise are defined in \ref{appendix:metrics}.
        \moveup
        \moveup
    \end{tablenotes}
\end{threeparttable}
\end{table*}

\subsection{Noise taxonomy}
\label{subsec:noise_orthogonality}

\begin{figure}[t]
\centering
\moveup
\moveup
\includegraphics[width=\columnwidth]{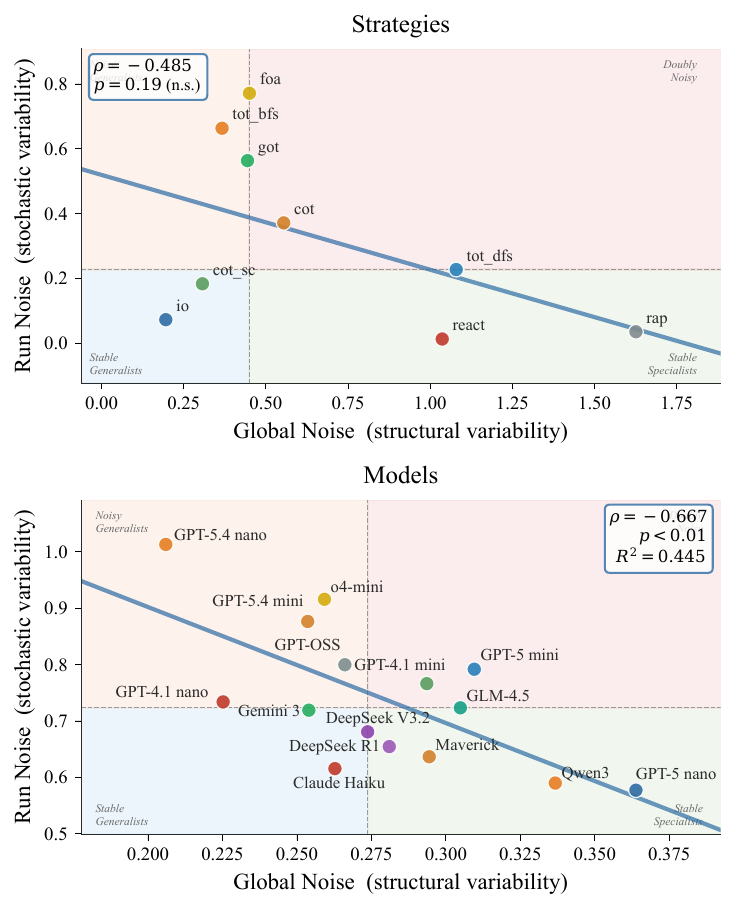}
\moveup
\moveup
\moveup
\moveup
\caption{\textbf{Noise taxonomy.} Each system is plotted in (Global Noise, Run Noise) space; dashed lines mark the median on each axis, defining four quadrants. Blue lines show linear regression with fit quality (inset boxes). \emph{Top:} Strategies show weak negative correlation. \emph{Bottom:} Models exhibit significant anti-correlation between structural and stochastic instability. Per quadrant discussion is in Appendix \ref{appendix:metrics}.}
\moveup
\moveup
\moveup
\moveup
\label{fig:noise_taxonomy}
\end{figure}

Tables~\ref{tab:strategies_quality_cost_ordered_type} and~\ref{tab:models_results} show that mean score, benchmark dependence, and within-benchmark stochasticity are separable axes. Plotting every system in the (Global Noise, Run Noise) plane reveals a relative taxonomy with four quadrants---\emph{Stable Generalists} (low on both), \emph{Noisy Generalists} (low Global, high Run), \emph{Stable Specialists} (high Global, low Run), and \emph{Doubly Noisy} (high on both); \Figref{fig:noise_taxonomy} shows the result. Models concentrate on the anti-diagonal: structural unevenness and stochastic instability anti-correlate across the panel. Strategies populate all four quadrants, and the quadrant a strategy occupies is predicted by its architecture---deterministic execution suppresses Run Noise, search-based execution amplifies it, and benchmark-specific successes or failures drive Global Noise. The taxonomy is therefore diagnostic: it tells a practitioner whether instability is structural (uneven across tasks) or stochastic (unreliable within a task), with different remedies for each.

\subsection{Variance decomposition}
\label{subsec:variance_decomposition}

\begin{figure}[t]
\centering
\moveup
\moveup
\includegraphics[width=\linewidth]{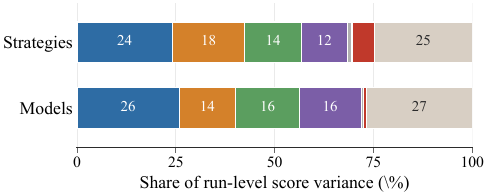}
\moveup
\moveup
\moveup
\caption{{\textbf{Run-level variance decomposition.} Hierarchical decomposition of item-level scores along the strategy and model axes; numbers are percentage points of total variance.
\swatch{vdBenchSys}\,benchmark + system,\;
\swatch{vdBenchXSys}\,benchmark\,$\times$\,system,\;
\swatch{vdItem}\,item (within benchmark),\;
\swatch{vdSysItem}\,system\,$\times$\,item,\;
\swatch{vdRepeat}\,repeat,\;
\swatch{vdUsage}\,token usage,\;
\swatch{vdResidual}\,residual.}}
\label{fig:variance_decomposition}
\end{figure}

To quantify what the taxonomy summarizes, we expand every run into item-level observations and fit a hierarchical decomposition in a fixed order: benchmark and system main effects, their interaction, item difficulty, system--item interactions, repeat effects within each benchmark--system cell, and token usage. The order is conservative---task, item, and method structure receive credit before usage---so a token covariate cannot poach variance that belongs to benchmark composition.

\Figref{fig:variance_decomposition} shows that roughly three quarters of variance is structured on both axes ($75.2\%$ for strategies, $73.3\%$ for models). What drives it is not rerun noise: benchmark--system interactions, item difficulty, and system--item interactions dominate, while pure repeat effects account for only $1.2\%$ (strategies) and $0.6\%$ (models). Token usage explains $5.7\%$ on the strategy axis but $0.8\%$ on the model axis---search procedures vary their footprint, models do not. The residual quarter is the load-bearing finding: after task, item, system, repeat, and usage are controlled, $24.8\%$ (strategies) and $26.7\%$ (models) of variance remains unexplained by any design factor we can observe.


\section{Sources of Instability}
\label{sec:analysis}

The previous section establishes that run-to-run variance is substantial and structured. We now ask which factors influence it. We test four candidates through controlled interventions holding the task and evaluation protocol fixed (\Tabref{tab:source_elimination}): decoding temperature, prompt and parser specification, evaluator quality, and reasoning effort. The key distinction throughout is between interventions that move the \emph{mean} (correcting bias) and those that move the \emph{variance} (reducing stochasticity); these are independent, and the former does not imply the latter. Appendix~\ref{appendix:model_scale} contains a brief case study on the effect of model scale on accuracy and variance.

\begin{table*}[t]
\centering
\caption{\textbf{Elimination study of instability sources.} Each row applies a controlled intervention and records whether mean quality and run-to-run variance change.}
\moveup
\moveups
\label{tab:source_elimination}
\footnotesize
\setlength{\tabcolsep}{4pt}
\begin{tabular}{@{}p{0.22\textwidth}p{0.28\textwidth}p{0.13\textwidth}p{0.10\textwidth}p{0.20\textwidth}@{}}
\toprule
\textbf{Source} & \textbf{Intervention} & \textbf{Mean} & \textbf{Variance} & \textbf{Verdict} \\
\midrule
Decoding randomness & $T \in \{0, 0.35, 0.7\}$ (Fig.~\ref{fig:temperature_grid}, \S\ref{appx:decoding}) & marginal & no change & not sufficient \\
Prompts \& parsers & refined templates (Tab.~\ref{tab:frameworks_quality_diff_appx}, \S\ref{appx-prompts}) & improves & no change & corrects bias only \\
Evaluation function & ground-truth on Game24 (Fig.~\ref{fig:evaluator_quality}) & improves & decreases & variance source \\
Reasoning effort & minimal $\to$ high (Fig.~\ref{fig:effort_panel_main}, \S\ref{appx-thinking}) & non-monotonic & no change & does not help \\
\bottomrule
\end{tabular}
\end{table*}

\begin{figure}[t]
\centering
\includegraphics[width=\linewidth]{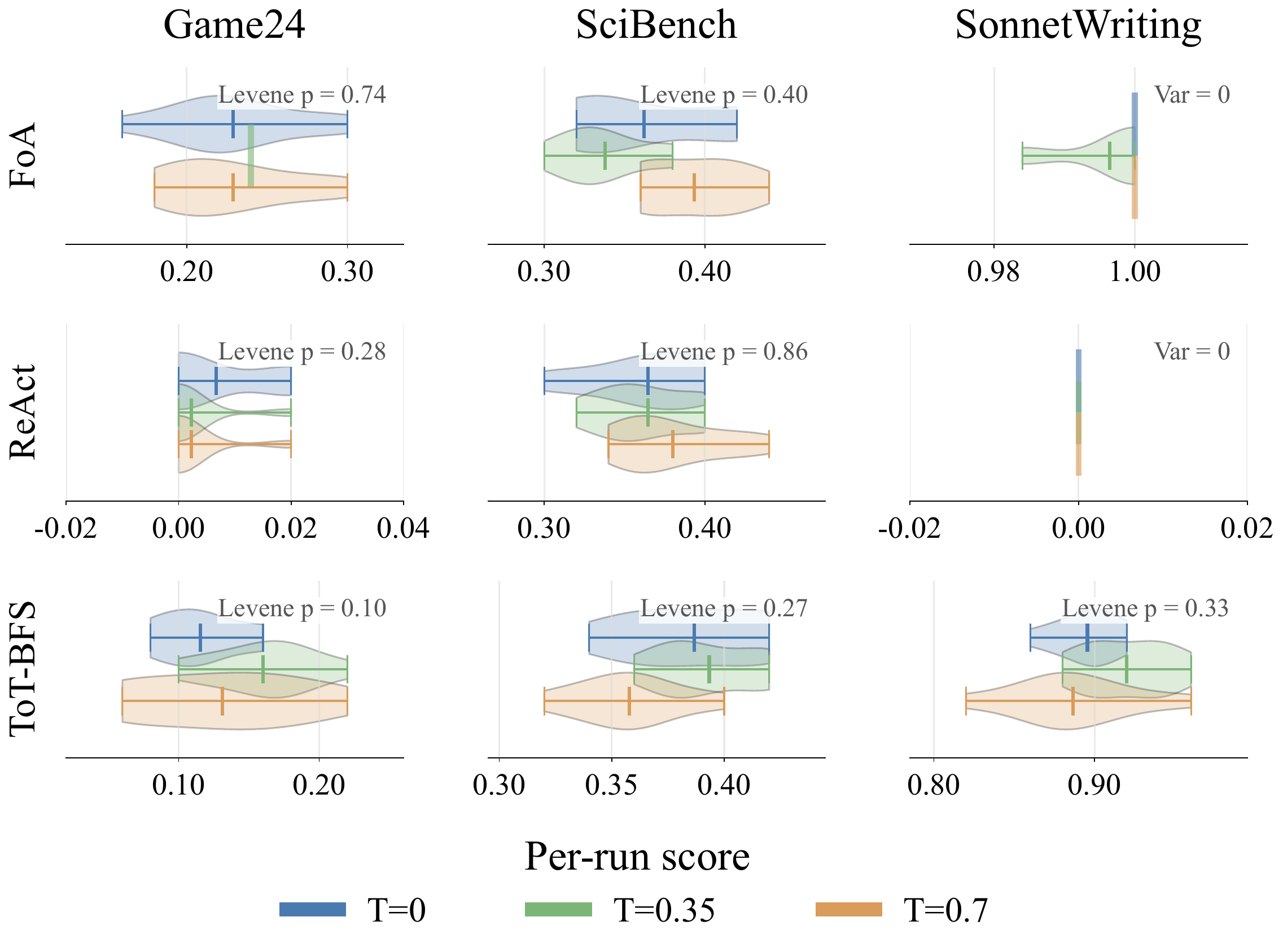}
\caption{\textbf{Variance persists at $T{=}0$.} Run-level score distributions for three strategies (rows) and three benchmarks (columns) at $T \in \{0, 0.35, 0.7\}$. Per-cell details in Appendix~\ref{appx:decoding}. $p$-value from Levene’s test for equality of variances ($T=0$ vs. $T=0.7$) is given.}
\label{fig:temperature_grid}
\end{figure}

The most common explanation for run-to-run variance is sampling at the decoding step. We sweep temperature over $T \in \{0, 0.35, 0.7\}$ for a representative subset of strategy--benchmark cells. If token-level sampling were the dominant source, greedy decoding ($T{=}0$) should collapse the run distribution. \Figref{fig:temperature_grid} shows the opposite: the inter-quartile range at $T{=}0$ is comparable to---and in several cells wider than---the range at $T{=}0.7$. This is a load-bearing negative result: temperature does not explain the instability we measure, and the variance persists even under fully deterministic token selection. Token-level sampling is thus not a sufficient explanation; upstream sources such as API-level batching, infrastructure non-determinism, or MoE routing, together with intrinsic strategy randomness, are candidates left for future work.

\xhdr{Prompt and parser specification}
Prompts may underspecify answer formats, and parsers may fail on harmless output variation. We refine both across all strategies and rerun on GPT-4.1 Nano. Refinement yields significant mean quality improvements across all strategies, with the largest gains in direct methods, yet confidence-interval \emph{widths} are essentially unchanged. However, prompt and parser fixes correct systematic error (bias) but not repeated-run variance (stochasticity).

\xhdr{Evaluation function}
Search-based strategies rely on an evaluator to score intermediate states. On Game of 24, where ground-truth evaluation is computable, we replace the heuristic evaluator with an oracle while holding the search procedure fixed. The oracle improves mean quality and \emph{reduces variance}: evaluator noise propagates through the search process, amplifying instability beyond the final-answer level. Evaluator calibration is, therefore, a direct stability lever for any strategy that consumes an evaluation signal.

\xhdr{Reasoning effort}
We test whether additional test-time compute reduces instability by sweeping the effort setting on three contemporary reasoning models over 10 independent runs each. Prior work~\citep{more_is_less} reports an inverted-U relation between reasoning length and accuracy; \Figref{fig:effort_panel_main} sharpens this into an operational claim: once repeated-run variability is accounted for, quality differences between effort settings are not statistically reliable, while output tokens rise monotonically. Spending more on test-time reasoning reliably increases cost, but buys neither higher quality nor lower variance---a useful negative result for the test-time compute scaling agenda.
\section{Models and Strategies Are Different Axes}
\label{sec:cross_axis}

\subsection{What changes when we vary each axis}
\label{subsec:cross_axis_analysis}
The two axes in \RB answer different scientific questions. Varying the model changes the base capability available to the system. Varying the strategy changes the procedure that uses that capability: whether the system samples one answer, searches a tree, reflects on failures, aggregates candidates, or calls an evaluator. Treating these as substitutes is misleading because they change different parts of the distribution.


\xhdr{Models move the ceiling; strategies move the procedure}
The two axes shift different parts of the distribution. Suite-level score spans $0.628$ across models against $0.360$ across strategies ($1.7\times$ wider on the model axis), while Global Noise spans $1.414$ vs $0.917$ ($1.5\times$) and Run Deviation $0.587$ vs $0.315$ ($1.9\times$ wider on the strategy axis). Model choice therefore decides whether a task is within reach at all; strategy choice decides how often the same underlying model reaches a correct answer through a particular reasoning process. The failure modes differ accordingly: a zero-score model cell typically reflects a capability ceiling, whereas a zero-score strategy cell under the same backbone reflects search failure, parser mismatch, or a brittle method--task interaction.

\begin{figure}[t]
\centering
\includegraphics[width=0.8\linewidth]{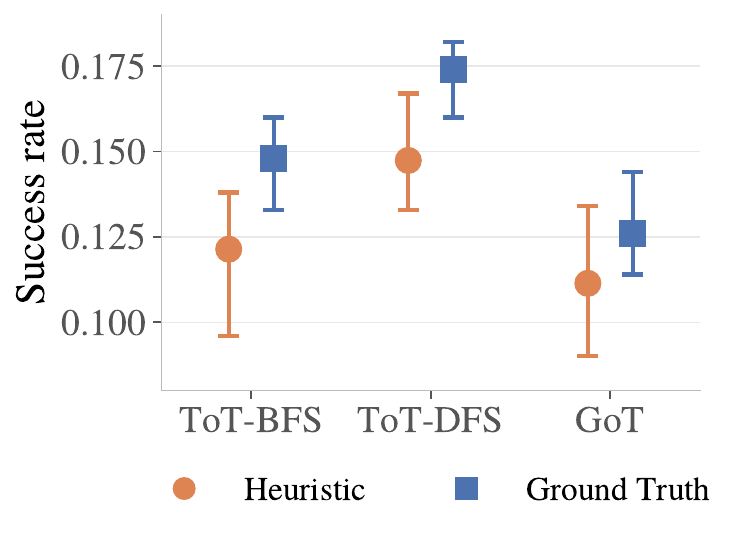}
\caption{\textbf{Oracle evaluation raises mean quality and tightens run distributions across all three search-based strategies on Game of 24.} The oracle raises mean performance and tightens run-to-run distributions across all three strategies.}
\label{fig:evaluator_quality}
\end{figure}
\begin{figure}[t]
\centering
\includegraphics[width=\linewidth]{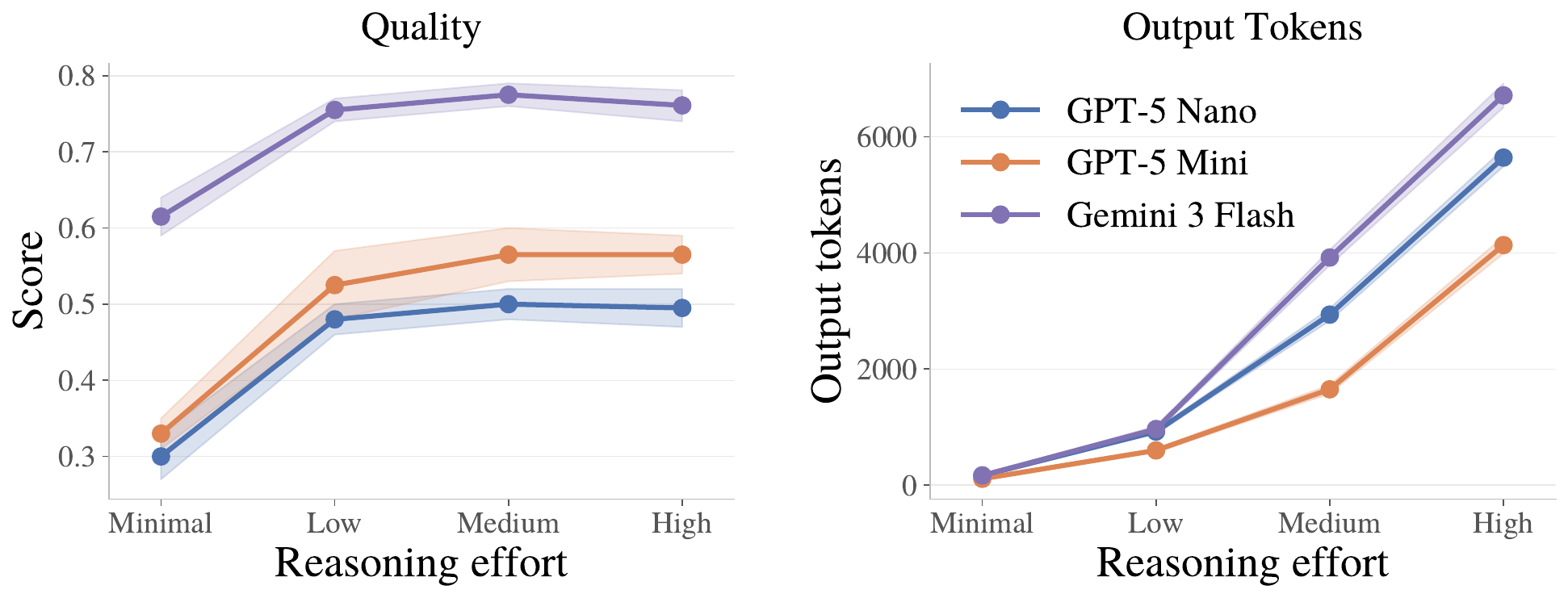}
\caption{{\textbf{Higher cost doesn't necessarily mean better quality.} Score changes across effort settings remain within run-to-run uncertainty (left), while output tokens rise monotonically (right).}}
\label{fig:effort_panel_main}
\end{figure}

\xhdr{Cost-quality decoupling}
Both axes break the assumption that higher cost reliably buys better quality. On the model axis, the most expensive model in the panel scores near the bottom in quality, while several mid-cost models reach competitive performance. On the strategy axis, search-based methods that cost substantially more than direct prompting can still score below it. Cost and quality are decoupled on both axes. This decoupling has a distributional consequence that is visible only under repeated runs. We define a run as a \textit{joint failure} if it simultaneously delivers below-median quality and above-median cost. Cheap methods are structurally immune to joint failure since they never exceed the cost threshold. Expensive methods remain exposed, regardless of their average quality: high joint failure rates are more common among the costliest models and strategies. This asymmetry is invisible from any single run and emerges only from the run-level distribution, and has a direct deployment implication: a nominally better but expensive method may dominate on mean quality while being substantially worse on joint failure risk.



\subsection{Implications for evaluation practice}
\label{subsec:cost_quality}

These findings change what an evaluation table should support. A suite designed to compare models should be validated for capability discrimination; the same suite, used to compare strategies, must also reveal procedural variance, parser failures, and cost instability. A single aggregate score cannot serve both purposes. \RB therefore treats repeated-run distributions as the object of evaluation: they expose whether a method is consistently better, better only in expectation, or better only under an unacceptable cost profile.

\section{Discussion}
\label{sec:discussion}

\subsection{Implications}
\label{sec:implications}

\xhdr{Stability is part of capability}
Reasoning benchmarks should not treat variance as formatting around a mean. A system that solves a task only on some runs has a different capability profile from one that solves it consistently, even if their averages match. Reporting confidence intervals, run deviation, benchmark dependence, and cost variability makes this distinction operational.

\xhdr{Models and strategies are non-substitutable, orthogonal axes}
The two axes answer different scientific questions. Stronger models raise the performance ceiling; structured strategies change the procedure that uses that ceiling and, with it, the shape of the run distribution. The axes are largely orthogonal in our data: a model that pushes the mean upward need not sit in the same stability quadrant as a strategy that does the same, and a benchmark suite that cleanly ranks one axis need not cleanly rank the other. Evaluation design should be validated for the axis being compared, and conclusions transported between axes only with care.

\xhdr{Cost is a reliability variable}
Cost is not merely a secondary efficiency metric. In repeated reasoning, expensive runs can be failed runs, and increasing effort can raise cost without reliably improving quality. The cheap-immune / expensive-exposed asymmetry in joint failure (\S\ref{sec:cross_axis}) means that cost-quality risk is structurally asymmetric. Evaluations that report quality alone miss this failure mode.

\xhdr{Negative results as positive guidance}
Three of our four source-elimination interventions yield negative results: temperature, prompts/parsers, and reasoning effort do not reduce variance. These are valuable findings. They tell practitioners that the most convenient decoding-time fixes are ineffective, and that stability should be optimized in the design of reasoning systems---through better evaluators and stronger models---rather than patched at inference time.

\xhdr{Toward co-developed reasoning systems}
The orthogonality of the two axes suggests a research program rather than a choice between them. The field currently concentrates effort on reasoning-trained \emph{models}, treating prompting and search strategies as a legacy concern. Our data argues the opposite: because models move ceilings while strategies move procedure and variance, the two should be co-developed---through meta-systems that match a problem to an appropriate model--strategy pair, route between direct answering and search under cost constraints, or adapt scaffolding to task type. Recent directions such as structured-trajectory post-training, recursive and looped-attention models, and hierarchical recursive architectures move in compatible spirit, but explicit co-development of reasoning models with the procedures that consume them remains, to our knowledge, largely open.

\subsection{What a practitioner should do differently}
\label{sec:recommendations}






Our findings imply four concrete changes to evaluation and deployment practice. First, \textbf{report distributions, not point estimates}: at minimum stratified-bootstrap CIs, ideally Global and Run Noise alongside the mean. Second, \textbf{validate the benchmark suite for the axis being compared}---a suite that discriminates models may not discriminate strategies. Third, \textbf{budget for joint cost--quality risk} by estimating the joint failure rate before deploying an expensive method; cheap alternatives often dominate on that metric. Fourth, \textbf{invest in evaluator calibration over decoding tuning}: for search-based strategies, a better intermediate evaluator reduces both bias and variance, lowering temperature does not.

\subsection{Future Work}
Three directions follow from the framework: a \emph{within-problem} variance decomposition, the natural finer-grained complement to the within- and across-benchmark levels we report; stability-aware routing that selects a model--strategy pair under explicit cost and reliability constraints, turning the orthogonality of the two axes into a design lever; and distribution-aware leaderboards as a community standard, so that single-run scores no longer set the default lens on reasoning progress. \RB is designed to support all three.

\section*{Limitations}
\label{sec:limitations}
Our study evaluates six tasks, ten strategies, and twelve model configurations; additional domains or tool-using settings may expose different variance structure. We test five candidate sources of instability but do not claim a mechanistic account of where every divergent reasoning path begins. Training data, instruction tuning, RLHF, provider-side infrastructure, and hardware non-determinism remain outside our intervention set. Finally, 30 runs per configuration support mean and moderate-tail analysis, but not extreme-quantile guarantees for high-stakes deployment. Humanity's Last Exam uses o3-mini as a judge for evaluation. Because the judge is itself a stochastic process, some of the Run Noise we attribute to reasoning systems may be partially attributable to evaluator variance.

\bibliography{references}

\appendix
\vanshv2{
\section*{Appendix}
\addcontentsline{toc}{section}{Appendix}

\noindent The appendix is organized into three parts. \textbf{Appendix~\ref{appendix:detailed-related-work}} extends the related-work discussion across point-estimate reasoning literature, input-side robustness, statistical evaluation practice, and the closest variance-aware precedents. \textbf{Appendix~\ref{appendix:additional-experimental-details}} consolidates the experimental setup---task descriptions, reasoning-strategy specifications, implementation, library design, sample-size and power analysis, and the formal definitions of Run Deviation, Global Noise, and Run Noise. \textbf{Appendix~\ref{appendix:additional-results}} collects the analyses deferred from the main paper: per-benchmark breakdowns on the model and strategy axes, decoding-configuration sweeps, the prompt-and-parser refinement study, the causal evaluator intervention, the full reasoning-effort sweep, run-level diagnosis traces, joint cost--quality tables, and cross-axis details.
}

\vanshv2{
\section{Detailed Related Work}
\label{appendix:detailed-related-work}

\xhdr{Reasoning methods and models as point estimates}
The reasoning procedures that anchor modern leaderboards---\chainofthought{}~\citep{cot}, self-consistency~\citep{cot_sc}, \tot{}~\citep{tot}, \react{}~\citep{react}, \got{}~\citep{got}, \rap{}~\citep{rap_reasoner}, \reflexion{}~\citep{reflexion}, LATS~\citep{lats}, \foa{}~\citep{foa}---and reasoning-trained models such as the o1- and DeepSeek-R1-class systems~\citep{openai_o1, deepseek_r1} are typically compared through benchmark-level mean scores or pass rates. Some of these methods do sample multiple candidates internally (votes, branches, reflections), but the unit of report remains a single number per task. This format answers a narrower question than the one a deployer asks: under an identical model, strategy, prompt, and task, what distribution of quality and cost should I expect across repeated executions? A point estimate captures expectation under one draw; a repeated-run benchmark captures the distribution itself. Rankings, reliability, and cost can shift when the latter is measured, and our experiments show that they often do.

\xhdr{Input-side robustness}
A first adjacent line of work studies brittleness by changing the input. RUPBench probes reasoning under prompt perturbations~\citep{rupbench}; MATCHA targets misalignment between reasoning and final answer~\citep{matcha}; deductive- and numerical-variation studies measure sensitivity to logical and surface form~\citep{investigating_deductive_reasoning, numerical_variations}; stress-test and out-of-distribution suites probe robustness through adversarial inputs~\citep{ar_checker, thinkbench}; and surveys document the same fragility at higher level~\citep{llm_for_mathematical_reasoning, beyond_accuracy}. These works ask whether reasoning survives a change in the stimulus. \RB asks a complementary question: whether the reported score survives \emph{no} change at all. A system can be robust to paraphrase yet inconsistent across repeated identical calls, or vice versa, so the two regimes are not interchangeable.

\xhdr{Statistical evaluation practice}
A second line argues that LLM evaluation should report uncertainty, not only averages. Miller imports standard statistical practice into LLM evaluation~\citep{adding_error_bars}; Mizrahi et al. show that multi-prompt aggregation can flip conclusions~\citep{state_of_what_art}; Blackwell et al. quantify benchmark-score uncertainty through repeated trials and observe that even seeded $T{=}0$ is not always deterministic~\citep{towards_reproducible_llm_evaluation}; MixEval reduces single-benchmark dependence~\citep{mixeval}; and general-purpose harnesses such as HELM and the LM Evaluation Harness standardize execution at scale~\citep{helm, lm_eval_harness}. We adopt the spirit of this agenda, but the unit of analysis is different. Rather than attach error bars to a model score on a task, we measure the rerun distribution of a reasoning \emph{system}---model, strategy, environment, parser, evaluator, and cost accounting taken together---and decompose its instability.

\xhdr{Variance-aware benchmarks and our delta}
The closest prior work treats variability as a measurement target. Hochlehnert et al.~\citep{sober_look} show that mathematical-reasoning conclusions are sensitive to decoding, seeds, formatting, and even hardware. Haase et al.~\citep{within_model_between_prompt_variability} partition creative-generation variance into model, prompt, interaction, and within-model components. Liu et al.~\citep{llms_stable_reasoning} extend pass@$k$ to G-Pass@$k_\tau$, capturing whether multiple sampled solutions are \emph{consistently} correct. Madaan et al.~\citep{quantifying_variance} quantify variance across benchmarks, checkpoints, and seeds, while uncertainty-quantification benchmarks~\citep{benchmarking_llms, cer-eval} and domain-specific reliability suites~\citep{autoeval, medomni} push in adjacent directions. Relative to these, \RB makes four choices that together define its contribution: it fixes the configuration and only varies the execution; it treats reasoning strategies, not only base models, as comparable systems; it makes cost a first-class random variable alongside quality; and it decomposes instability into Global Noise (cross-benchmark) and Run Noise (within-benchmark). The first turns variance from a property of evaluation design into a property of the system itself; the rest let us see effects---joint cost--quality failure, architecture-predicted noise quadrants, axis-dependent residual variance---that accuracy-only stability cannot.

}

\section{Additional Experimental Details}
\label{appendix:additional-experimental-details}

\subsection{Detailed Task Descriptions}
\label{appendix:tasks}
\subsubsection{Game of 24}
The Game of 24 is a math puzzle where players are given four numbers and must use each of them exactly once, along with the basic arithmetic operations ($+$, $-$, $\times$, $\div$), to form an expression that evaluates to 24.

Our benchmark includes 1,362 such puzzles collected from 4nums.com, organized in ascending order of difficulty. Each puzzle provides four input numbers, and the goal is to generate a valid equation that results in 24. Following the approach of ToT~\cite{tot}, we designate puzzles numbered 901 to 1000 as our test set.

\subsubsection{SciBench}
SciBench~\cite{scibench} is a scientific reasoning benchmark designed to evaluate college-level problem-solving abilities across subjects such as mathematics, physics, and chemistry. Each task presents an open-ended problem that requires multi-step reasoning, domain-specific knowledge, and advanced computations, including calculus and differential equations. Problems are drawn from widely used textbooks and university exams. 

Following the approach of ReST-MCTS~\cite{restmcts}, we sampled 109 problems spanning different subjects to form the test set. Quality is measured using an \emph{accuracy} metric, defined as the proportion of problems correctly solved according to the official solutions (exact matching).

\subsubsection{HumanEval}
HumanEval~\cite{humaneval} is a code generation benchmark where participants are given natural language docstrings and must generate Python functions that correctly implement the described behavior. Each problem includes a hidden test suite used to verify functional correctness. 

Following the setup from \reflexion{}~\cite{reflexion}, the benchmark consists of 100 programming tasks in the test set. We evaluate performance using the \emph{pass@1} metric, which measures the proportion of problems solved correctly on the first attempt.

\subsubsection{HotpotQA}
HotpotQA~\cite{hotpotqa} is a large-scale question answering benchmark that tests an agent's ability to perform multi-hop reasoning across multiple documents. Multi-step approaches, such as ToT, are permitted to interact with an API that enables document retrieval and targeted information lookup.

Following prior work~\cite{lats, reflexion}, we evaluate on a set of 100 randomly selected questions. The quality of a response is judged based on \emph{exact match} (EM) with the oracle answer. 

\subsubsection{Shakespearean Sonnet Writing}
Shakespearean Sonnet Writing~\cite{meta_prompting} is a creative generation task where the goal is to compose a 14-line sonnet adhering to the classic rhyme scheme ``ABAB CDCD EFEF GG''. Each sonnet must include three provided words verbatim.

Following Suzgun et al.~\cite{meta_prompting}, we randomly sampled 50 datapoints to form the test set. Quality is measured using an \emph{accuracy} metric, which reflects the proportion of sonnets that both satisfy the rhyme scheme and include all three required words exactly as given.

\subsubsection{Humanity's Last Exam}
Humanity's Last Exam~\cite{hle} is a challenging, multidisciplinary benchmark designed to probe the upper limits of general reasoning and knowledge in large language models. The benchmark consists of carefully curated questions spanning mathematics, natural sciences, humanities, and abstract reasoning, with an emphasis on problems that require deep understanding, precise reasoning, and resistance to shallow pattern matching.

Each task is presented as a standalone question with a single correct answer, typically requiring multi-step logical inference, symbolic manipulation, or synthesis of domain knowledge. The benchmark is designed to be closed-book and does not permit external tool use.

Following the official evaluation protocol, we evaluate models on a 50-sample subset and report performance using an \emph{accuracy} metric, defined as the proportion of questions answered exactly correctly. Correct answers are determined using the original authors' LLM-as-judge system with the recommended prompts and models (GPT-o3 Mini).

\subsection{Detailed Descriptions of Reasoning Strategies}
\label{appendix:reasoning_strategies}

\begin{table}[tbp]
\centering
\caption{\textbf{Reasoning strategies and their type classification.} Companion to \Tabref{tab:strategies_quality_cost_ordered_type}.}
\label{tab:strategy_types_appx}
\small
\begin{tabular}{ll}
\toprule
\textbf{Strategy} & \textbf{Type} \\
\midrule
\io{}        & Direct \\
\cot{}       & Direct \\
\cotsc{}    & Direct \\
\midrule
\react{}     & Adaptive \\
\reflexion{} & Adaptive \\
\midrule
\totdfs{}   & Structured \\
\totbfs{}   & Structured \\
\got{}       & Structured \\
\midrule
\rap{}       & Planning \\
\midrule
\foa{}       & Evolutionary \\
\bottomrule
\end{tabular}
\end{table}

\subsubsection{\io{} (Input-Output)}
A direct prompting strategy where the model maps an input to an output in a single step, without generating or exposing intermediate reasoning. \io{} relies entirely on the model's internal representations and is typically used as a baseline for comparison with more explicit reasoning methods.

\subsubsection{\chainofthought{} (\cot{})}
Encourages the model to generate an explicit sequence of intermediate reasoning steps before producing a final answer. By verbalizing its reasoning process, \cot{} is expected to improve performance on multi-step and compositional reasoning tasks \cite{cot}.

\subsubsection{\chainofthought{} with \selfconsistency{} (\cotsc{})}
Extends \chainofthought{} by sampling multiple independent reasoning chains and aggregating their final answers via a consistency-based voting mechanism. This approach mitigates errors from individual reasoning paths and improves robustness and accuracy \cite{cot_sc}.

\subsubsection{\reflexion{}}
A reasoning framework that enables models to iteratively reflect on and critique their own outputs using feedback from prior attempts or environment interactions. \reflexion{} leverages self-evaluation to generate corrective insights, which are incorporated into subsequent reasoning steps to improve task performance over time \cite{reflexion}.

\subsubsection{Tree of Thoughts (\tot{})}
Decomposes the problem into multiple chains of thoughts, organized in a tree structure. Thought evaluation and search traversal algorithms are utilized to solve the problem \cite{tot}.

\subsubsection{Fleet of Agents (\foa{})}
Decomposes the problem into multiple chains of thoughts. Employs a genetic-type particle filtering approach to navigate through dynamic tree searches to solve the problem \cite{foa}.

\subsubsection{Graph of Thoughts (\got{})}
Allows the organization of thoughts in a graph structure~\cite{got}. It introduces arbitrary graph-based thought transformations such as thought aggregation and thought refinement.

\subsubsection{\react{}}
A reasoning method that interleaves reasoning (thought generation) and acting (taking environment-interacting actions) to solve problems interactively. Each action's output informs subsequent reasoning, enabling adaptive and dynamic problem-solving \cite{react}.

\subsubsection{Reasoning via Planning (\rap{})}
\rap{} is a reasoning framework that equips large language models with an internal world model and employs Monte Carlo Tree Search (MCTS) for strategic exploration of reasoning paths. \rap{} repurposes the LLM to simulate future states and evaluate potential actions, enabling deliberate planning and improved problem-solving performance \cite{rap_reasoner}.


\begin{table}[tbp]
\centering
\caption{\textbf{Reasoning models and their providers.} Companion to \Tabref{tab:models_results}.}
\label{tab:model_providers_appx}
\small
\begin{tabular}{ll}
\toprule
\textbf{Reasoning Model} & \textbf{Provider} \\
\midrule
DeepSeek R1          & DeepSeek \\
DeepSeek V3.2        & DeepSeek \\
Qwen3 235B Thinking  & Alibaba \\
GPT-OSS 120B         & OpenAI \\
GPT-5 mini           & OpenAI \\
GPT-5 nano           & OpenAI \\
GPT-5.4 mini         & OpenAI \\
GPT-5.4 nano         & OpenAI \\
o4-mini              & OpenAI \\
Claude Haiku 4.5     & Anthropic \\
Gemini 3 Flash       & Google \\
GLM-4.5 Air          & Z.ai \\
\bottomrule
\end{tabular}
\end{table}

\subsection{Implementation Details}
\label{appendix:implementation-details}

\subsubsection{Platforms}
\label{appendix:platforms}
GPT models were accessed through the \href{https://platform.openai.com/docs/overview}{OpenAI API}, Google models through the \href{https://ai.google.dev/gemini-api/docs}{Gemini API}, and the remaining models through the \href{https://docs.together.ai/docs/introduction}{TogetherAI API}.

\subsubsection{Model checkpoints and prices}
\label{appendix:checkpoints-and-prices}
To compute the costs of our experiments, we used the model prices published by OpenAI, Gemini, and Together AI for each provider. The specific model snapshots we used, along with their respective prices, are presented in \Tabref{tab:snapshot_prices}.
\begin{table}[ht]
\centering
\caption{Cost of each model that we have used, at this project's time of execution.}
\label{tab:snapshot_prices}
\scriptsize
\setlength{\tabcolsep}{3pt}
\resizebox{\linewidth}{!}{%
\begin{tabular}{l l r r}
\toprule
Model & Provider used & Input (\$/1M) & Output (\$/1M) \\
\midrule
DeepSeek R1          & OpenRouter API    & 3.00 & 7.00 \\
DeepSeek V3.2        & OpenRouter API    & 0.25 & 0.38 \\
Qwen3 235B Thinking  & Together API      & 0.65 & 3.00 \\
GPT-OSS 120B         & Together API      & 0.15 & 0.60 \\
GPT-5 mini           & OpenAI API        & 0.25 & 2.00 \\
GPT-5 nano           & OpenAI API        & 0.05 & 0.40 \\
GPT-5.4 mini         & OpenAI API        & 0.75 & 4.50 \\
GPT-5.4 nano         & OpenAI API        & 0.20 & 1.25 \\
o4-mini              & OpenAI API        & 1.10 & 4.40 \\
Claude Haiku 4.5     & Anthropic API     & 1.00 & 5.00 \\
Gemini 3 Flash       & Google Gemini API & 0.50 & 3.00 \\
GLM-4.5 Air          & OpenRouter API    & 0.13 & 0.85 \\
\bottomrule
\end{tabular}
}
\end{table}

\subsubsection{Model configurations}
\label{appendix:model-configurations}
Generation parameters were specified when making calls to the models used throughout this study. These parameters were inherited from the original implementations where the respective prompts were introduced. For newer models, we adjusted only the maximum allowed completion tokens as needed to ensure compatibility and successful completion of responses.
\begin{table}[ht]
\centering
\caption{Generation parameters specified when making requests to a base LLM.}
\label{tab:model_config}
\scriptsize
\setlength{\tabcolsep}{3pt}
\resizebox{\linewidth}{!}{%
\begin{tabular}{l r r r l}
\toprule
 & \textbf{max\_tokens} & \textbf{temperature} & \textbf{top\_p} & \textbf{stop} \\
\midrule
\textbf{Game of 24}     & 200 & 0.7 & 1 & Null \\
\textbf{SciBench}       & 300 & 0.7 & 1 & Null \\
\textbf{Humanity's Last Exam}            & 300 & 0.7 & 1 & Null \\
\textbf{HumanEval}      & 200 & 0.7 & 1 & Null \\
\textbf{HotpotQA}       & 300 & 0.7 & 1 & Null \\
\textbf{Sonnet Writing} & 800 & 1.0 & 1 & Null \\
\bottomrule
\end{tabular}
}
\end{table}

\subsubsection{On random seeds}
\label{appendix:random-seeds}
We do not fix random seeds across runs. The vast majority of providers in our evaluation (including OpenAI, Google, Anthropic, and DeepSeek) do not expose a seed parameter through their public APIs, and even where a nominal seed is accepted it is not guaranteed to produce deterministic outputs across infrastructure changes. The variability we measure therefore reflects the stochasticity that practitioners actually observe when calling these models, rather than an artificial seed sweep, and is precisely the reason we rely on multi-run statistics rather than fixed-seed reproducibility.

\subsubsection{Prompts}
\label{appendix:prompts}
Due to the large number of methods and tasks presented in this paper, including all corresponding prompts would be impractical in the appendix. We therefore provide a comprehensive collection of all prompts used in our experiments, both original and improved, on our GitHub repository: \url{https://github.com/au-clan/ReasonBench}.

\subsection{Library Design Details}
\label{appendix:library-design}

This section expands on the five core abstractions of \RB summarized in \S\ref{subsec:library_design}. We integrate the abstractions with CacheSaver~\citep{cachesaver}, which caches provider calls across repeated experiments and keeps the multi-run protocol affordable at scale.

\xhdr{Method Abstraction}
The method abstraction specifies the overall logic of a reasoning strategy independently of the underlying model or task. A method integrates agents, which construct prompts and parse responses; the environment, which maintains and updates the task state; and the model, which produces candidate outputs. Each method exposes a standard interface for solving tasks by generating and updating sequences of states, and a benchmarking routine that runs multiple problem instances in parallel. This makes methods interchangeable and extensible: once the interface is implemented, a new reasoning algorithm can be evaluated consistently across models, tasks, and metrics within the benchmarking pipeline.

\xhdr{Environment Abstraction}
The environment abstraction formalizes the task-specific dynamics of reasoning. It governs how a state evolves in response to an action, how to determine whether an action is valid, when a trajectory has reached a terminal condition, and how to evaluate the final outcome. By encapsulating these rules, the environment decouples domain logic from reasoning algorithms, allowing the same method to be applied consistently across tasks while ensuring that actions and evaluations remain faithful to each benchmark.

\xhdr{Agent Abstraction}
The agent abstraction defines the interface between methods, models, and states. Agents specify how prompts are constructed from the current state, how queries are issued to the model, and how responses are parsed into actions that update the environment. This unified interface makes it possible to express a wide spectrum of reasoning strategies: from simple input--output prompting to multi-step reasoning, search procedures, candidate aggregation, and self-evaluation. By isolating prompt construction and response handling, \RB supports diverse reasoning paradigms without altering the abstractions for methods, environments, or models.

\xhdr{State Abstraction}
The state abstraction captures the intermediate configuration of a reasoning process. It provides a standardized way to represent progress on a task and to handle states with controlled randomness. Methods interact only with states, while environments define how actions modify them and how final outcomes are assessed. This separation ensures that reasoning trajectories can be reproduced, compared, and analyzed independently of the underlying task domain.

\xhdr{Model Abstraction}
The model abstraction provides a uniform interface for interacting with language models, supporting both single and batched queries across diverse providers. Built on top of asynchronous execution (via \textit{asyncio}) and integrated with response caching through \textit{CacheSaver}, it is both extensible and accountable: new models can be added without modifying the framework, and every interaction logs latency, token usage, and generation metadata. This combination enables deterministic reproducibility across repeated experiments while distinguishing between newly generated, reused, and deduplicated outputs.

\subsection{Sample Size and Power Analysis}
\label{appendix:power-analysis}

To justify the sample size of 30 runs per configuration, we conduct a two-stage statistical power analysis. We assume that benchmark scores are approximately normally distributed, since they are scaled sums of independent and identically distributed Bernoulli variables (per-problem successes); this assumption is used \emph{only} for sample-size planning, while all confidence intervals reported in the paper are computed by stratified bootstrap resampling and are therefore non-parametric.

\xhdr{Stage 1: Critical variance threshold (two-sample $t$-test)} We first identify the maximum run-to-run variance compatible with reliably detecting a meaningful difference between two configurations. Following typical claims in the literature, we target a true mean difference of $\Delta\mu = 0.05$ (a 5\% effect size). Using a two-sided, equal-variance two-sample $t$-test with $n$ runs per group, the test statistic follows a non-central $t$-distribution with $2n-2$ degrees of freedom and non-centrality parameter
\[
\lambda \;=\; \frac{\Delta\mu}{\sigma\sqrt{2/n}}.
\]
Solving for the critical maximum standard deviation $\sigma_0$ such that statistical power $1-\beta$ remains at $0.90$ for significance level $\alpha = 0.05$ yields the threshold $\sigma_0 \approx 0.021$. Above this level, the test loses the sensitivity required to detect a $5\%$ effect.

\xhdr{Stage 2: Required sample size (one-sided $\chi^2$ variance test)} We then ask how many samples are needed to demonstrate that empirical variance significantly \emph{exceeds} this threshold in practice. Assuming a true underlying standard deviation of $\sigma = 0.04$ (the level we expect to encounter for moderately noisy strategy--benchmark cells), we use a one-sided $\chi^2$ test with hypotheses
\[
H_0: \sigma^2 \leq \sigma_0^2 \quad \text{vs.} \quad H_1: \sigma^2 > \sigma_0^2,
\]
where the test statistic $(n-1)s^2/\sigma_0^2$ follows a $\chi^2_{n-1}$ distribution under $H_0$. Targeting power $1-\beta' = 0.80$ at $\alpha = 0.05$, the analysis indicates that $n \geq 9$ samples are sufficient to reject $H_0$.

\xhdr{Choice of $n=30$} The threshold $n \geq 9$ derived above is the minimum required to detect that empirical variance exceeds the safe level. Collecting 30 runs per configuration provides over $3\times$ this minimum, leaving substantial headroom for variance-aware analyses (bootstrap confidence intervals, run-deviation, and z-score noise) that are more demanding than the planning-stage $t$-test.

\xhdr{Empirical confirmation} The choice of $\sigma = 0.04$ as the underlying standard deviation in Stage 2 proved appropriate: empirical checks confirmed that 10 distinct model--benchmark pairs and 9 strategy--benchmark combinations natively exhibit standard deviations meeting or exceeding this level, with 22 and 24 combinations respectively statistically rejecting the safer $\sigma_0 = 0.021$ threshold.

\subsubsection{Confidence-Interval Convergence}
\label{appendix:ci-convergence}
\begin{figure}[tbp]
    \centering
    \includegraphics[width=0.99\linewidth]{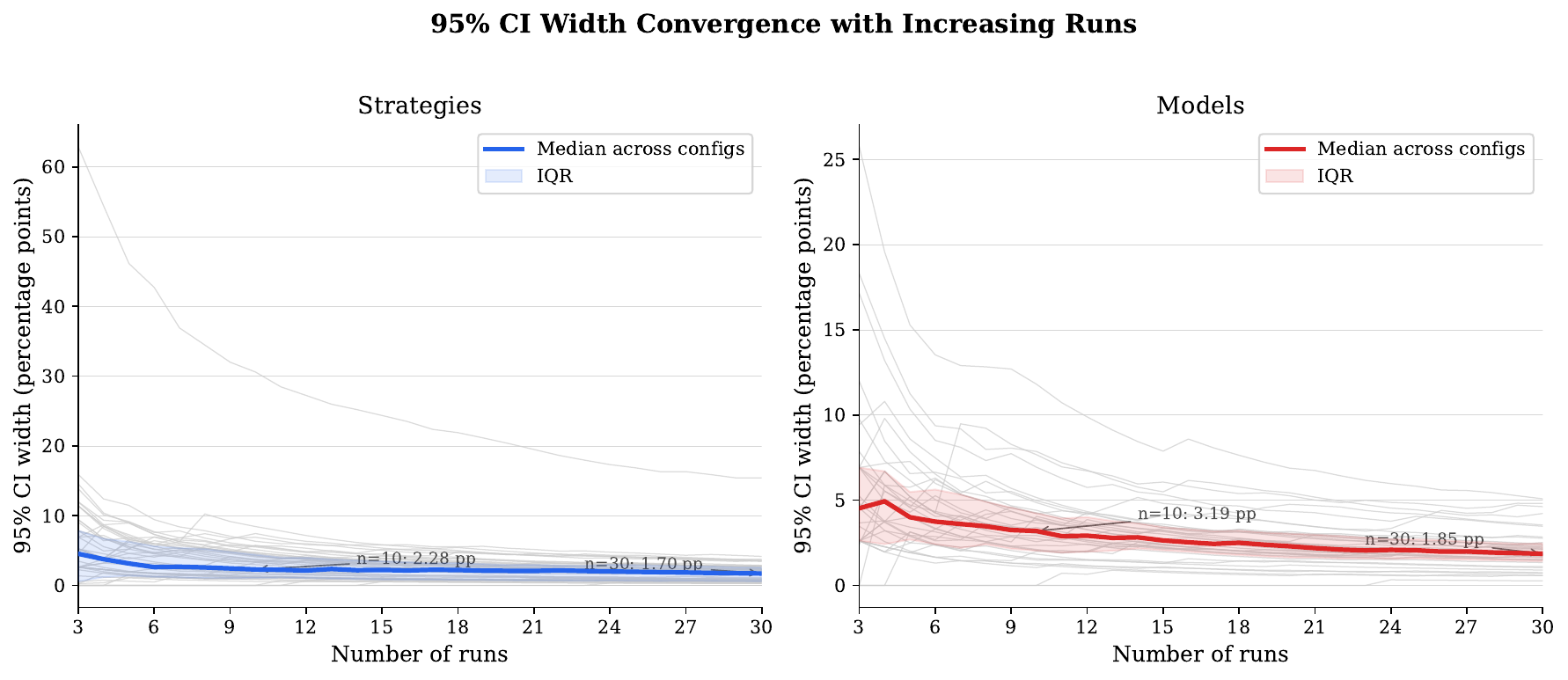}
    \caption{\textbf{Convergence of bootstrap $95\%$ confidence intervals as the number of runs $n$ grows from $3$ to $30$.} Solid line: median CI width across configurations; shaded band: inter-quartile range; light gray: individual per-configuration traces. \textbf{Left:} strategy comparison under GPT-4.1 Nano. \textbf{Right:} model comparison. In both panels, the median CI width tightens sharply over the first $\sim 10$ runs and continues to shrink more slowly to below $2$ pp by $n=30$.}
    \label{fig:ci_convergence}
\end{figure}

As a complementary check, we measure the width of the bootstrap $95\%$ confidence intervals as the number of runs $n$ grows from $3$ to $30$, separately for the strategy comparison (Tab.~\ref{tab:strategies_quality_cost_ordered_type}) and the model comparison (Tab.~\ref{tab:models_results}). \Figref{fig:ci_convergence} reports both the median CI width across configurations (solid line) and the inter-quartile range (shaded band), with individual per-configuration traces overlaid in light gray.

The median CI width drops sharply over the first roughly ten runs and then continues to tighten more gradually: from $\sim 5$ pp at $n=3$ to $2.28$ pp at $n=10$ and $1.70$ pp at $n=30$ for strategies, and from $\sim 5$ pp at $n=3$ to $3.19$ pp at $n=10$ and $1.85$ pp at $n=30$ for models. Two observations are worth highlighting. First, even the few configurations whose CIs start very wide at small $n$ (visible as outlier traces above $30$ pp on the strategy panel) collapse onto the rest of the distribution well before $n=30$. Second, by $n=30$ the median CI width is below $2$ pp in both panels, comfortably narrower than the $5$ pp effect size targeted in our power analysis (\S\ref{appendix:power-analysis}), so the chosen sample size leaves headroom for resolving practically meaningful differences without requiring further runs.

\subsection{Evaluation Metrics: Definitions and Interpretation}
\label{appendix:metrics}

This section expands the metric definitions sketched in \S\ref{subsec:experimental_setup}. We consider a model $m$ evaluated on benchmark $b$ over $n$ independent runs, where $x_{i,b}$ denotes the score (or cost) of run $i$ on benchmark $b$, and the per-cell mean is
\[
\mu_{m,b} \;=\; \mathbb{E}\!\left[X \,\middle|\, \text{model}=m,\ \text{benchmark}=b\right].
\]

\xhdr{Average and confidence intervals} We report the per-cell mean $\mu_{m,b}$ together with $95\%$ confidence intervals computed by stratified bootstrap resampling over runs, where each benchmark is treated as a stratum. The intervals reflect the spread expected under independent reruns of the same benchmark suite, rather than parametric uncertainty around a single point estimate. Two configurations with similar means can still differ in stability, so comparisons should consider the full distribution rather than means alone.

\xhdr{Run Deviation (relative error)} For each run $i$ on cell $(m,b)$ we compute the normalized absolute deviation from the cell mean,
\[
\text{RelError}_i \;=\; \frac{|x_i - \mu_{m,b}|}{\mu_{m,b} + \varepsilon},
\]
where $\varepsilon$ is a small constant that prevents division by zero on benchmarks with mean near zero. Run Deviation is reported as the bootstrapped average of $\text{RelError}_i$ across runs and benchmarks, with $95\%$ CIs computed by the same resampling scheme.

\xhdr{Z-score normalization} To compare scores across benchmarks of different difficulty, we standardize each sample with respect to the within-benchmark distribution:
\[
z_{i,b} \;=\; \frac{x_{i,b} - \mu_b}{\sigma_b},
\]
where $\mu_b$ and $\sigma_b$ are the mean and standard deviation computed over \emph{all} samples within benchmark $b$, pooled across models. Z-scores are intermediate quantities that underlie the two noise metrics that follow.

\xhdr{Global Noise} For each model $m$, Global Noise is the variance of its z-scores pooled across all $(i,b)$ samples:
\begin{align}
\mu_m
&= \frac{1}{N_m}
   \sum_{(i',b')\in m} z_{i',b'} ,
\\[0.5em]
\text{GlobalNoise}_m
&= \mathrm{Var}\!\left(
    \{z_{i,b} \mid \text{model}=m\}
\right) \\
&= \frac{1}{N_m - 1}
   \sum_{(i,b)\in m}
   \left(
      z_{i,b} - \mu_m
   \right)^2 .
\end{align}
where $N_m$ is the total number of $(i,b)$ samples for model $m$. Global Noise is dominated by \emph{between-benchmark} dispersion: a model that performs uniformly across benchmarks has low Global Noise, while one that excels on some benchmarks and fails on others has high Global Noise, regardless of run-to-run consistency.

\xhdr{Run Noise} Run Noise is the \emph{within-benchmark} average of z-score variance,
\begin{align}
\mu_{m,b}
&= \frac{1}{N_{m,b}}
   \sum_{i : (i,b)\in m} z_{i,b},
\\[0.5em]
\text{RunNoise}_m
&= \frac{1}{|B|}
   \sum_{b \in B}
   \frac{1}{N_{m,b}-1}
   \sum_{i : (i,b)\in m}
   \left(
      z_{i,b} - \mu_{m,b}
   \right)^2 .
\end{align}
where $B$ is the set of benchmarks, and $N_{m,b}$ is the number of samples for model $m$ on benchmark $b$. By averaging variance \emph{within} each benchmark, Run Noise isolates the stochastic component of variability under repeated trials and discards the structural component captured by Global Noise.

\xhdr{Score-band amplification of Rel Error} Tab.~\ref{tab:score_band_amplification} expands the discussion of denominator amplification in \S\ref{subsec:noise_orthogonality}. Grouping every $(\text{system}, \text{benchmark})$ cell by score band, the ratio of Rel Error to Run Noise is roughly two orders of magnitude larger for cells with score below $0.05$ than for cells in the medium range, confirming that the disagreement between the two metrics is concentrated entirely at the boundary of the score range.

\begin{table}[tbp]
\centering
\caption{\textbf{The Rel Error / Run Noise gap localizes to near-zero cells.} For each (system, benchmark) cell we compute the ratio Rel Error / Run Noise and average within score bands. Near-zero cells inflate the ratio by roughly two orders of magnitude; medium and high scores leave it close to unity. Strategies additionally include 6 zero-score cells where both metrics are exactly $0$ and which are excluded from the table.}
\label{tab:score_band_amplification}
\small
\begin{tabular}{lcccc}
\toprule
& \multicolumn{2}{c}{\textbf{Models}} & \multicolumn{2}{c}{\textbf{Strategies}} \\
\cmidrule(lr){2-3}\cmidrule(lr){4-5}
\textbf{Score band} & \# cells & RE / RN & \# cells & RE / RN \\
\midrule
$< 0.05$ (near-zero)  & 7  & ${\approx}\,95\times$  & 7  & ${\approx}\,112\times$ \\
$[0.05,\, 0.2)$ (low) & 13 & ${\approx}\,2.6\times$ & 15 & ${\approx}\,1.3\times$ \\
$[0.2,\, 0.5)$ (medium) & 15 & ${\approx}\,2.0\times$ & 15 & ${\approx}\,0.9\times$ \\
$[0.5,\, 0.8)$ (high) & 8  & ${\approx}\,1.7\times$ & 7  & ${\approx}\,2.0\times$ \\
$\geq 0.8$ (very high) & 16 & ${\approx}\,2.6\times$ & 4  & ${\approx}\,3.3\times$ \\
\bottomrule
\end{tabular}
\end{table}

\xhdr{Validation correlations} Global Noise tracks z-score variability (Spearman $\rho = 0.988$, Pearson $r = 0.984$ for models) rather than raw score spread ($\rho = -0.18$, $r = -0.13$); the same pattern holds for strategies ($\rho = 0.983$ for z-score std, $\rho = 0.50$ for raw std). Run Noise tracks the per-cell run-level standard deviation at $\rho = 0.90$ for strategies and $\rho = 0.42$ for models; the rank correlation is weaker for models because Run Noise is dominated by a few high-variance cells (especially HLE).

\xhdr{Metric-polarizing vs metric-neutral benchmarks} The benchmarks in our suite split into two regimes. \emph{Metric-neutral}: HotpotQA and SciBench produce consistent rankings under all three noise metrics, because most systems score in the $0.3$--$0.85$ range where the denominator effect is mild. \emph{Metric-polarizing}: Game24 and SonnetWriting host many near-zero cells and therefore drive the denominator-effect ranking flips; HLE additionally introduces heterogeneous Run Noise (range $31.8\times$ across models). A benchmark suite that mixes both regimes, as ours does, cannot be summarized by any single noise metric.

\xhdr{Noise-taxonomy interpretation} The combined taxonomy in \Figref{fig:noise_taxonomy} separates structural unevenness (Global Noise) from stochastic rerun instability (Run Noise). Whereas the model plane is close to one-dimensional, the strategy plane populates all four quadrants, and the quadrant a strategy lands in is predicted by its architecture.


\section{Additional Results}
\label{appendix:additional-results}

\Tabref{tab:source_elimination} summarizes the source-elimination study referenced in \Secref{sec:analysis}. Each row corresponds to one of the four candidates examined in the main paper (decoding temperature, prompt and parser specification, evaluator quality, reasoning effort) and reports the controlled intervention used, its effect on mean quality and run-to-run variance, and the resulting verdict. The subsections below expand on each row in turn, followed by a model-scale case study, per-benchmark breakdowns, run-level diagnoses, joint cost--quality tables, and cross-axis details.

\subsection{Reasoning Strategies}
\label{appendix:additional-results-strategies}
The tables below complement \Tabref{tab:strategies_quality_cost_ordered_type} with cost and output-token breakdowns across the ten strategies, each evaluated on \texttt{GPT-4.1-Nano} over 30 runs per cell.

\begin{table*}[t]
\centering
\caption{\textbf{Cost variability across reasoning frameworks under GPT-4.1 Nano.} Best in \highlightblue{blue}, worst in \highlightorange{orange}. Strategy types are listed in \Tabref{tab:strategy_types_appx}.}
\scriptsize

\begin{threeparttable}
    \resizebox{0.99\linewidth}{!}{
    \begin{tabular}{l|cccc}
    \toprule
    \textbf{Strategy} &
    \textbf{Average*} &
    \textbf{Run Deviation (\%)*} &
    \textbf{Noise (Global)} &
    \textbf{Noise (Run)} \\
    \midrule

    \io{} &
    \highlightblue{0.54 [0.47, 0.61]} &
    3.37 [1.24, 6.28] &
    0.0109 & \highlightblue{0.0000} \\

    \cot{}~\cite{cot} &
    1.31 [1.22, 1.42] &
    4.53 [2.19, 7.32] &
    \highlightblue{0.0106} & 0.0000 \\

    \cotsc{}~\cite{cot_sc} &
    6.83 [6.75, 6.90] &
    \highlightblue{0.8 [0.37, 1.27]} &
    0.0636 & 0.0000 \\

    \hdashline

    \react{}~\cite{react} &
    6.24 [5.59, 7.07] &
    \highlightorange{10.08 [4.68, 19.04]} &
    0.0242 & 0.0011 \\

    \hdashline

    \totdfs{}~\cite{tot} &
    10.32 [9.45, 11.14] &
    3.99 [1.76, 6.72] &
    0.3547 & 0.0049 \\

    \totbfs{}~\cite{tot} &
    43.68 [41.90, 45.20] &
    4.61 [1.82, 9.13] &
    1.0032 & \highlightorange{0.0057} \\

    \got{}~\cite{got} &
    49.46 [48.13, 50.95] &
    1.94 [0.86, 3.20] &
    1.2931 & 0.0033 \\

    \hdashline

    \rap{}~\cite{rap_reasoner} &
    \highlightorange{53.23 [52.28, 54.35]} &
    6.22 [1.07, 13.20] &
    \highlightorange{1.6912} & 0.0020 \\

    \hdashline

    \foa{}~\cite{foa} &
    32.28 [31.52, 33.05] &
    3.5 [1.22, 7.05] &
    0.3285 & 0.0011 \\

    \bottomrule
    \end{tabular}
    }
    \begin{tablenotes}
        \small \item[*] Reports average value and 95\% confidence intervals in brackets.
    \end{tablenotes}
\end{threeparttable}
\moveup
\moveup
\label{tab:strategies_cost}
\end{table*}

\begin{table*}[t]
\centering
\caption{\textbf{Output-token variability across reasoning frameworks under GPT-4.1 Nano.} Best in \highlightblue{blue}, worst in \highlightorange{orange}. Strategy types are listed in \Tabref{tab:strategy_types_appx}.}
\scriptsize

\begin{threeparttable}
    \resizebox{0.99\linewidth}{!}{
    \begin{tabular}{l|cccc}
    \toprule
    \textbf{Strategy} &
    \textbf{Average*} &
    \textbf{Run Deviation (\%)*} &
    \textbf{Noise (Global)} &
    \textbf{Noise (Run)} \\
    \midrule

    \io{} &
    \highlightblue{1.08 [0.92, 1.26]} &
    4.86 [2.04, 8.33] &
    0.0065 & \highlightblue{0.0001} \\

    \cot{}~\cite{cot} &
    2.85 [2.62, 3.11] &
    5.08 [2.47, 8.23] &
    \highlightblue{0.0064} & 0.0002 \\

    \cotsc{}~\cite{cot_sc} &
    13.13 [12.95, 13.31] &
    0.98 [0.47, 1.56] &
    0.1151 & 0.0001 \\

    \hdashline

    \react{}~\cite{react} &
    4.56 [4.12, 5.83] &
    \highlightorange{14.26 [6.34, 35.12]} &
    0.0426 & 0.0015 \\

    \hdashline

    \totdfs{}~\cite{tot} &
    11.73 [10.80, 12.56] &
    3.87 [1.61, 6.66] &
    0.4673 & 0.0053 \\

    \totbfs{}~\cite{tot} &
    \highlightorange{50.16 [48.31, 51.91]} &
    5.24 [2.15, 11.36] &
    0.9659 & \highlightorange{0.0069} \\

    \got{}~\cite{got} &
    44.66 [43.55, 46.02] &
    \highlightblue{1.94 [0.85, 3.33]} &
    1.3081 & 0.0030 \\

    \hdashline

    \rap{}~\cite{rap_reasoner} &
    35.77 [34.93, 36.91] &
    8.92 [1.93, 24.09] &
    \highlightorange{1.7387} & 0.0021 \\

    \hdashline

    \foa{}~\cite{foa} &
    34.98 [34.10, 35.82] &
    3.34 [1.45, 6.14] &
    0.3486 & 0.0019 \\

    \bottomrule
    \end{tabular}
    }
    \begin{tablenotes}
        \small \item[*] Reports average value and 95\% confidence intervals in brackets.
    \end{tablenotes}
\end{threeparttable}
\moveup
\moveup
\label{tab:strategies_tokensout}
\end{table*}

\subsection{Reasoning Models}
\label{appendix:additional-results-models}
The tables below complement \Tabref{tab:models_results} with cost and output-token breakdowns for the model panel, using the same shared input--output prompting protocol and up to 30 runs per cell.

\begin{table*}[t]
\centering
\caption{\textbf{Cost variability of contemporary reasoning models across all benchmarks.} Best in \highlightblue{blue}, worst in \highlightorange{orange}. Model providers are listed in \Tabref{tab:model_providers_appx}.}
\scriptsize

\begin{threeparttable}
\resizebox{0.99\linewidth}{!}{
    \begin{tabular}{l|cccc}
    \toprule
    \textbf{Reasoning Model} &
    \textbf{Average*} & \textbf{Run Deviation (\%)*} & \textbf{\makecell{Noise (Global)}} & \textbf{Noise (Run)} \\
    \midrule


    DeepSeek R1 &
    0.0268 [0.0261, 0.0275] &
    \highlightorange{67.11 [10.73, 120.16]} &
    0.0200 & 0.2088 \\

    DeepSeek V3.2 &
    \highlightblue{0.0024 [0.0023, 0.0024]} &
    83.81 [48.26, 140.33] &
    \highlightblue{0.0011} & 0.0878 \\

    \hdashline


    GPT-4.1 mini &
    0.0003 [0.0003, 0.0004] &
    124.87 [11.57, 290.37] &
    0.0002 & 0.4457 \\

    GPT-4.1 nano &
    0.0005 [0.0005, 0.0006] &
    89.89 [5.19, 284.47] &
    0.0001 & 0.1315 \\

    Llama 4 Maverick &
    0.0007 [0.0007, 0.0007] &
    58.49 [4.43, 139.98] &
    0.0002 & 0.1017 \\

    \hdashline


    Qwen3 235B Thinking &
    \highlightorange{0.0130 [0.0125, 0.0135]} &
    78.33 [35.63, 124.76] &
    \highlightorange{0.0102} & 0.2335 \\

    GPT-5 mini &
    0.0041 [0.0040, 0.0042] &
    86.04 [20.97, 166.13] &
    0.0020 & 0.2261 \\

    GPT-5 nano &
    0.0015 [0.0014, 0.0015] &
    66.39 [18.34, 115.67] &
    0.0006 & 0.1177 \\

    GPT-5.4 mini &
    0.0036 [0.0035, 0.0037] &
    116.80 [41.10, 190.28] &
    0.0027 & 0.2862 \\

    GPT-5.4 nano &
    0.0007 [0.0006, 0.0007] &
    99.66 [25.27, 165.91] &
    0.0004 & 0.2179 \\

    o4-mini &
    0.0092 [0.0090, 0.0094] &
    83.21 [33.31, 130.87] &
    0.0050 & 0.2551 \\

    GPT-OSS 120B &
    0.0006 [0.0005, 0.0006] &
    80.68 [34.41, 143.91] &
    0.0002 & 0.2191 \\

    \hdashline


    Claude Haiku 4.5 &
    0.0055 [0.0050, 0.0061] &
    67.87 [3.62, 199.13] &
    0.0012 & \highlightblue{0.1799} \\

    Gemini 3 Flash &
    0.0196 [0.0193, 0.0198] &
    \highlightblue{59.18 [29.69, 100.83]} &
    0.0061 & 0.2654 \\

    GLM-4.5 Air &
    0.0027 [0.0027, 0.0028] &
    111.64 [50.19, 231.39] &
    0.0033 & \highlightorange{1.0172} \\

    \bottomrule
    \end{tabular}
    }
\begin{tablenotes}
    \small \item[*] Reports average value and 95\% confidence intervals in brackets.
    \item Note: Models are ordered by release date (2025). Dashed horizontal rules indicate models released in the same quarter.
\end{tablenotes}
\end{threeparttable}

\label{tab:models_cost}
\end{table*}

\begin{table*}[t]
\centering
\caption{\textbf{Output-token variability of contemporary reasoning models across all benchmarks.} Best in \highlightblue{blue}, worst in \highlightorange{orange}. Model providers are listed in \Tabref{tab:model_providers_appx}.}
\scriptsize

\begin{threeparttable}
\resizebox{0.99\linewidth}{!}{
    \begin{tabular}{l|cccc}
    \toprule
    \textbf{Reasoning Model} &
    \textbf{Average*} & \textbf{Run Deviation (\%)*} & \textbf{\makecell{Noise (Global)}} & \textbf{Noise (Run)} \\
    \midrule


    DeepSeek R1 &
    \highlightblue{3821.63 [3724.03, 3919.24]} &
    \highlightorange{68.75 [11.04, 122.38]} &
    \highlightblue{2765.8176} & 0.2335 \\

    DeepSeek V3.2 &
    3766.59 [3674.45, 3858.74] &
    87.45 [48.52, 144.93] &
    2826.5806 & 0.1014 \\

    \hdashline


    GPT-4.1 mini \textsuperscript{\dag} &
    135.49 [124.88, 146.10]  &
    161.07 [12.50, 452.65] &
    160.2179 & 0.6850 \\

    GPT-4.1 nano \textsuperscript{\dag} &
    216.02 [206.69, 225.35] &
    69.55 [5.64, 138.59] &
    312.8960 & 0.2096 \\

    Llama 4 Maverick \textsuperscript{\dag} &
    342.85 [335.29, 350.41] &
    67.44 [4.92, 144.76] &
    245.9663 & 0.1769 \\

    \hdashline


    Qwen3 235B Thinking &
    4286.57 [4130.32, 4442.83] &
    79.24 [36.17, 126.20] &
    3353.3641 & 0.2400 \\

    GPT-5 mini &
    1650.30 [1614.83, 1685.76] &
    88.13 [21.01, 170.75] &
    983.9251 & 0.2389 \\

    GPT-5 nano &
    2917.21 [2867.24, 2967.17] &
    67.14 [18.36, 116.71] &
    1411.0590 & 0.1207 \\

    GPT-5.4 mini &
    1442.46 [1399.38, 1485.53] &
    125.94 [41.35, 219.63] &
    1118.9662 & 0.3350 \\

    GPT-5.4 nano &
    1366.90 [1327.51, 1406.29] &
    106.58 [25.45, 180.65] &
    912.7655 & 0.2536 \\

    o4-mini &
    1844.51 [1803.25, 1885.77] &
    86.85 [33.42, 134.29] &
    1145.1105 & 0.2762 \\

    GPT-OSS 120B &
    \highlightorange{894.71 [872.71, 916.71]} &
    \highlightblue{87.19 [35.15, 149.81]} &
    412.6778 & 0.2586 \\

    \hdashline


    Claude Haiku 4.5 &
    387.14 [372.47, 401.81] &
    72.69 [3.81, 215.72] &
    222.8693 & \highlightblue{0.2897} \\

    Gemini 3 Flash &
    6474.82 [6389.47, 6560.17] &
    59.55 [29.71, 101.81] &
    2018.9410 & 0.2728 \\

    GLM-4.5 Air &
    5499.17 [5345.06, 5653.28] &
    112.19 [50.31, 231.91] &
    \highlightorange{3816.3473} & \highlightorange{1.0332} \\

    \bottomrule
    \end{tabular}
    }
\begin{tablenotes}
    \small \item[*] Reports average value and 95\% confidence intervals in brackets.
    \small \item[\dag] Non-reasoning models.
    \item Note: Models are ordered by release date (2025). Dashed horizontal rules indicate models released in the same quarter.
\end{tablenotes}
\end{threeparttable}

\label{tab:models_tokensout}
\end{table*}

\subsection{Decoding Configuration Sweeps}
\label{appx:decoding}

The decoding-randomness study sweeps temperature for a representative subset of \emph{reasoning strategies}, using \textsc{GPT-4.1 Nano} as the underlying model. We expand on the per-cell interpretation here.

\xhdr{Per-cell strategy view} \Figref{fig:temperature_grid} in the main paper shows the run-level score distribution at $T \in \{0, 0.35, 0.7\}$ for every (strategy, benchmark) cell in the sweep: \foa{}, \react{}, and \totbfs{} over Game24, SciBench, and SonnetWriting. Within each cell the inter-quartile range at $T=0$ is comparable to, and in several cases larger than, the corresponding range at $T=0.7$. Two cells degenerate to a point mass and contain no run-level variance to begin with (\foa{} on SonnetWriting saturates at $1.00$; \react{} on SonnetWriting collapses at $0.00$); these are uninformative for the question at hand and we exclude them from the discussion.

\subsection{Prompt and Parser Specification}
\label{appx-prompts}

\Tabref{tab:frameworks_quality_diff_appx} reports the impact of prompt and parser refinements on mean quality with point estimates, 95\% bootstrap confidence intervals, the absolute change $\Delta$, strategy type, and statistical-significance markers. All ten strategies show a positive $\Delta$ that is statistically significant at $p < 0.05$, while the within-cell confidence-interval widths remain comparable across the original and refined conditions. Refinement therefore reduces bias without affecting run-to-run variance.

\begin{table}[t]
\centering
\caption{
\textbf{Impact of prompt and parsing refinements on strategy performance.}
Enhancing clarity and standardizing output parsing significantly improves accuracy without affecting the stability. Direct prompting methods show the largest gains while the rest showcase similar ones, except \rap{} which improves the least.
The best performance is shown in \highlightblue{blue} whereas the worst is shown in \highlightorange{orange}.
}
\label{tab:frameworks_quality_diff_appx}
\begin{threeparttable}
\resizebox{1.0\linewidth}{!}{
    \begin{tabular}{ll|c|c|c}
    \toprule
    \textbf{Strategy} & \textbf{Type} &
    \textbf{Original Prompts*} &
    \textbf{Improved Prompts*} &
    $\Delta$ \\
    \midrule
    
    \io{} & Direct &
    \highlightorange{0.106 [0.10, 0.12]} &
    0.313 [0.28, 0.34] &
    \highlightblue{+0.207}\textsuperscript{$\dagger$} \\
    
    \cot{} & Direct &
    0.276 [0.25, 0.30] &
    0.398 [0.35, 0.43] &
    +0.122\textsuperscript{$\dagger$} \\
    
    \cotsc{} & Direct &
    0.228 [0.21, 0.24] &
    0.410 [0.40, 0.45] &
    +0.182\textsuperscript{$\dagger$} \\ 
    \hdashline
    
    \react{} & Adaptive &
    0.295 [0.28, 0.31] &
    0.391 [0.36, 0.42] &
    +0.096\textsuperscript{$\dagger$} \\
    
    \reflexion{} & Adaptive &
    0.282 [0.27, 0.30] &
    0.411 [0.39, 0.42] &
    +0.129\textsuperscript{$\dagger$} \\
    \hdashline
    
    \totdfs{} & Structured &
    0.127 [0.10, 0.14] &
    \highlightorange{0.177 [0.15, 0.20]} &
    +0.050\textsuperscript{$\dagger$} \\
    
    \got{} & Structured &
    0.3361 [0.31, 0.36] &
    0.420 [0.39, 0.46] &
    +0.084\textsuperscript{$\dagger$} \\

    \totbfs{} & Structured &
    0.407 [0.38, 0.44] &
    0.506 [0.47, 0.54] &
    +0.099\textsuperscript{$\dagger$} \\
    \hdashline
    
    \rap{} & Planning &
    0.367 [0.35, 0.38] &
    0.403 [0.39, 0.41] &
    \highlightorange{+0.036}\textsuperscript{$\dagger$} \\
    \hdashline
    
    \foa{} & Evolutionary &
    \highlightblue{0.4580 [0.43, 0.48]} &
    \highlightblue{0.546 [0.52, 0.58]} &
    +0.088\textsuperscript{$\dagger$} \\
    
    \bottomrule
    \end{tabular}
}
\vspace{2pt}
\begin{tablenotes}
        \small \item[\dag] Indicates statistical significance ($p<0.05$) from original. 
        \item[*] Reports average quality and 95\% confidence intervals in \\brackets.
    \end{tablenotes}
\end{threeparttable}
\end{table}

\subsection{Causal Analysis: Evaluator Quality}
\label{appendix:causal-analysis}
The causal intervention isolates the effect of replacing heuristic evaluation with ground-truth evaluation. Holding the search procedure fixed, the oracle evaluator raises mean quality and tightens the run distribution for all three search-based strategies on Game of 24 (see \Figref{fig:evaluator_quality} in the main paper). Evaluation quality therefore affects not only expected performance but also run-to-run variability.

\subsection{Reasoning Effort}
\label{appx-thinking}

We extend the reasoning-effort study by jointly analyzing mean behavior and stability under test-time compute scaling. We evaluate \textsc{GPT-5 Nano}, \textsc{GPT-5 Mini}, and \textsc{Gemini-3 Flash} at three effort settings (\emph{low}, \emph{medium}, \emph{high}) on the full \RB suite, with 10 independent runs per (model, effort) configuration. For each model we measure (\emph{i}) \textbf{Score} as the primary quality metric, (\emph{ii}) \textbf{tokens\_out} as a direct proxy for generation cost, and (\emph{iii}) \textbf{Efficiency} as a quality--cost trade-off. For each quantity we report \textbf{Average} performance using stratified bootstrap confidence intervals across benchmarks; \textbf{Run Deviation} (relative instability) as the typical percent deviation of runs from their benchmark-level mean; and \textbf{Global Noise} as the variance of z-scored outcomes across benchmarks, capturing benchmark-dependence after normalizing for task difficulty. The main-paper summary is shown in \Figref{fig:effort_panel_main}; the per-sample and per-benchmark breakdowns below provide the supporting detail.

\subsubsection{Per-sample significance under reasoning-effort scaling}
\label{appendix:effort-sample-significance}

To complement aggregate averages, we analyze how reasoning effort affects outcomes at the level of individual benchmark instances. For each model, benchmark, and effort setting, we estimate per-sample quality together with uncertainty by aggregating over 10 independent runs and computing a confidence interval for each sample. We then compare effort levels pairwise (\emph{low}$\rightarrow$\emph{medium}, \emph{medium}$\rightarrow$\emph{high}, \emph{low}$\rightarrow$\emph{high}) and classify each sample as \emph{improved}, \emph{worsened}, or \emph{unaffected} based on whether the confidence intervals indicate a statistically significant increase, a statistically significant decrease, or an overlap. We report, for each of the three models and each benchmark, the percentage of samples in each category, quantifying how often increased effort yields meaningful gains or regressions beyond what is explained by run-to-run variability.

\begin{figure}[tbp]
    \centering
    \includegraphics[width=0.96\linewidth]{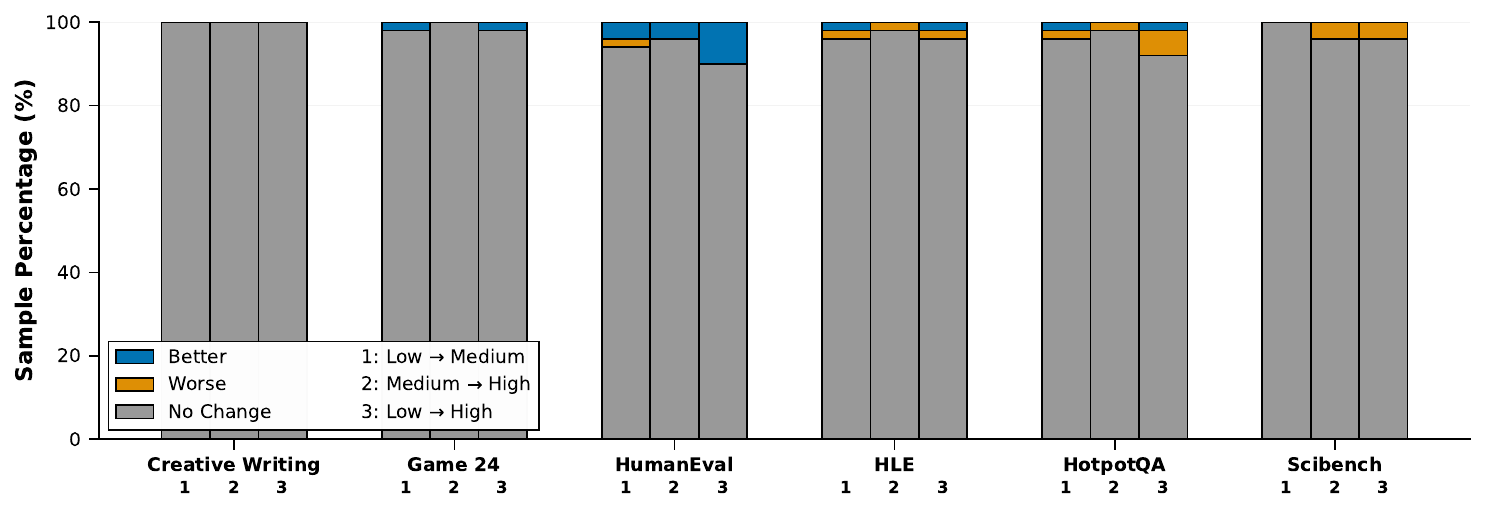}
    \caption{Per-sample significance analysis for \textsc{GPT-5 Nano}: for each benchmark, stacked bars show the percentage of instances whose quality is \emph{better}, \emph{worse}, or shows \emph{no statistically significant change} when increasing \texttt{reasoning\_effort}, based on confidence-interval comparisons over 10 independent runs.}
    \label{fig:effort_addendum_gpt5nano}
\end{figure}

\begin{figure}[tbp]
    \centering
    \includegraphics[width=\linewidth]{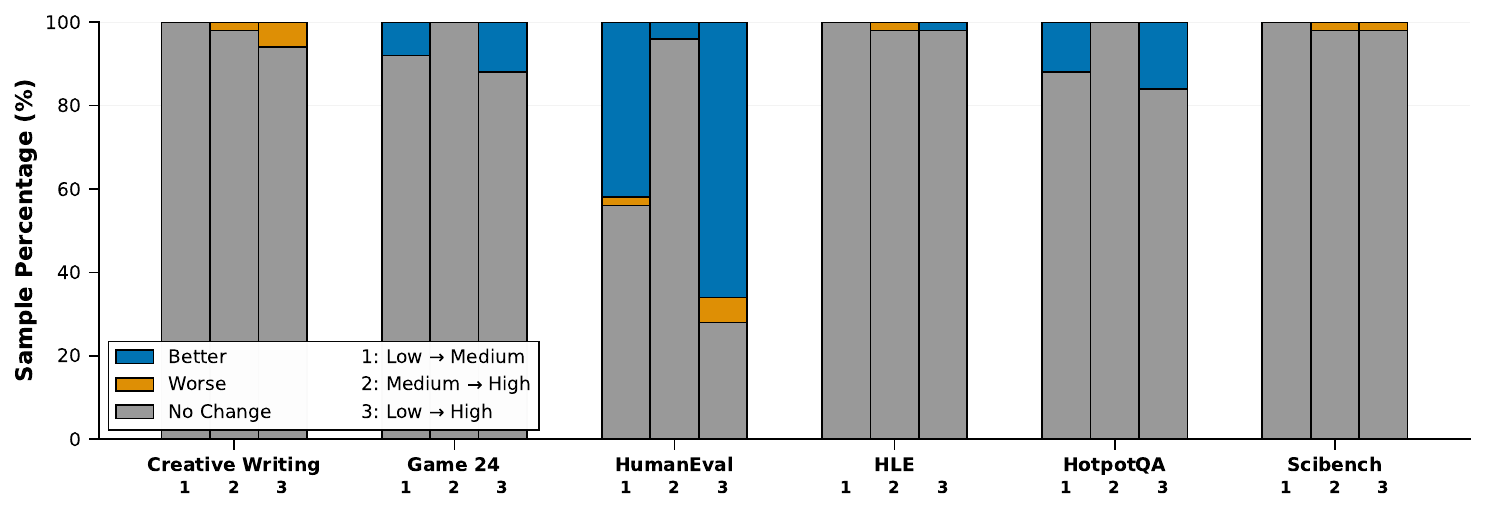}
    \caption{Per-sample significance analysis for \textsc{GPT-5 Mini}; conventions as in \Figref{fig:effort_addendum_gpt5nano}.}
    \label{fig:effort_addendum_gpt5mini}
\end{figure}

\begin{figure}[tbp]
    \centering
    \includegraphics[width=\linewidth]{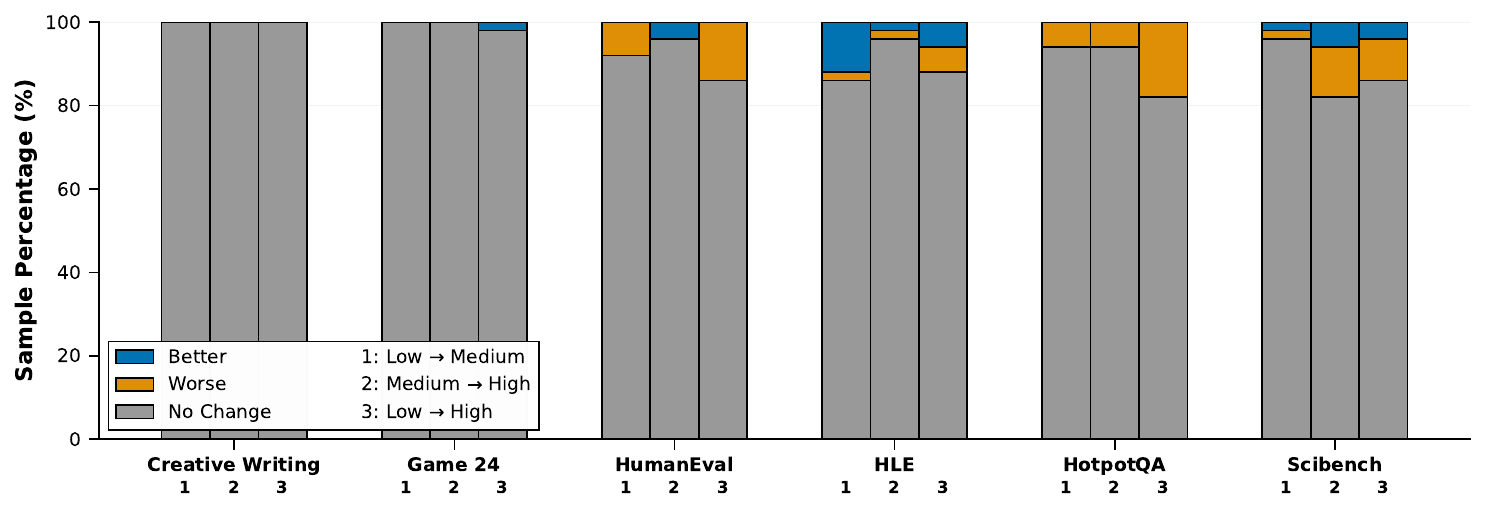}
    \caption{Per-sample significance analysis for \textsc{Gemini-3 Flash}; conventions as in \Figref{fig:effort_addendum_gpt5nano}.}
    \label{fig:effort_addendum_gemini3}
\end{figure}

\subsubsection{Per-benchmark effort results}
\label{appendix:effort-per-benchmark}

\Figref{fig:effort_per_benchmark} reports the per-benchmark counterpart of the aggregated reasoning-effort study. Each row corresponds to one benchmark and shows the two performance axes of the effort sweep side by side: mean quality (left, Score) and generation cost (right, output tokens) at four effort settings (\emph{minimal}, \emph{low}, \emph{medium}, \emph{high}) for \textsc{GPT-5 Nano}, \textsc{GPT-5 Mini}, and \textsc{Gemini-3 Flash}. Markers report bootstrap means over 10 independent runs with 95\% confidence intervals.

\begin{figure}[t]
    \centering
    \begin{subfigure}{0.78\linewidth}
        \centering
        \includegraphics[width=\linewidth]{"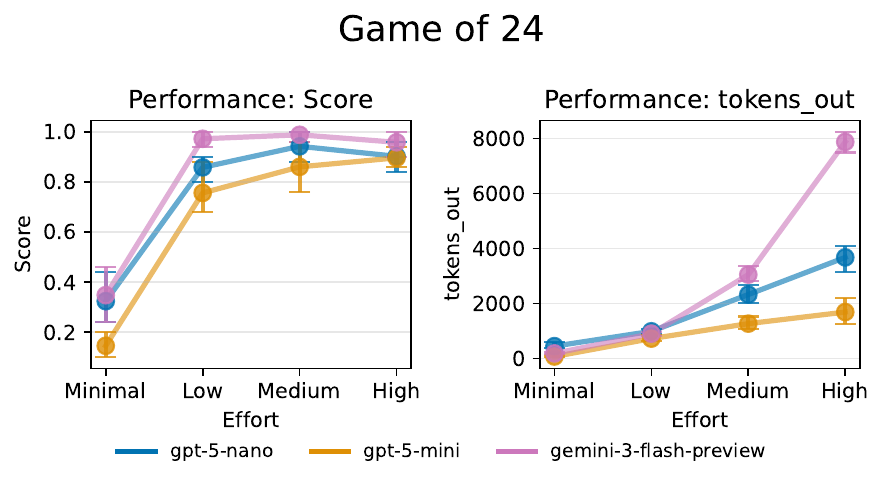"}
        \caption{Game of 24}
        \label{fig:effort_pb_game24}
    \end{subfigure}\\[2pt]
    \begin{subfigure}{0.78\linewidth}
        \centering
        \includegraphics[width=\linewidth]{"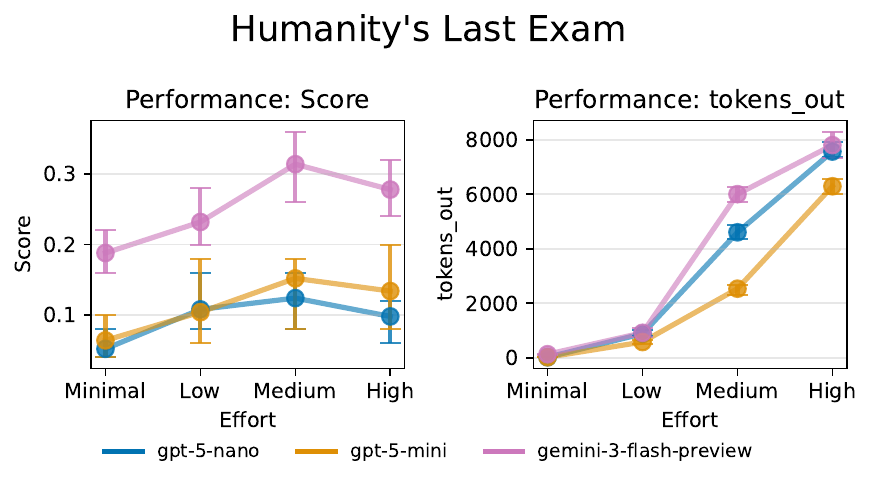"}
        \caption{Humanity's Last Exam}
        \label{fig:effort_pb_hle}
    \end{subfigure}\\[2pt]
    \begin{subfigure}{0.78\linewidth}
        \centering
        \includegraphics[width=\linewidth]{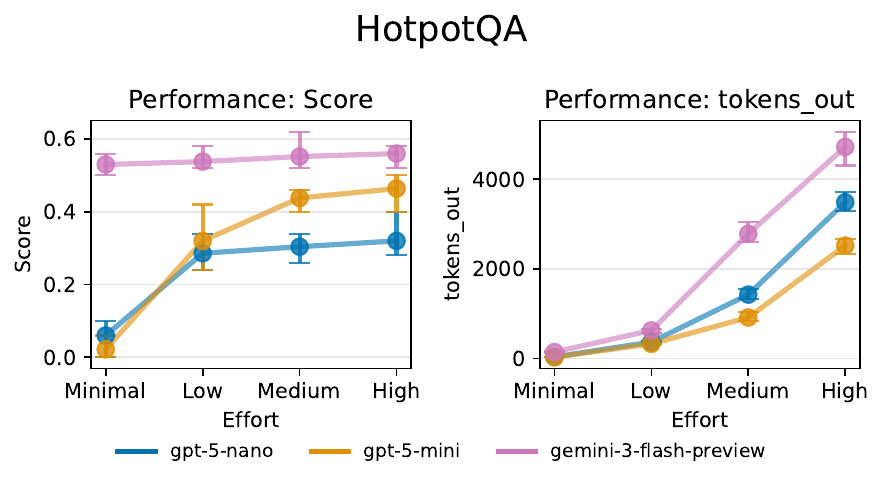}
        \caption{HotpotQA}
        \label{fig:effort_pb_hotpotqa}
    \end{subfigure}
    \caption{\textbf{Reasoning-effort sweep, per-benchmark (1/2).} Each row shows mean Score and output-token cost across four effort settings for \textsc{GPT-5 Nano}, \textsc{GPT-5 Mini}, and \textsc{Gemini-3 Flash}, with 95\% bootstrap CIs over 10 runs. Continued on the next page.}
    \label{fig:effort_per_benchmark}
\end{figure}

\begin{figure}[t]\ContinuedFloat
    \centering
    \begin{subfigure}{0.78\linewidth}
        \centering
        \includegraphics[width=\linewidth]{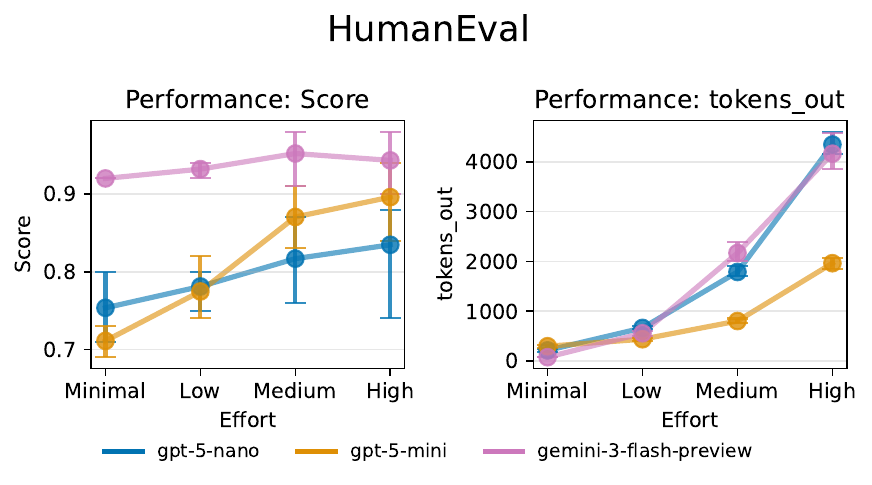}
        \caption{HumanEval}
        \label{fig:effort_pb_humaneval}
    \end{subfigure}\\[2pt]
    \begin{subfigure}{0.78\linewidth}
        \centering
        \includegraphics[width=\linewidth]{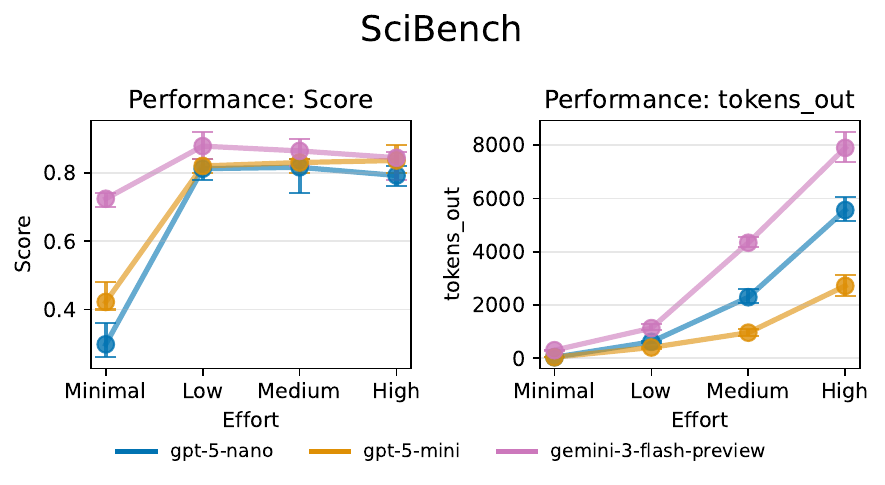}
        \caption{SciBench}
        \label{fig:effort_pb_scibench}
    \end{subfigure}\\[2pt]
    \begin{subfigure}{0.78\linewidth}
        \centering
        \includegraphics[width=\linewidth]{"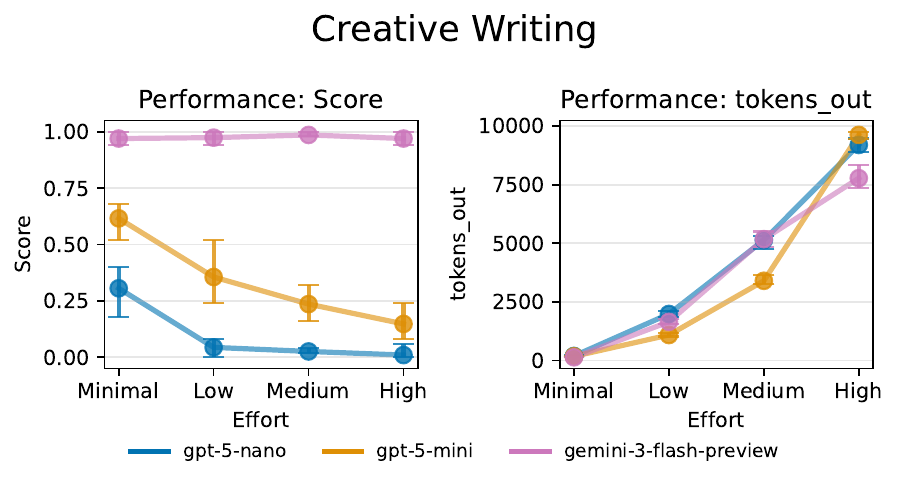"}
        \caption{Creative Writing}
        \label{fig:effort_pb_sonnetwriting}
    \end{subfigure}
    \caption{\textbf{Reasoning-effort sweep, per-benchmark (2/2).}}
    \label{fig:effort_per_benchmark_cont}
\end{figure}

\subsection{Model Scale}
\label{appendix:model_scale}
Comparing GPT-4.1 Nano with GPT-4.1 Mini, GPT-5 Nano with GPT-5 Mini, and GPT-5.4 Nano with GPT-5.4 Mini under the same strategies and prompts, the larger sibling in each pair improves mean quality and tightens run distributions across the panel. Model capacity is therefore a stabilizing factor within a family, although it does not eliminate variance entirely.

\subsection{Reasoning Models: Per-Benchmark Results}
\label{appendix:model-per-benchmark}

\subsubsection{Score}
\label{appendix:model-per-benchmark-score}
The figures below report per-benchmark run-level score distributions for the model panel, complementing \Tabref{tab:models_results} which aggregates across benchmarks.

\begin{figure}[tbp]
    \centering
    \includegraphics[width=0.96\linewidth]{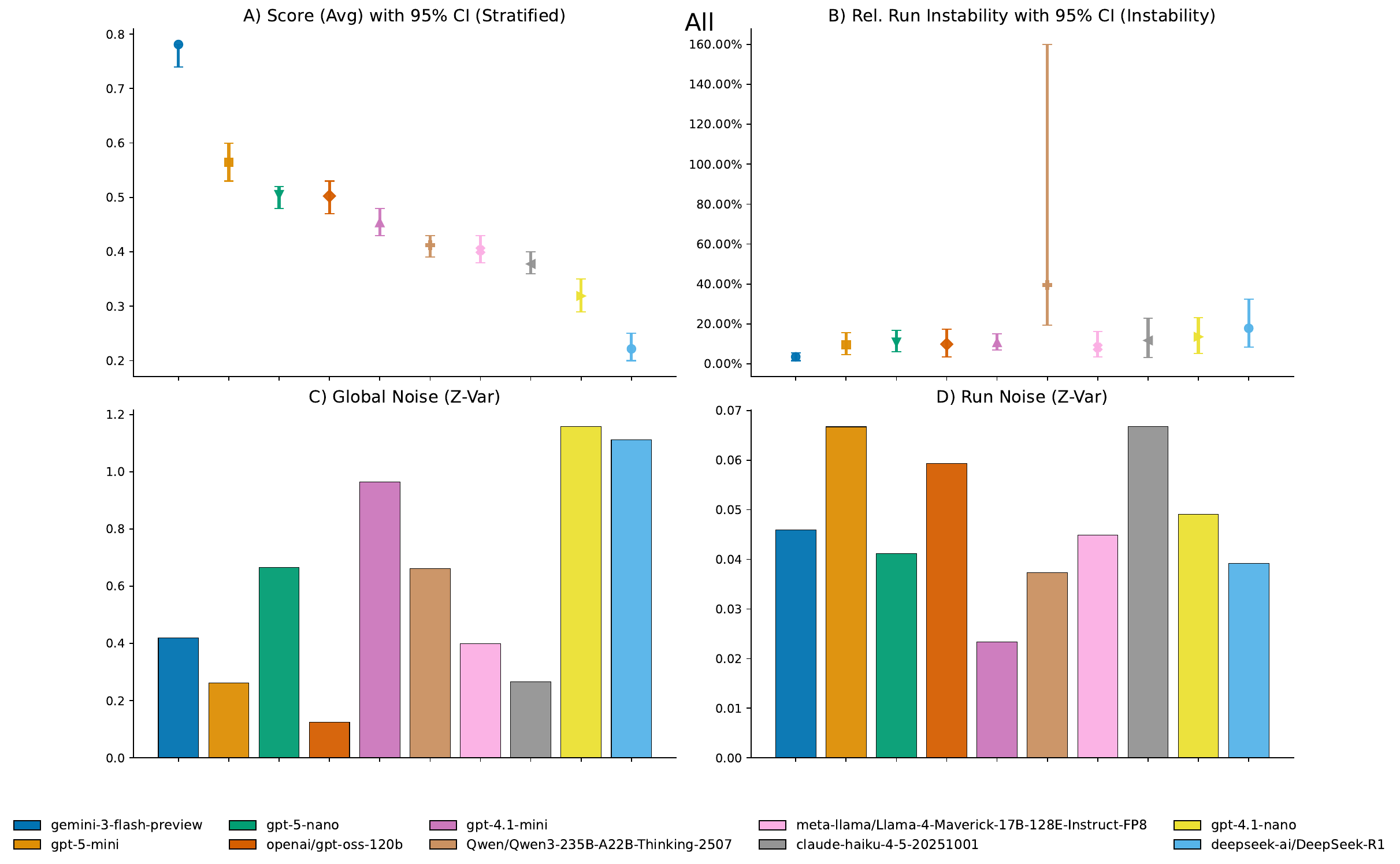}
    \caption{Score distributions across all benchmarks.}
    \label{fig:model_score_all}
\end{figure}

\begin{figure}[tbp]
    \centering
    \includegraphics[width=0.96\linewidth]{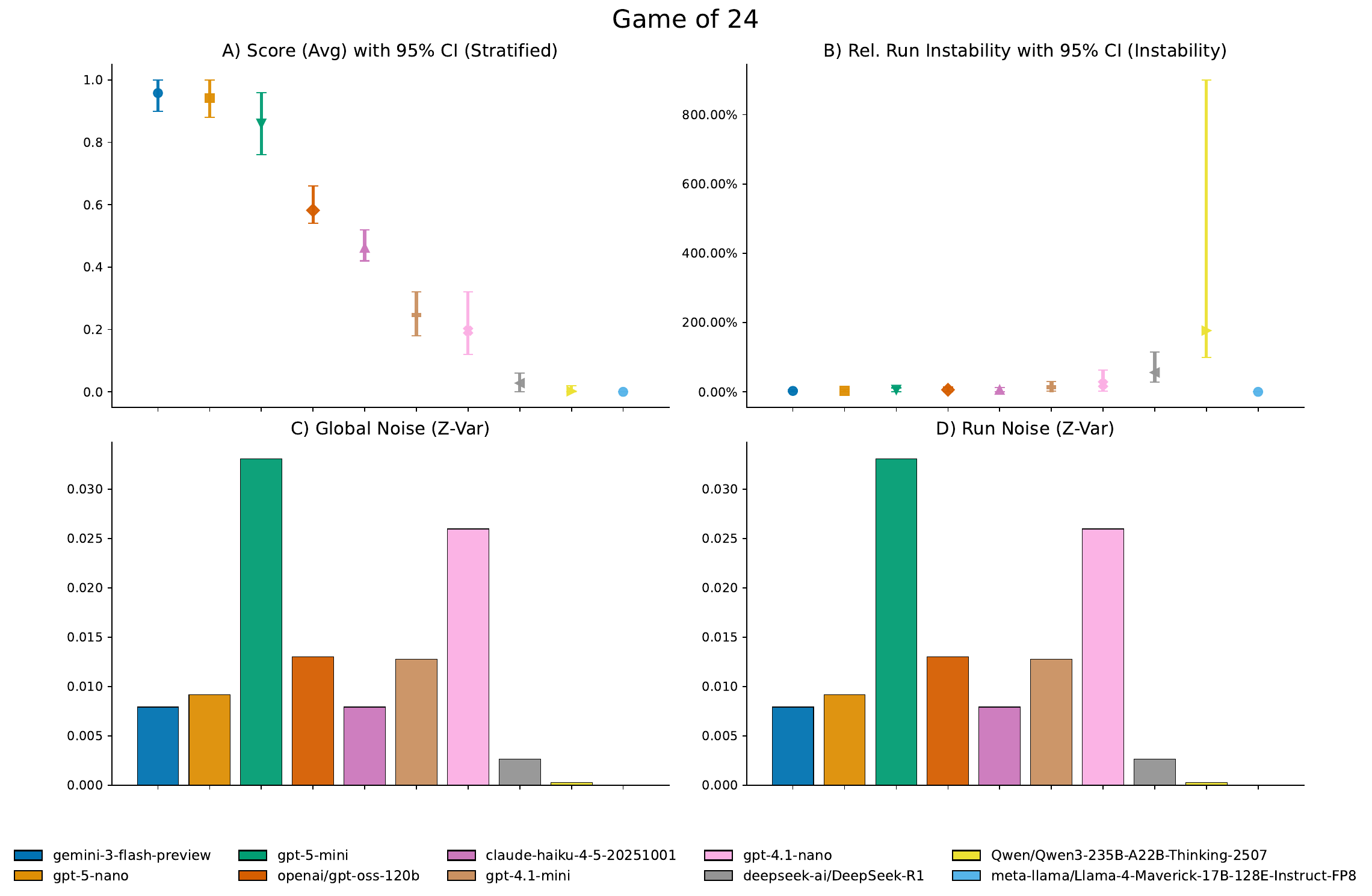}
    \caption{Score distributions on Game of 24.}
    \label{fig:model_score_game24}
\end{figure}

\begin{figure}[tbp]
    \centering
    \includegraphics[width=0.96\linewidth]{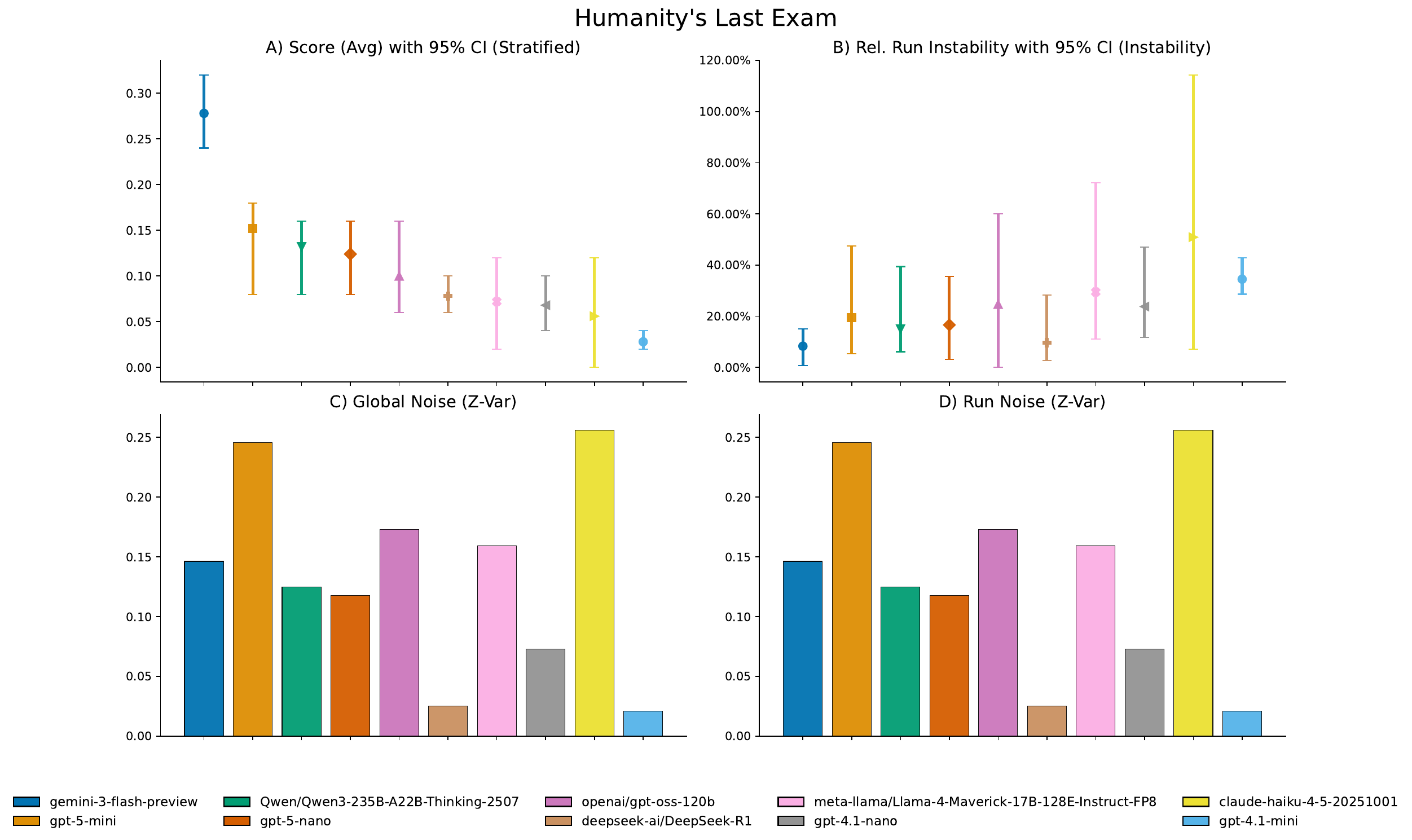}
    \caption{Score distributions on Humanity's Last Exam.}
    \label{fig:model_score_hle}
\end{figure}

\begin{figure}[tbp]
    \centering
    \includegraphics[width=0.96\linewidth]{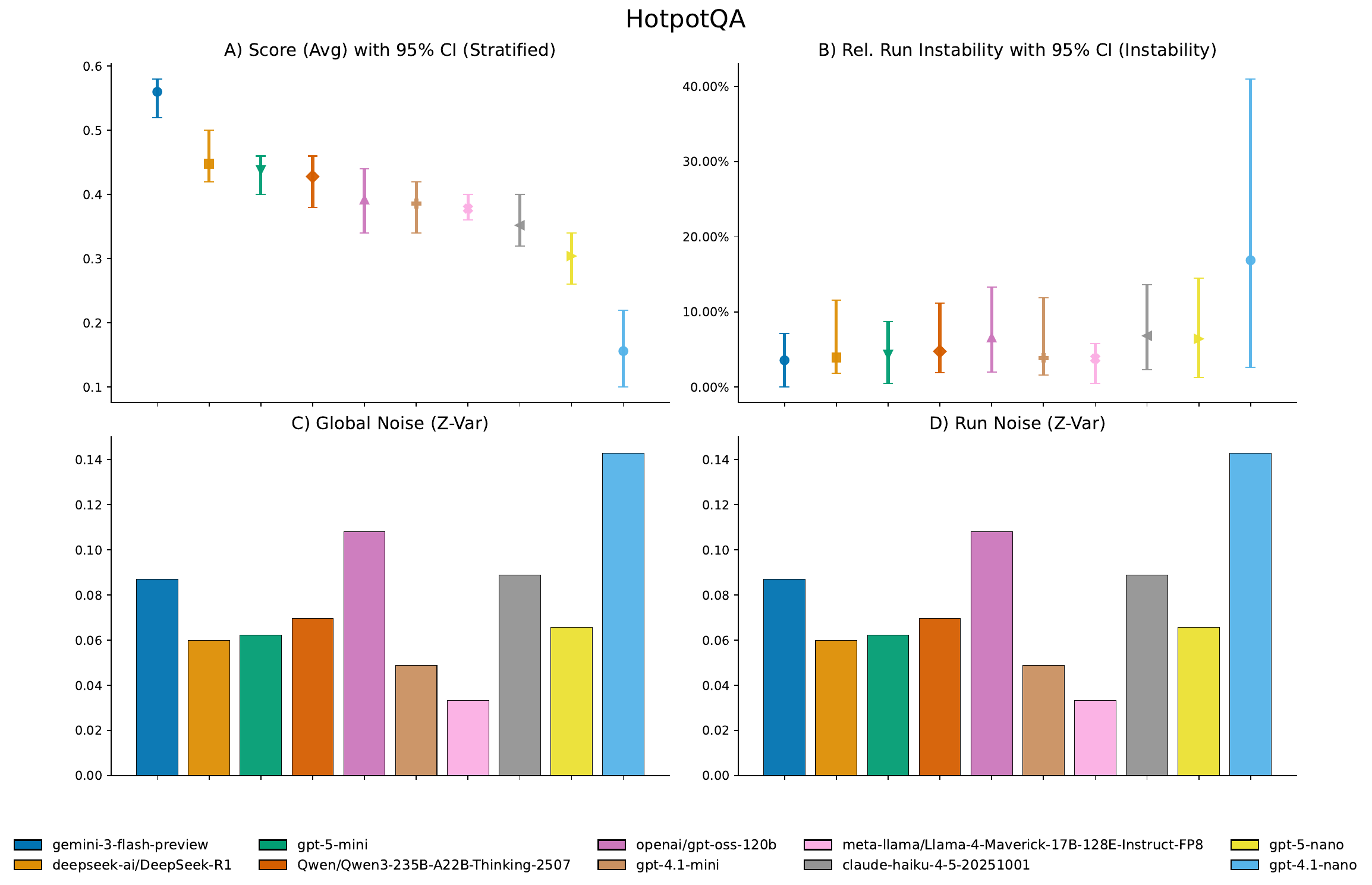}
    \caption{Score distributions on HotpotQA.}
    \label{fig:model_score_hotpotqa}
\end{figure}

\begin{figure}[tbp]
    \centering
    \includegraphics[width=0.96\linewidth]{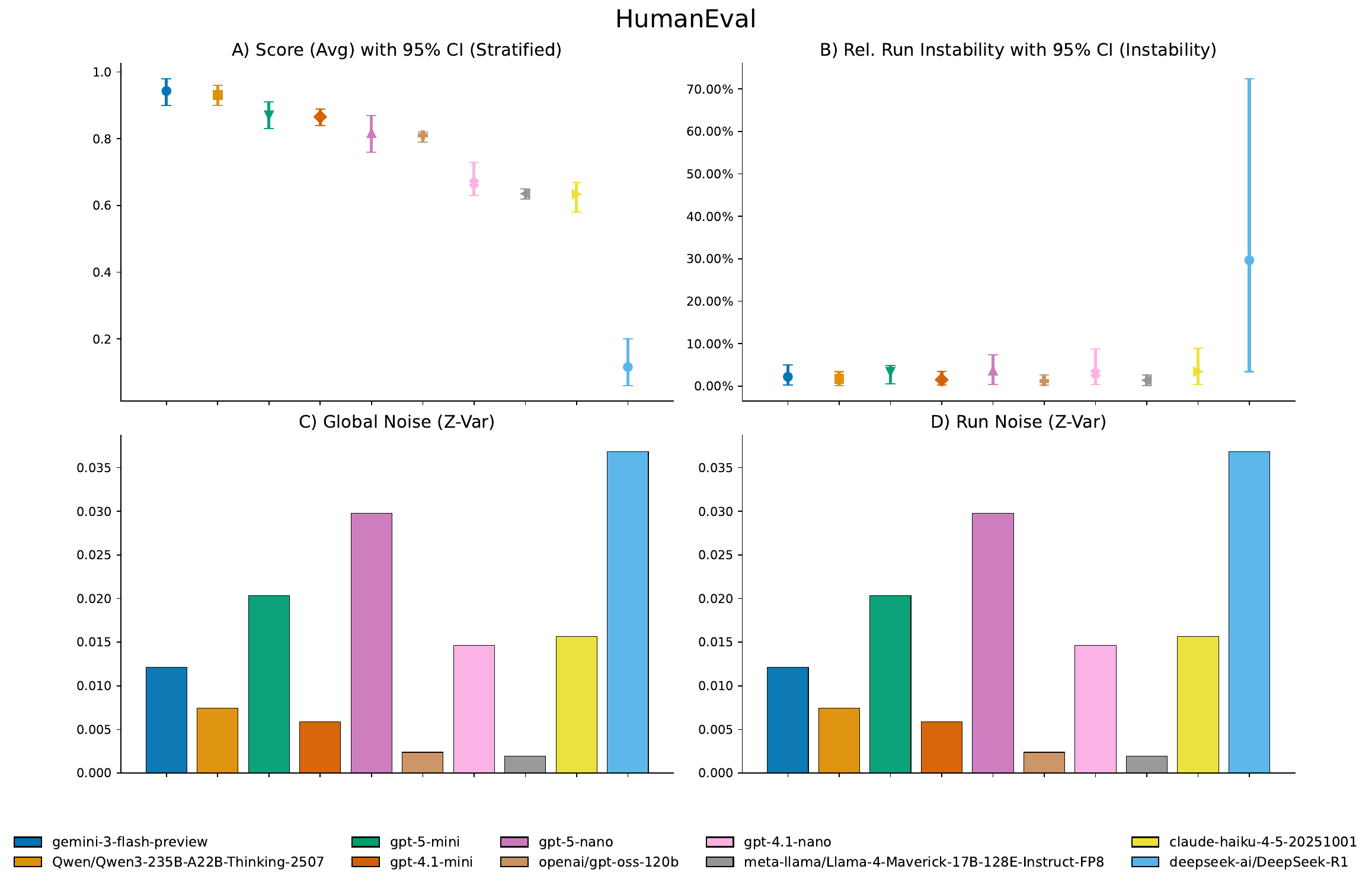}
    \caption{Score distributions on HumanEval.}
    \label{fig:model_score_humaneval}
\end{figure}

\begin{figure}[tbp]
    \centering
    \includegraphics[width=0.96\linewidth]{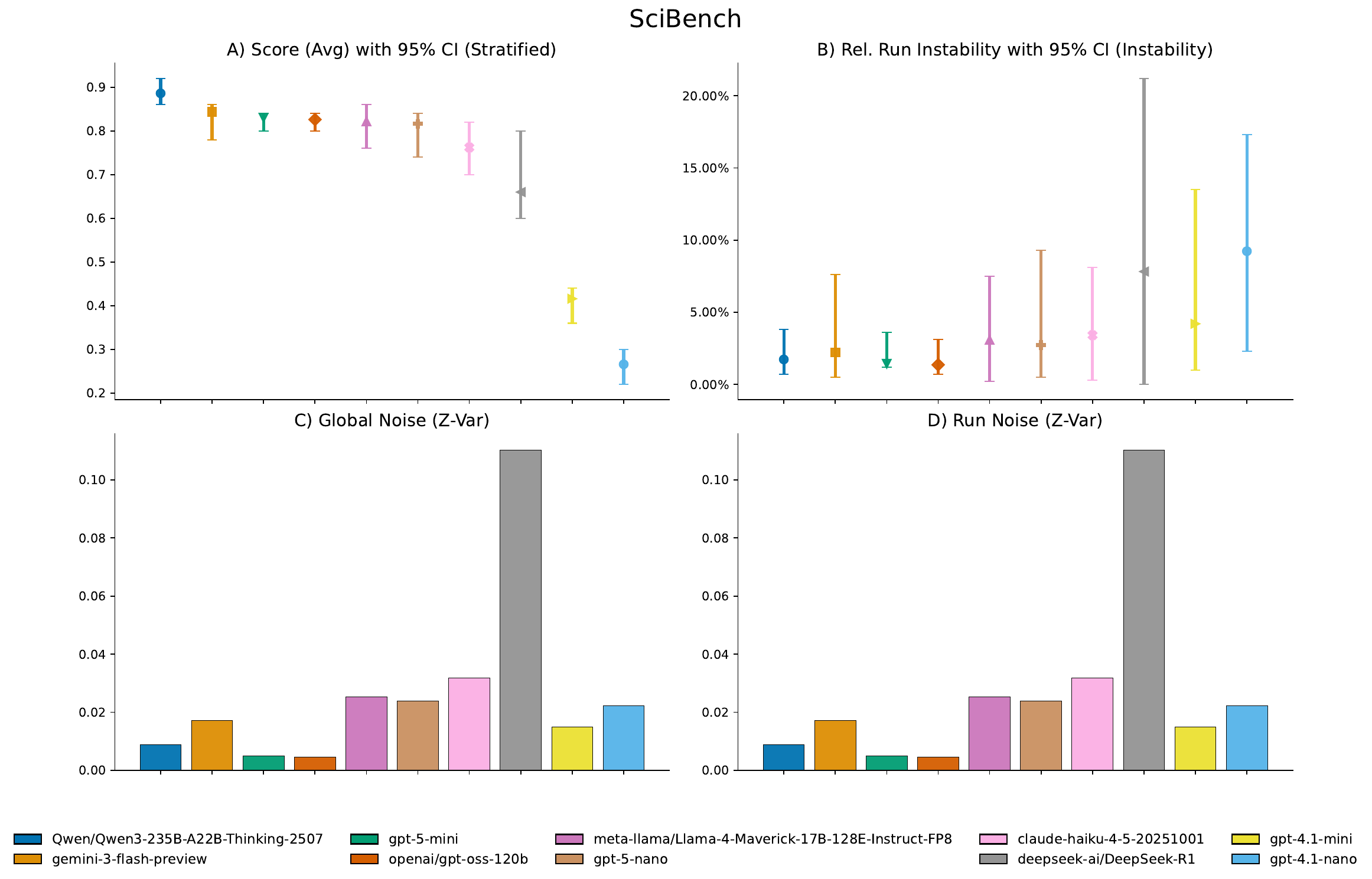}
    \caption{Score distributions on SciBench.}
    \label{fig:model_score_scibench}
\end{figure}

\begin{figure}[tbp]
    \centering
    \includegraphics[width=0.96\linewidth]{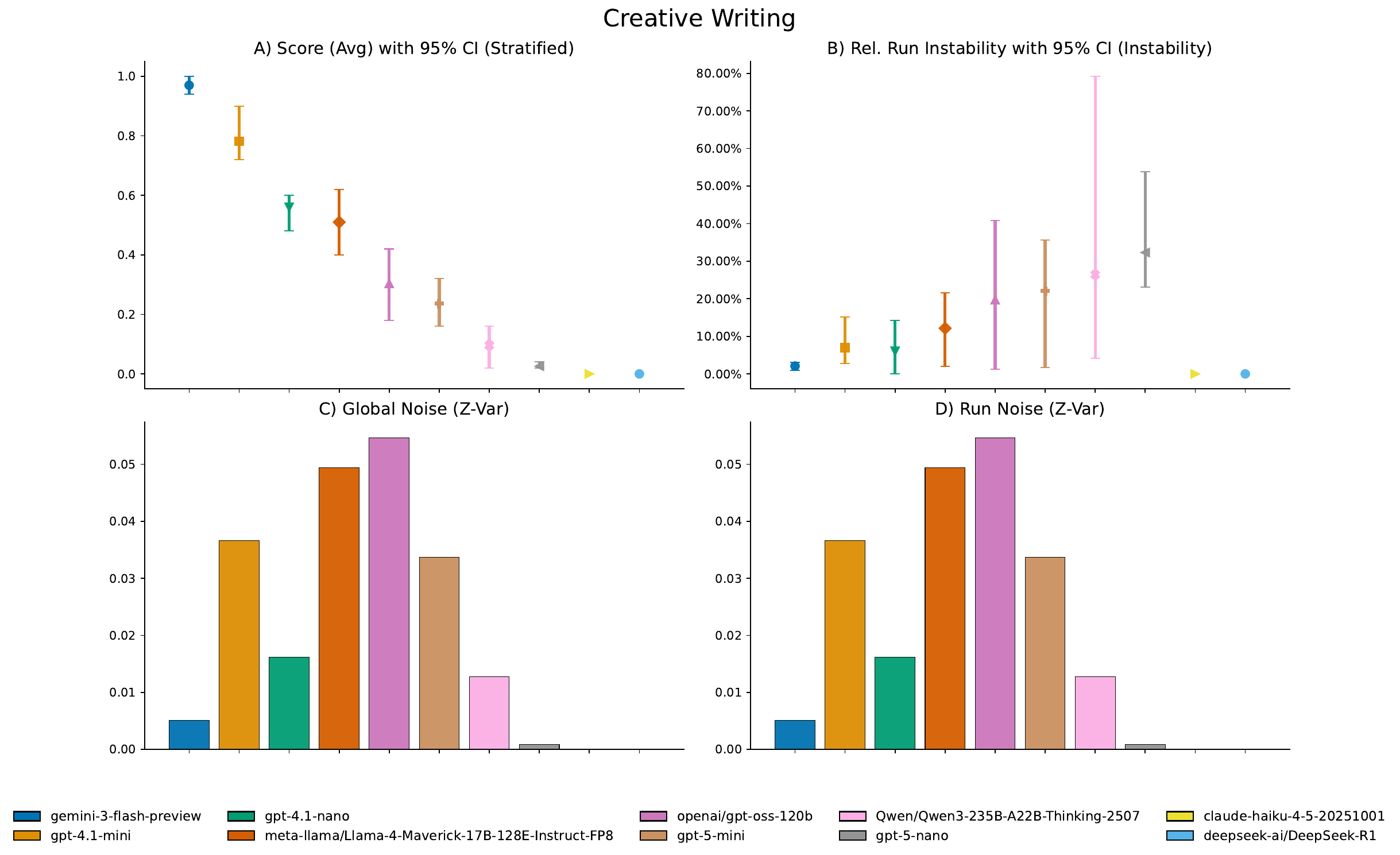}
    \caption{Score distributions on Shakespearean Sonnet Writing.}
    \label{fig:model_score_sonnetwriting}
\end{figure}

\subsubsection{Cost}
\label{appendix:model-per-benchmark-cost}
The figures below report per-benchmark run-level cost distributions for the model panel.

\begin{figure}[tbp]
    \centering
    \includegraphics[width=0.96\linewidth]{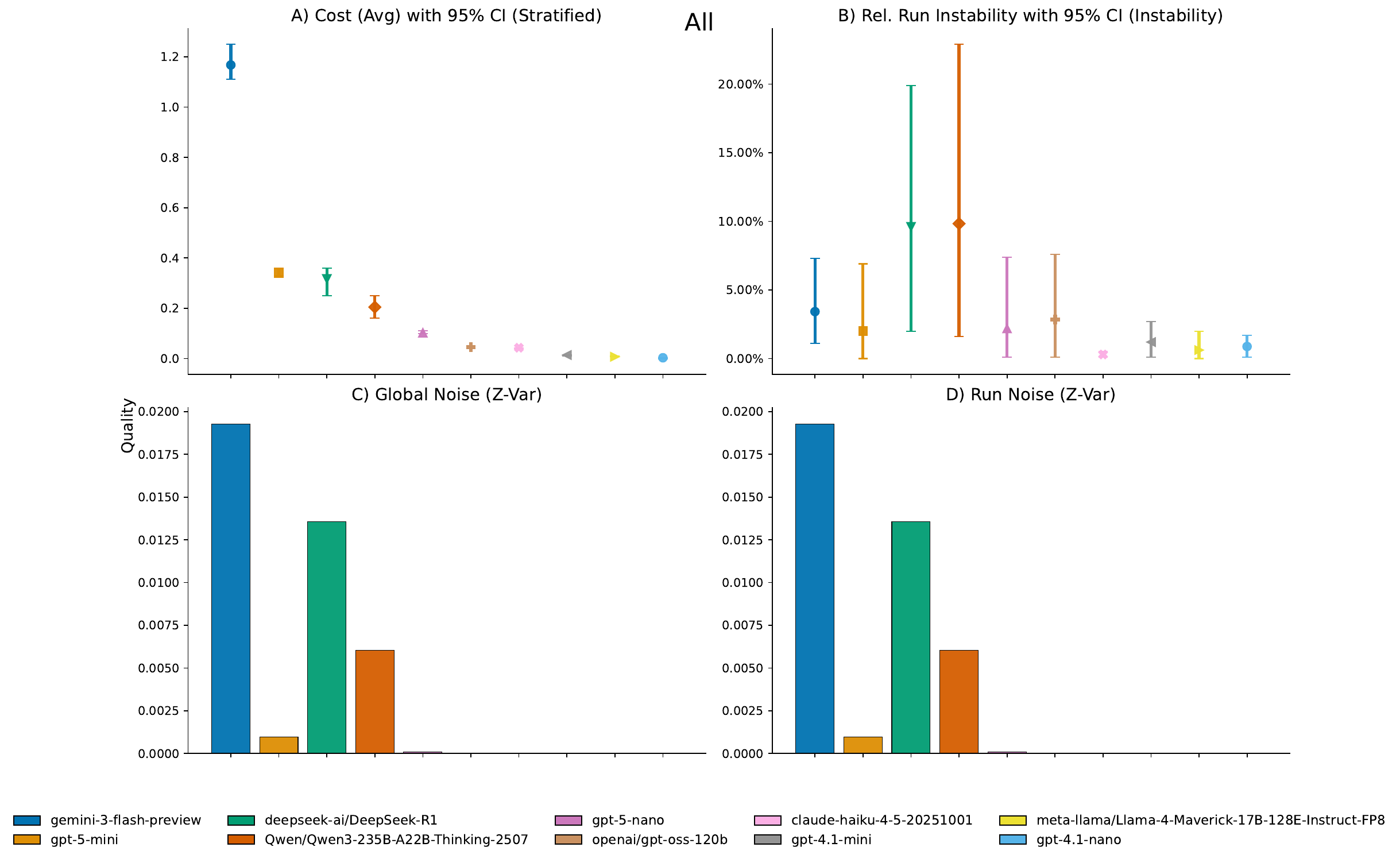}
    \caption{Cost distributions across all benchmarks.}
    \label{fig:model_cost_all}
\end{figure}

\begin{figure}[tbp]
    \centering
    \includegraphics[width=0.96\linewidth]{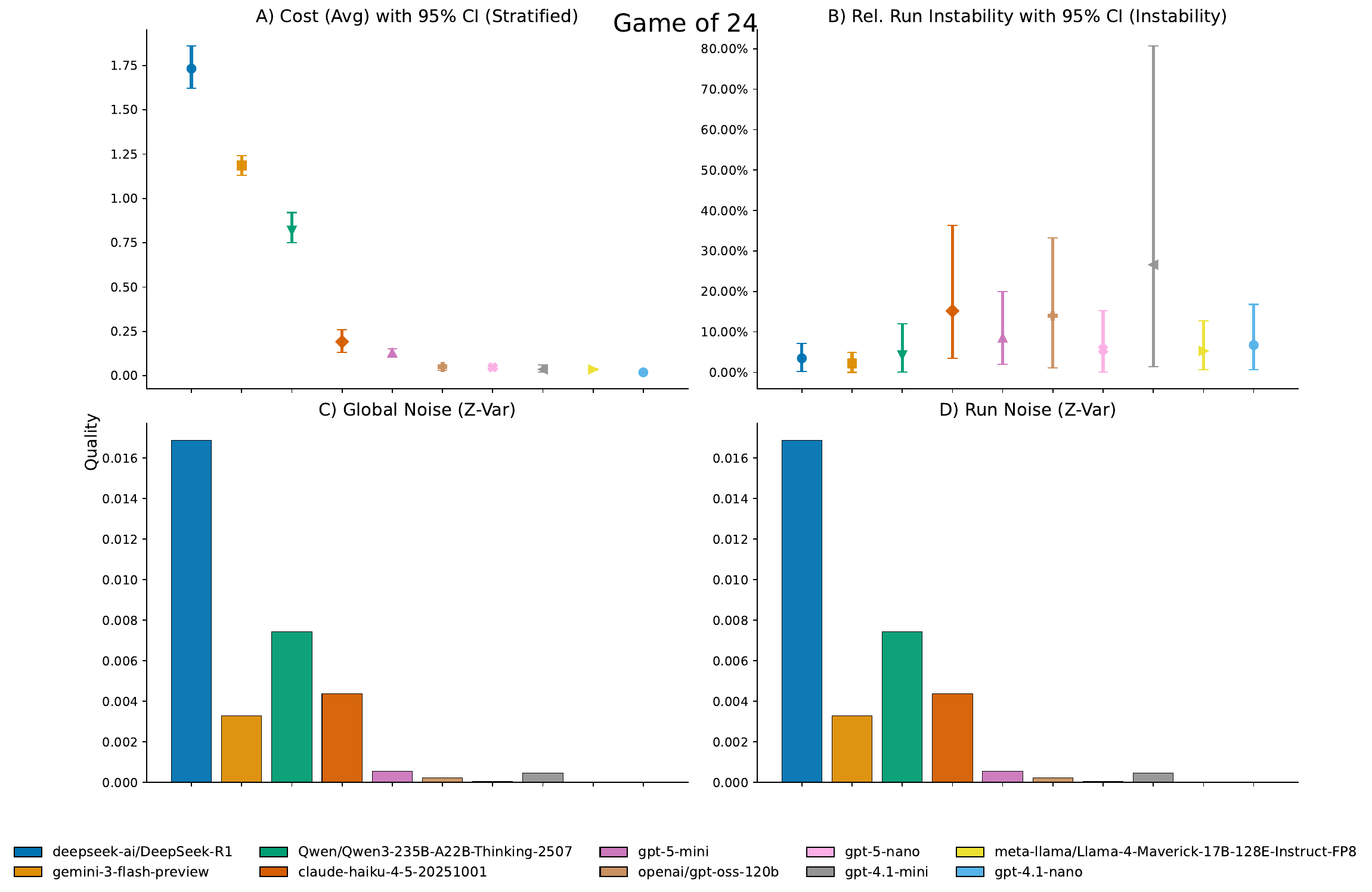}
    \caption{Cost distributions on Game of 24.}
    \label{fig:model_cost_game24}
\end{figure}

\begin{figure}[tbp]
    \centering
    \includegraphics[width=0.96\linewidth]{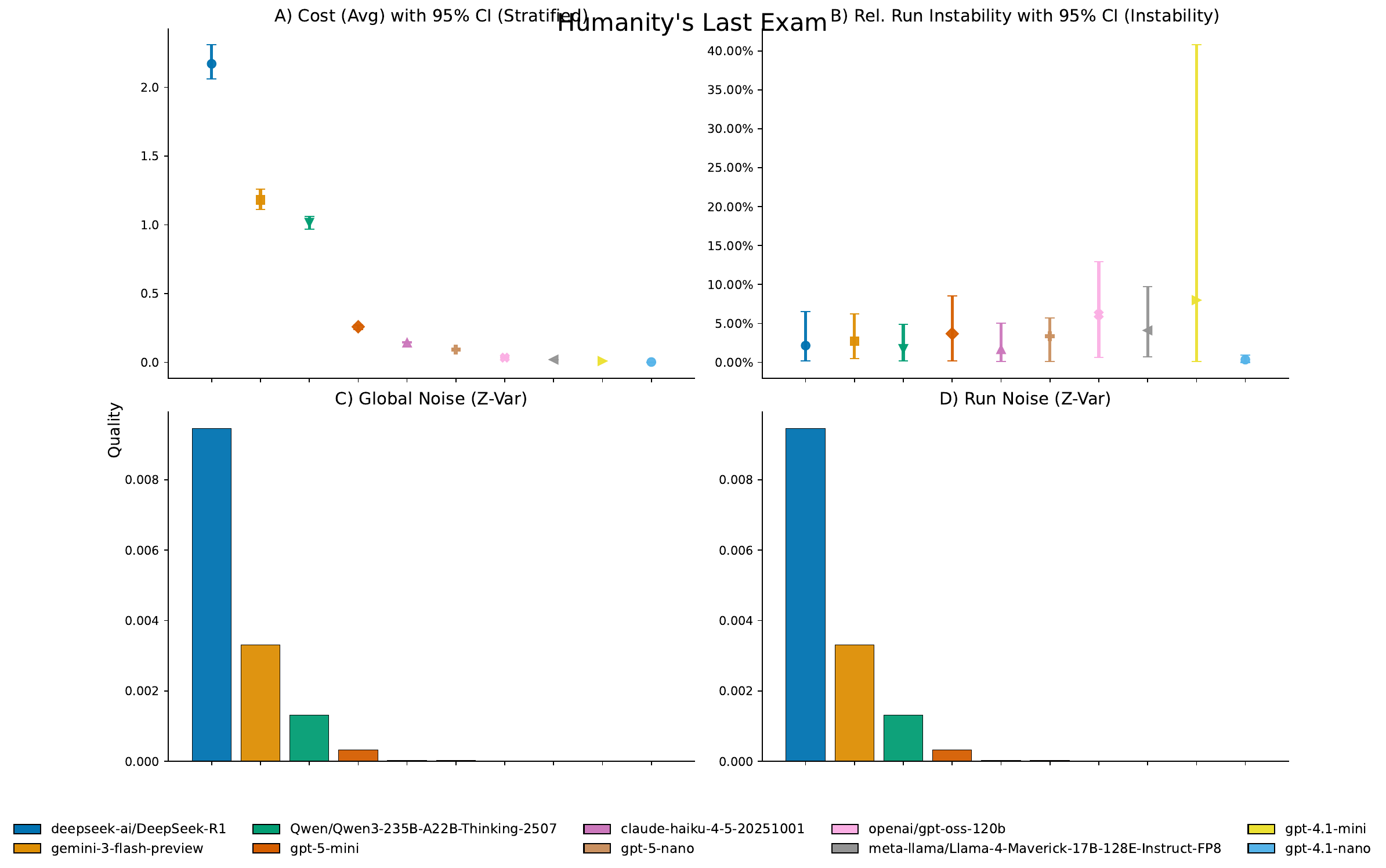}
    \caption{Cost distributions on Humanity's Last Exam.}
    \label{fig:model_cost_hle}
\end{figure}

\begin{figure}[tbp]
    \centering
    \includegraphics[width=0.96\linewidth]{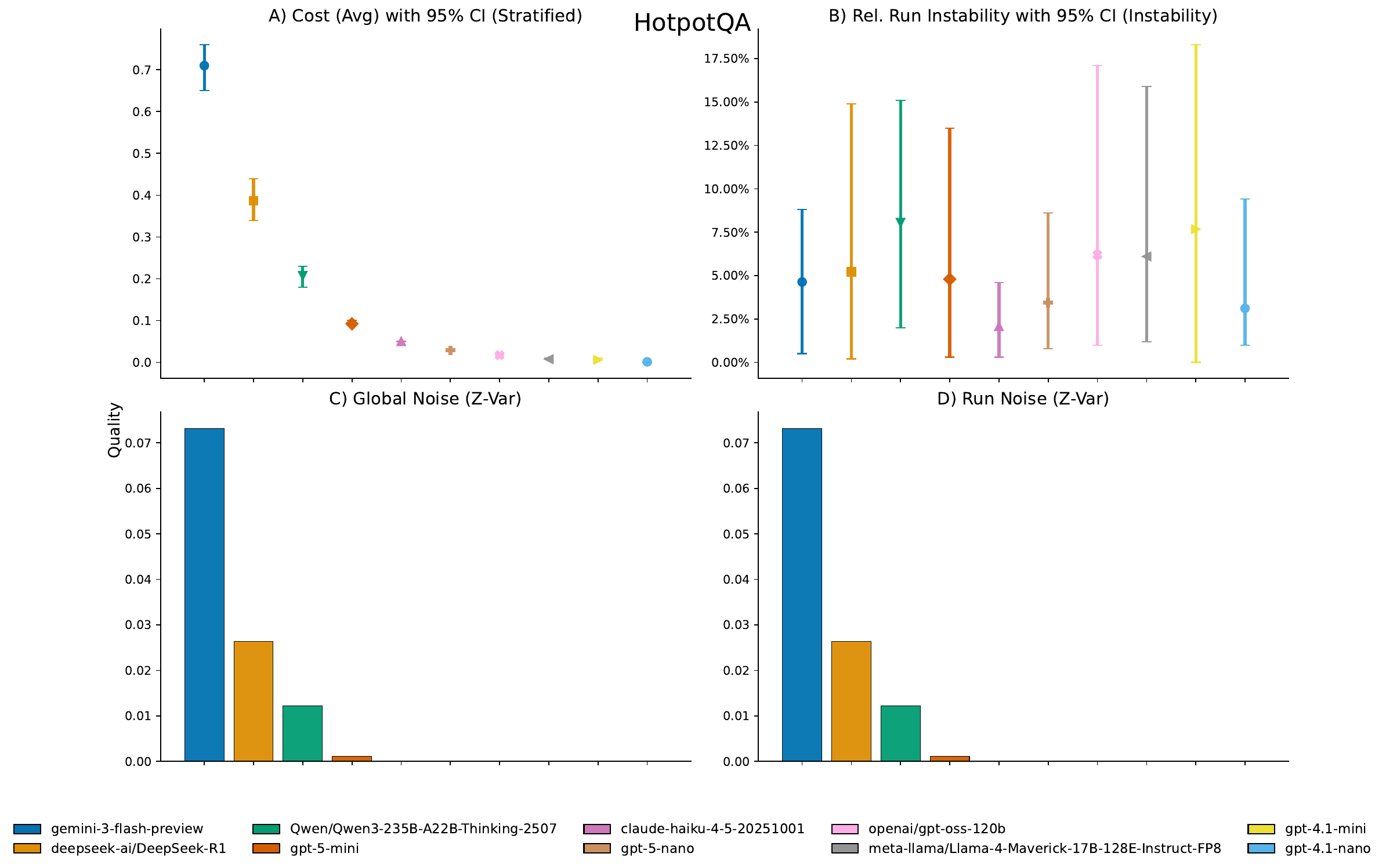}
    \caption{Cost distributions on HotpotQA.}
    \label{fig:model_cost_hotpotqa}
\end{figure}

\begin{figure}[tbp]
    \centering
    \includegraphics[width=0.96\linewidth]{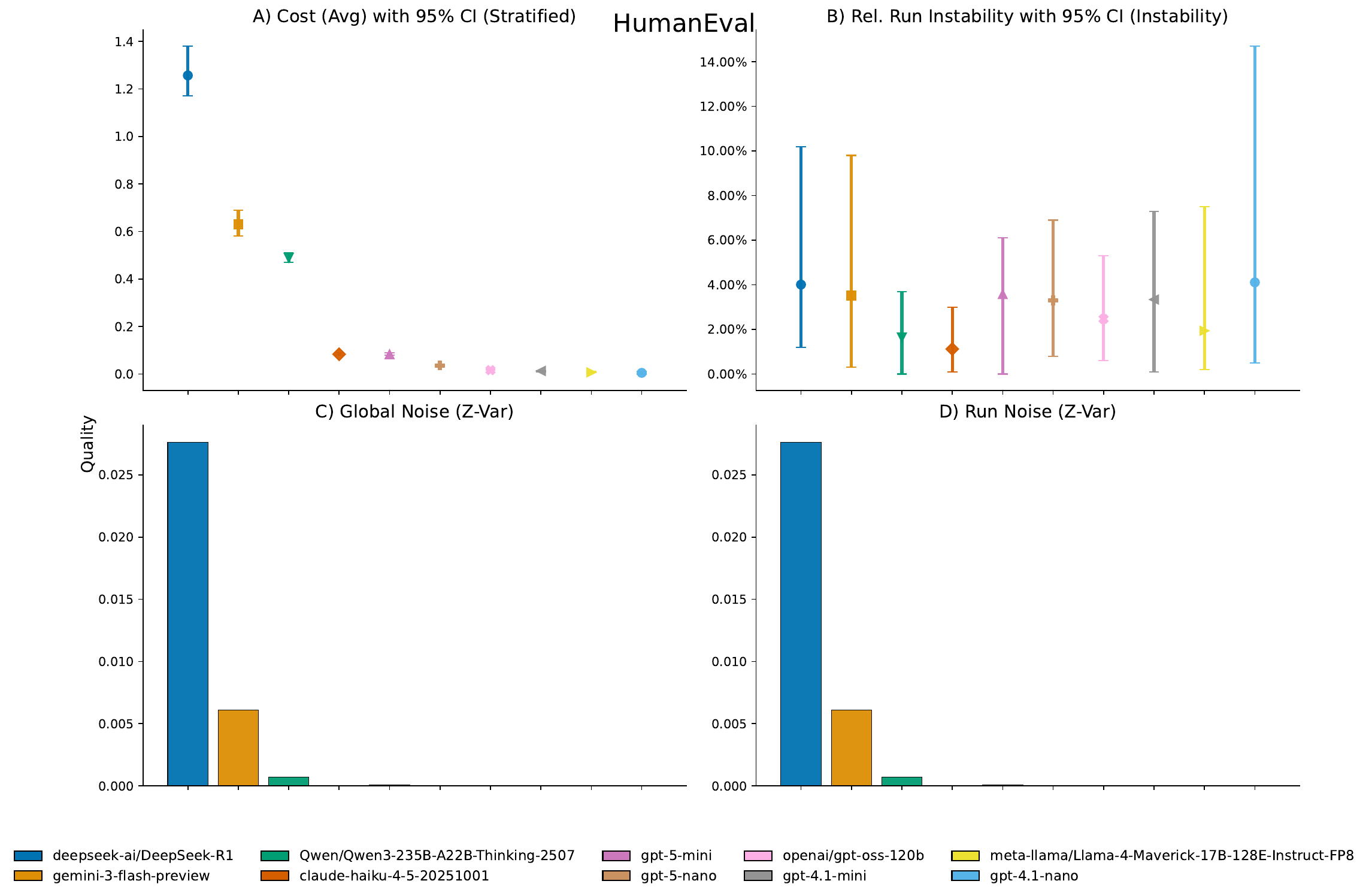}
    \caption{Cost distributions on HumanEval.}
    \label{fig:model_cost_humaneval}
\end{figure}

\begin{figure}[tbp]
    \centering
    \includegraphics[width=0.96\linewidth]{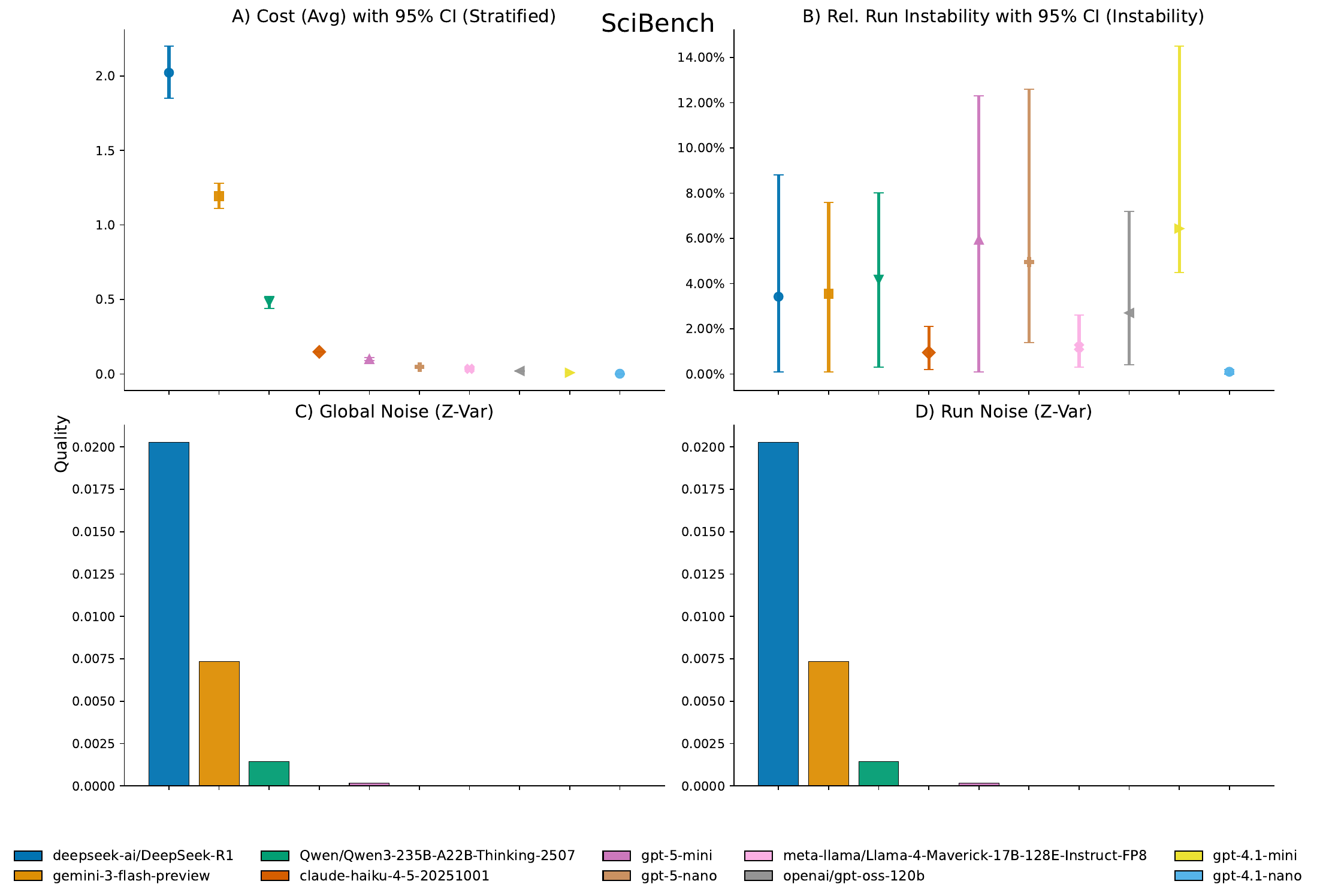}
    \caption{Cost distributions on SciBench.}
    \label{fig:model_cost_scibench}
\end{figure}

\begin{figure}[tbp]
    \centering
    \includegraphics[width=0.96\linewidth]{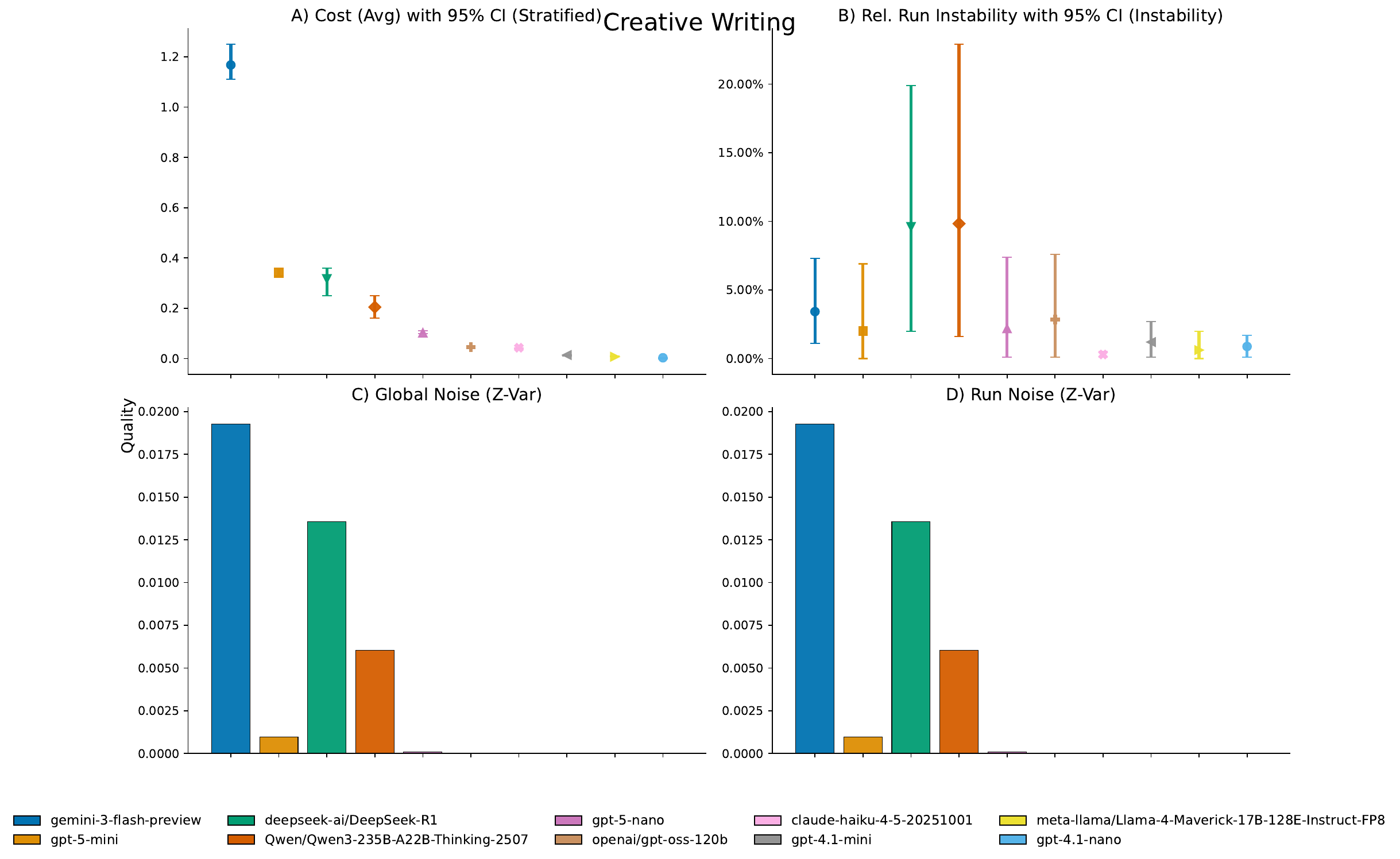}
    \caption{Cost distributions on Shakespearean Sonnet Writing.}
    \label{fig:model_cost_sonnetwriting}
\end{figure}

\subsection{Run-level Diagnosis Traces}
\label{appendix:diagnosis}
For the models with diagnosis traces, we plot per-run outcomes grouped by benchmark and report benchmark-normalized z-scores together with absolute and relative deviations from the benchmark mean. These traces visualize the run-level distributions that underlie \Tabref{tab:models_results}, and together with the noise-taxonomy plane (\Figref{fig:noise_taxonomy}) provide a per-model lens on the stability quadrants. Conventions are identical across panels; captions name only the model.

\begin{figure}[tbp]
    \centering
    \includegraphics[width=0.96\linewidth]{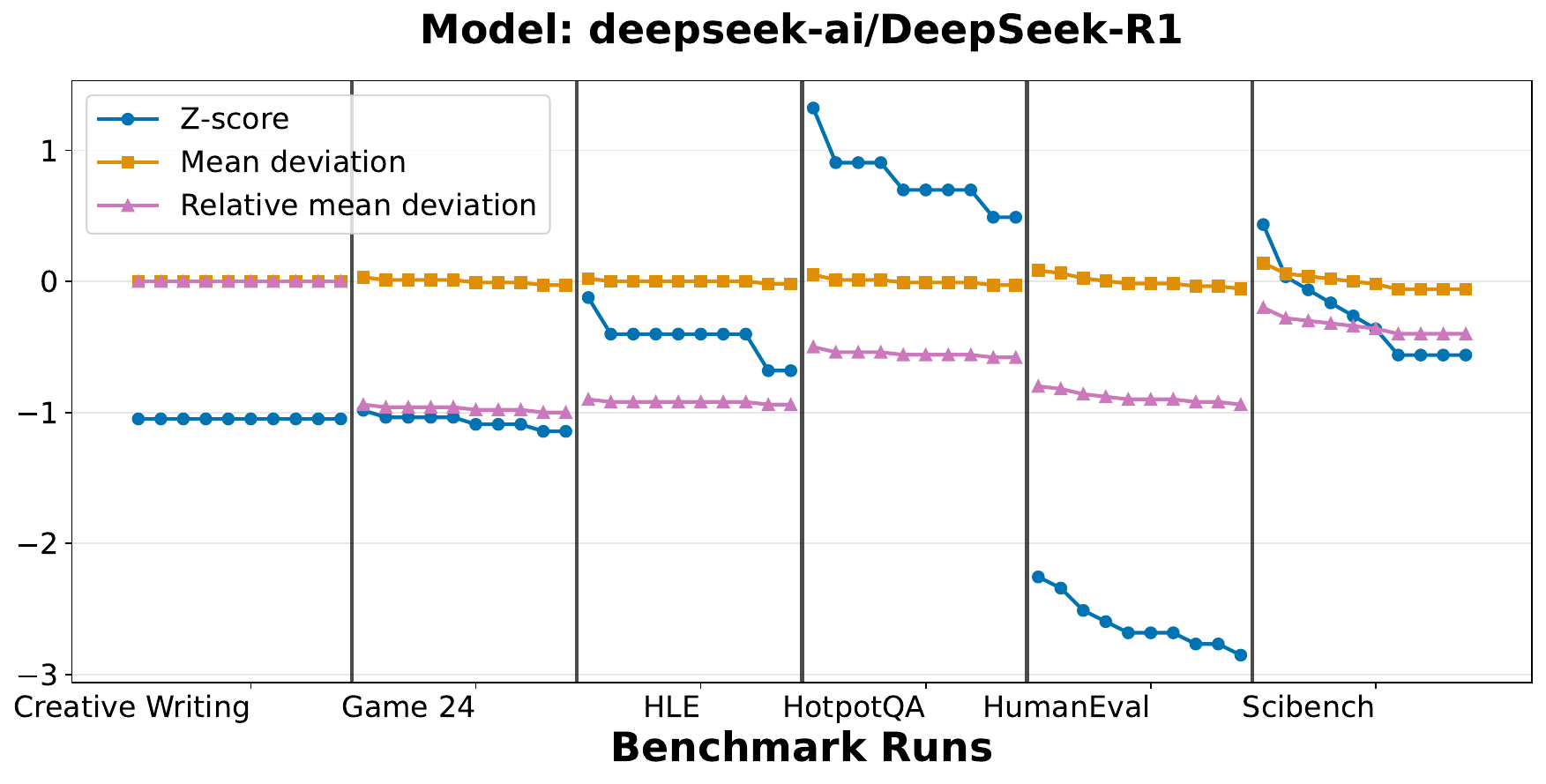}
    \caption{Stability trace, \textsc{DeepSeek R1}.}
    \label{fig:diagnosis_deepseek_r1}
\end{figure}

\begin{figure}[tbp]
    \centering
    \includegraphics[width=0.96\linewidth]{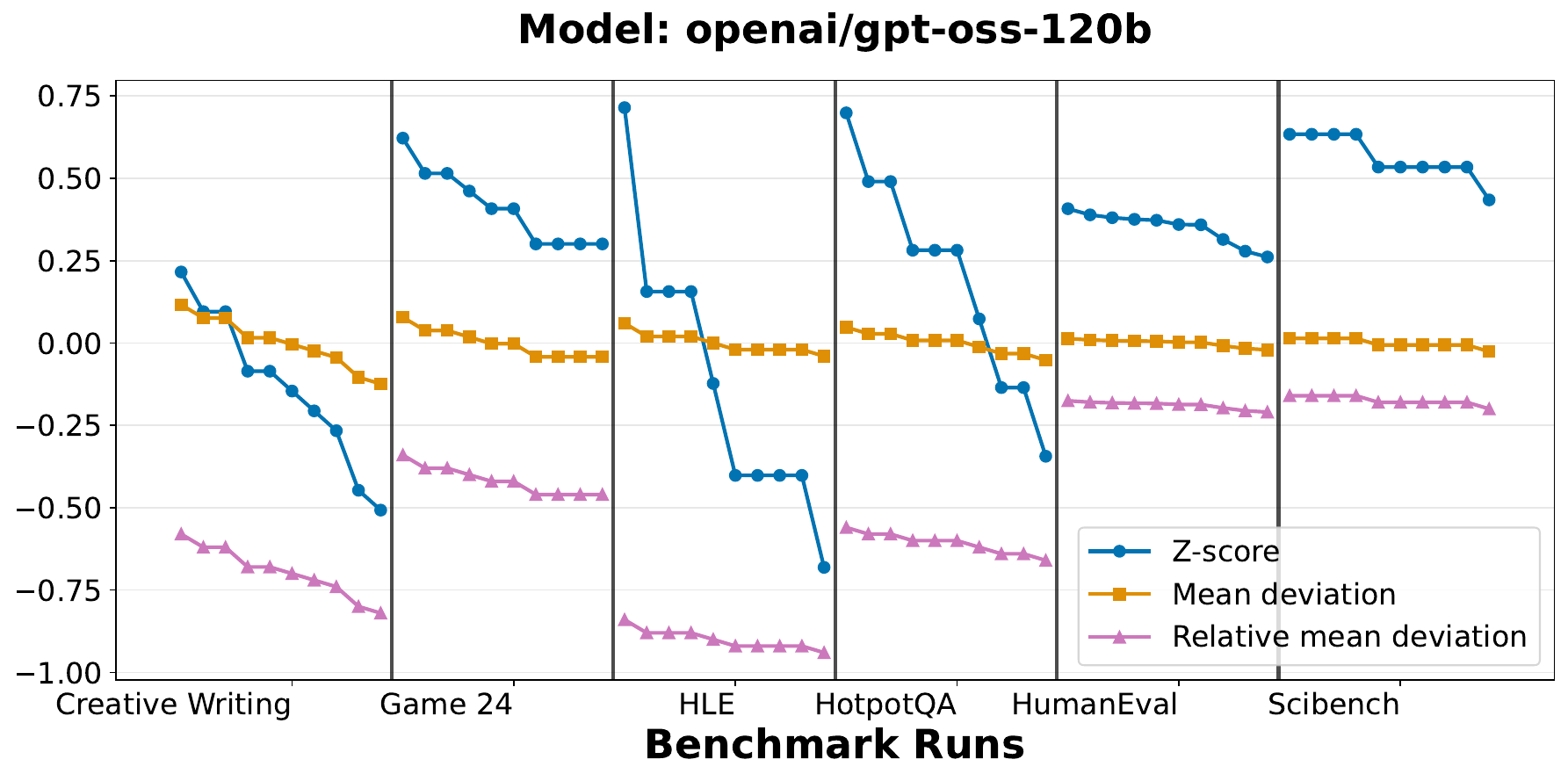}
    \caption{Stability trace, \textsc{GPT-OSS 120B}.}
    \label{fig:diagnosis_gpt_oss_120b}
\end{figure}

\begin{figure}[tbp]
    \centering
    \includegraphics[width=0.96\linewidth]{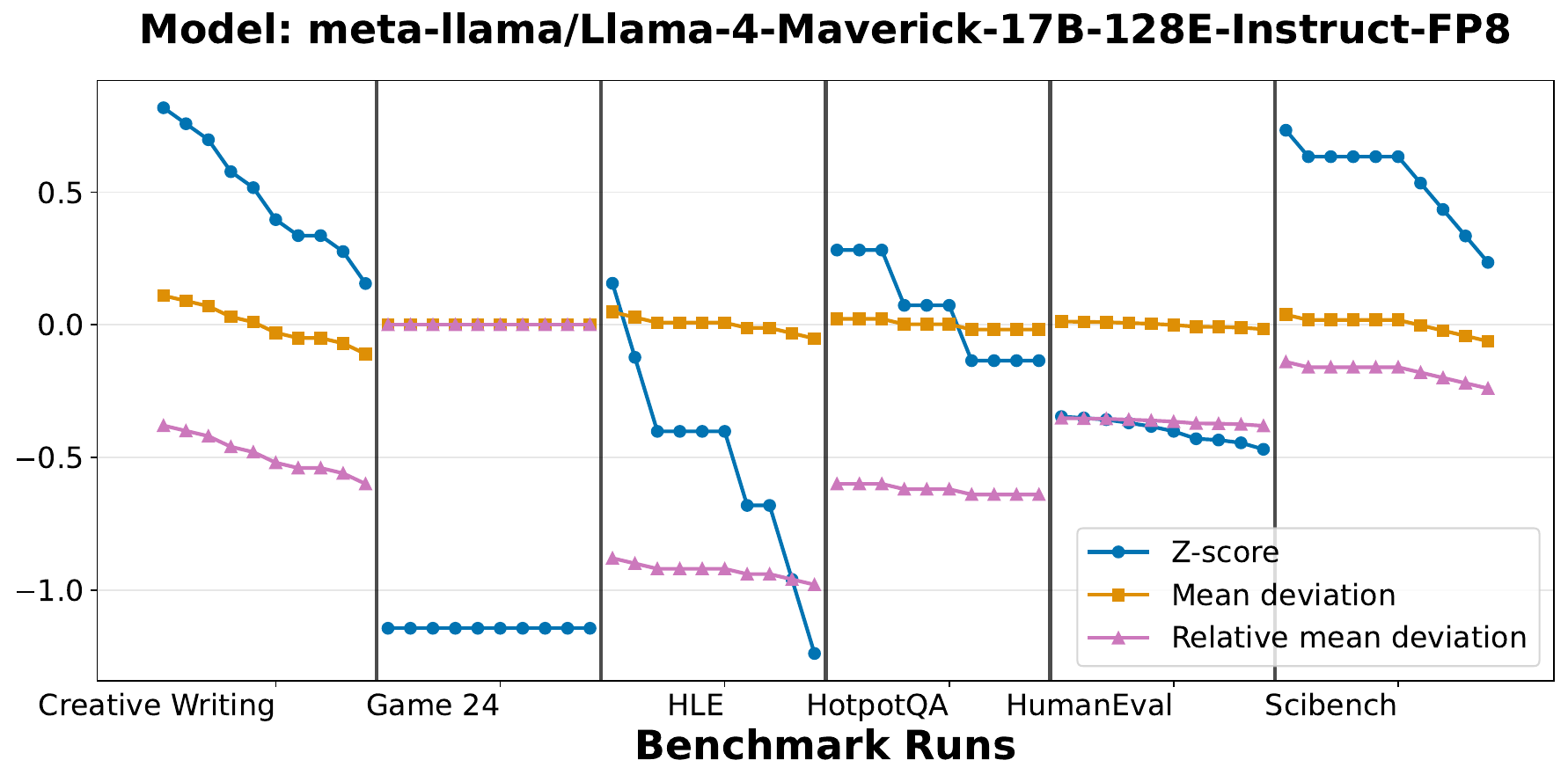}
    \caption{Stability trace, \textsc{Llama-4 Maverick}.}
    \label{fig:diagnosis_llama4_maverick}
\end{figure}

\begin{figure}[tbp]
    \centering
    \includegraphics[width=0.96\linewidth]{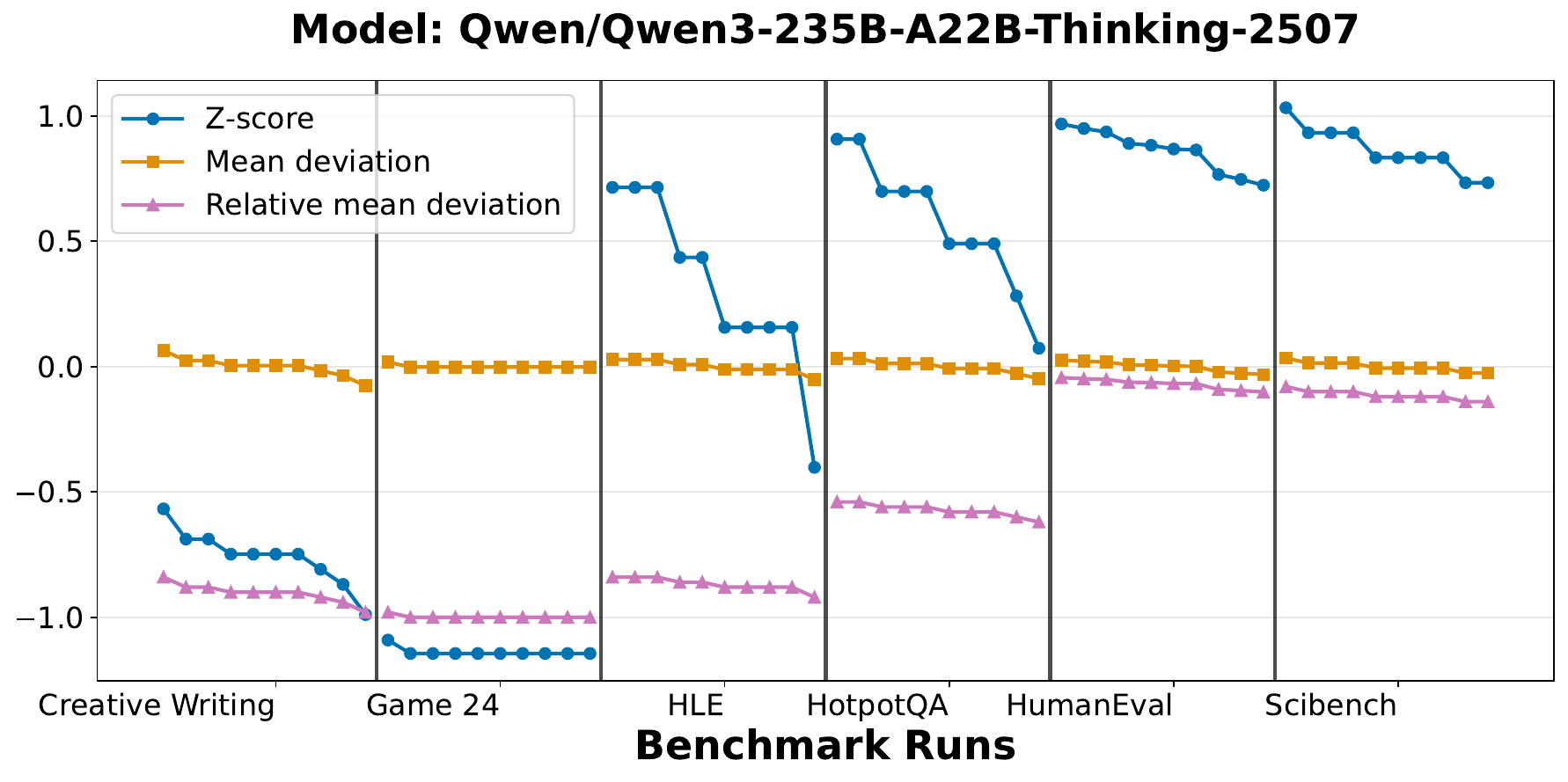}
    \caption{Stability trace, \textsc{Qwen3-235B Thinking}.}
    \label{fig:diagnosis_qwen3_235b}
\end{figure}

\begin{figure}[tbp]
    \centering
    \includegraphics[width=0.96\linewidth]{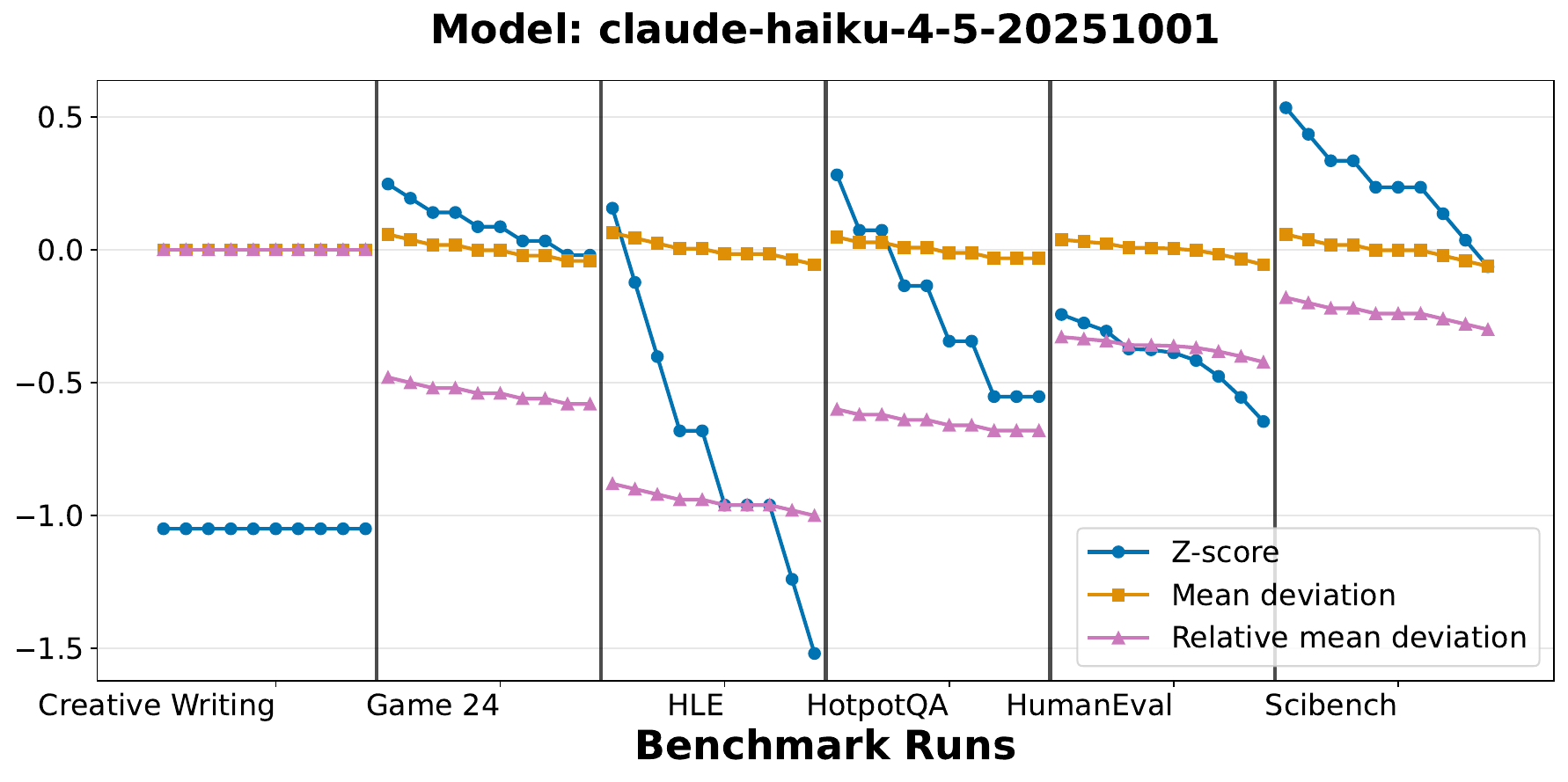}
    \caption{Stability trace, \textsc{Claude Haiku 4.5}.}
    \label{fig:diagnosis_claude_haiku_4_5}
\end{figure}

\begin{figure}[tbp]
    \centering
    \includegraphics[width=0.96\linewidth]{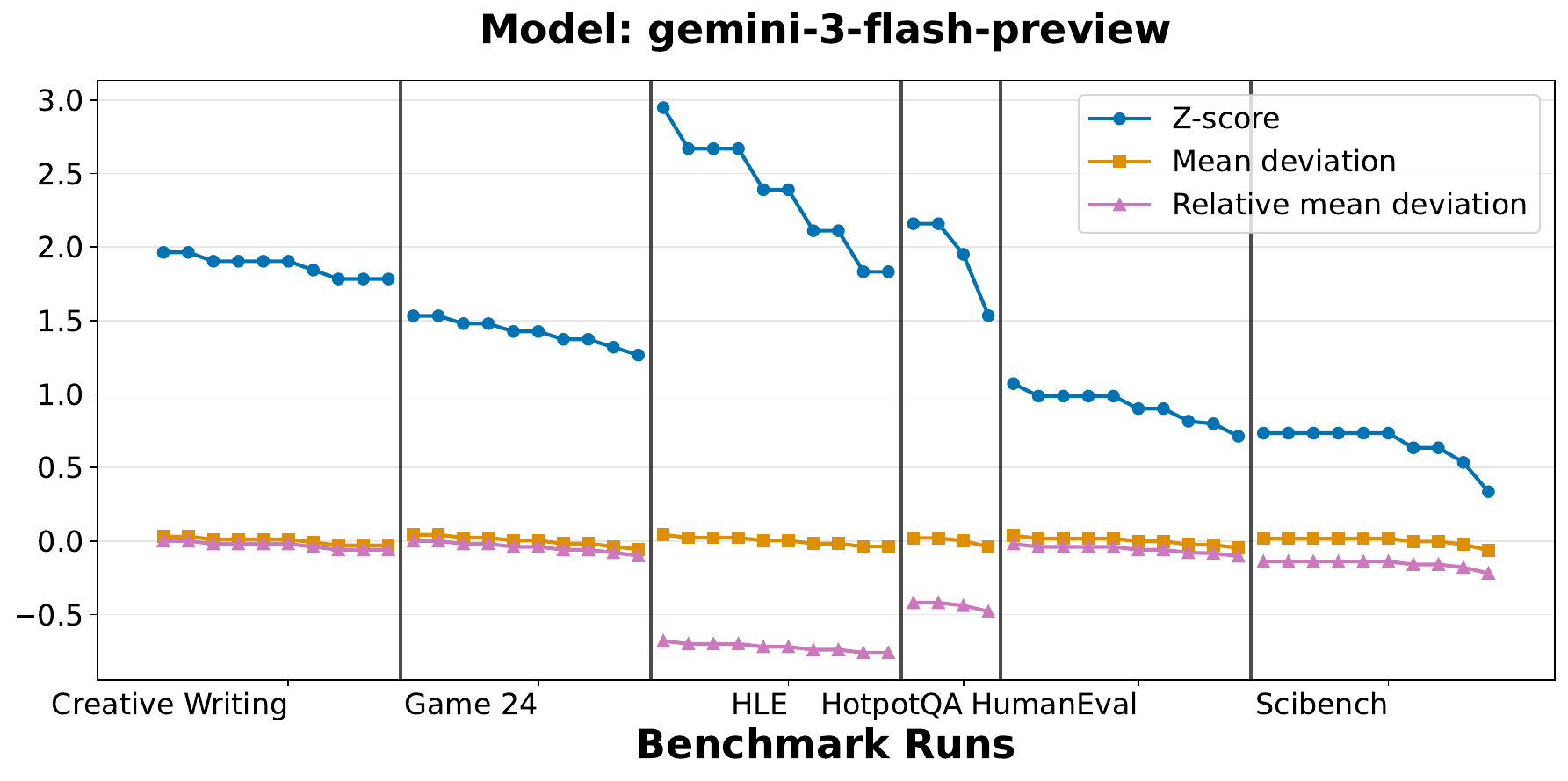}
    \caption{Stability trace, \textsc{Gemini-3 Flash}.}
    \label{fig:diagnosis_gemini_3_flash}
\end{figure}

\begin{figure}[tbp]
    \centering
    \includegraphics[width=0.96\linewidth]{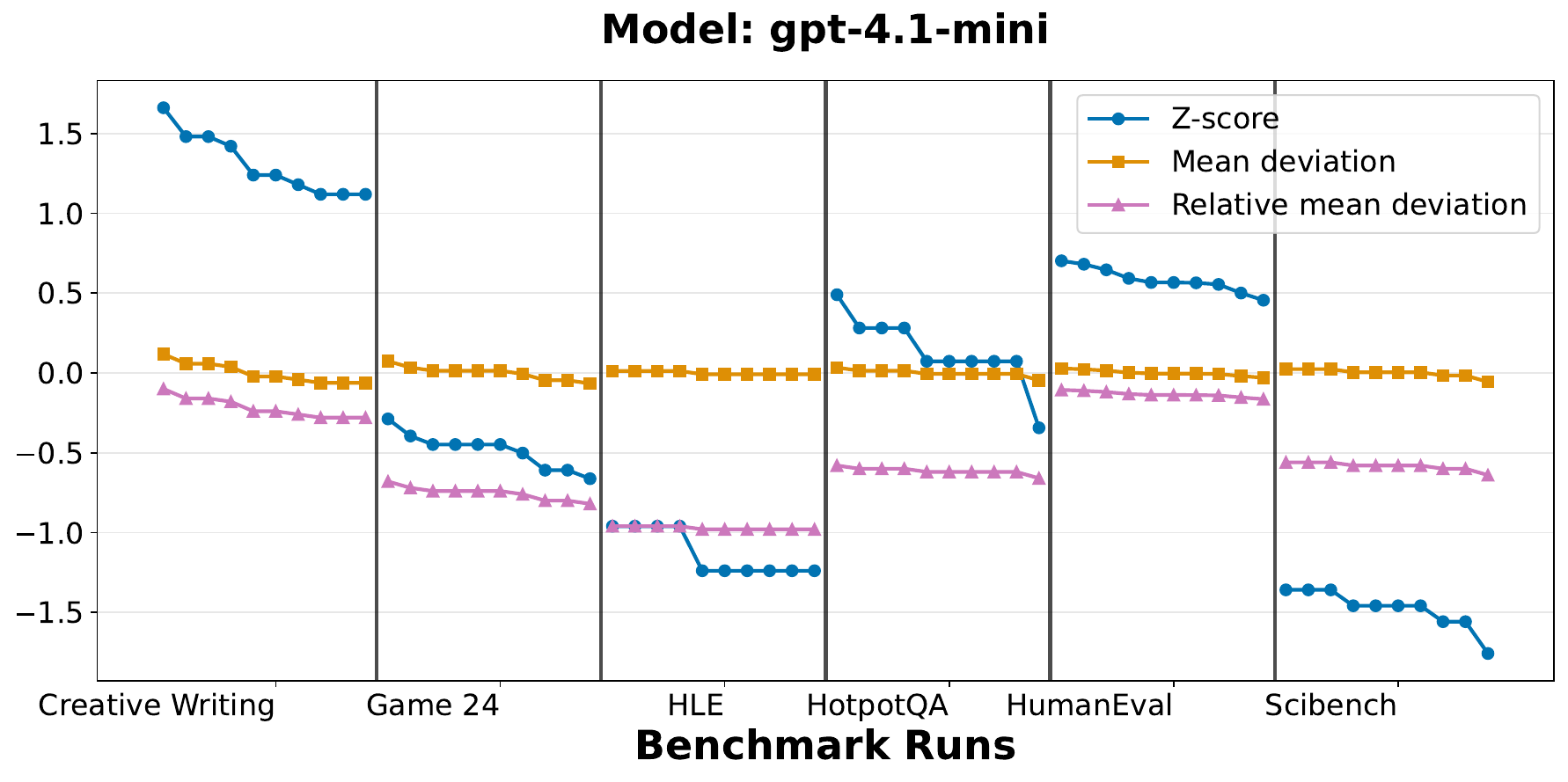}
    \caption{Stability trace, \textsc{GPT-4.1 Mini} (non-reasoning baseline).}
    \label{fig:diagnosis_gpt_4_1_mini}
\end{figure}

\begin{figure}[tbp]
    \centering
    \includegraphics[width=0.96\linewidth]{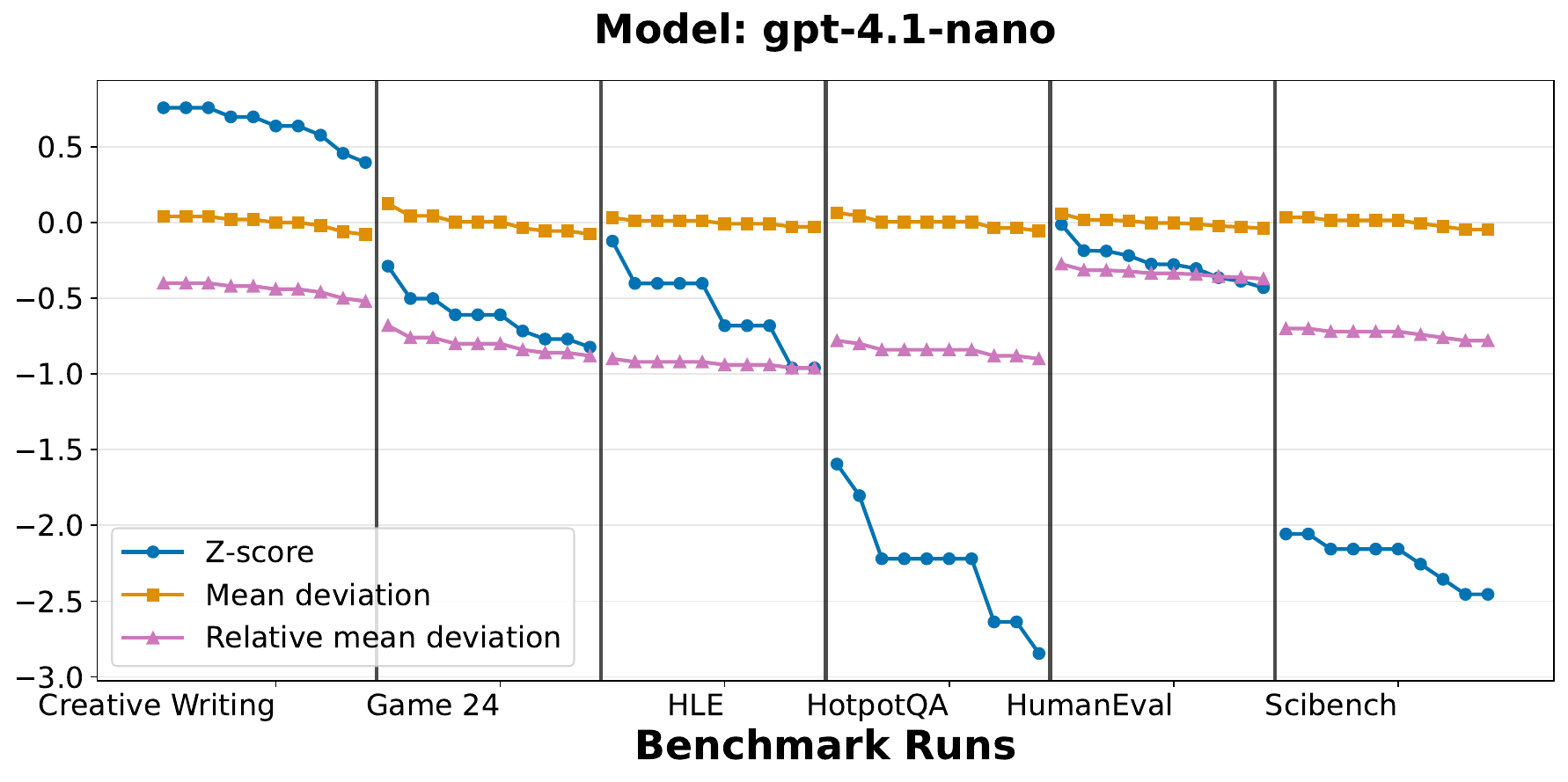}
    \caption{Stability trace, \textsc{GPT-4.1 Nano} (non-reasoning baseline).}
    \label{fig:diagnosis_gpt_4_1_nano}
\end{figure}

\begin{figure}[tbp]
    \centering
    \includegraphics[width=0.96\linewidth]{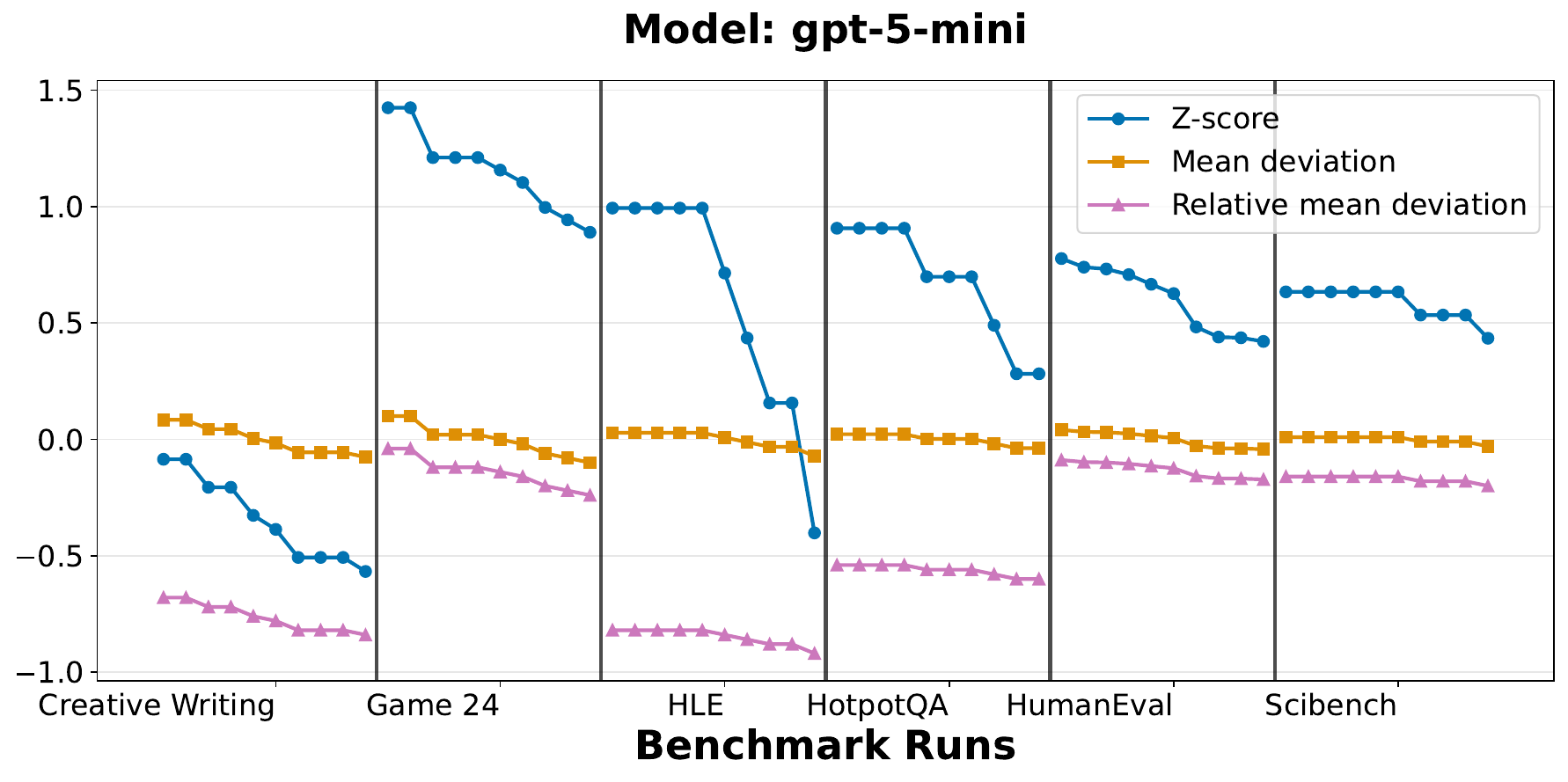}
    \caption{Stability trace, \textsc{GPT-5 Mini}.}
    \label{fig:diagnosis_gpt_5_mini}
\end{figure}

\begin{figure}[tbp]
    \centering
    \includegraphics[width=0.96\linewidth]{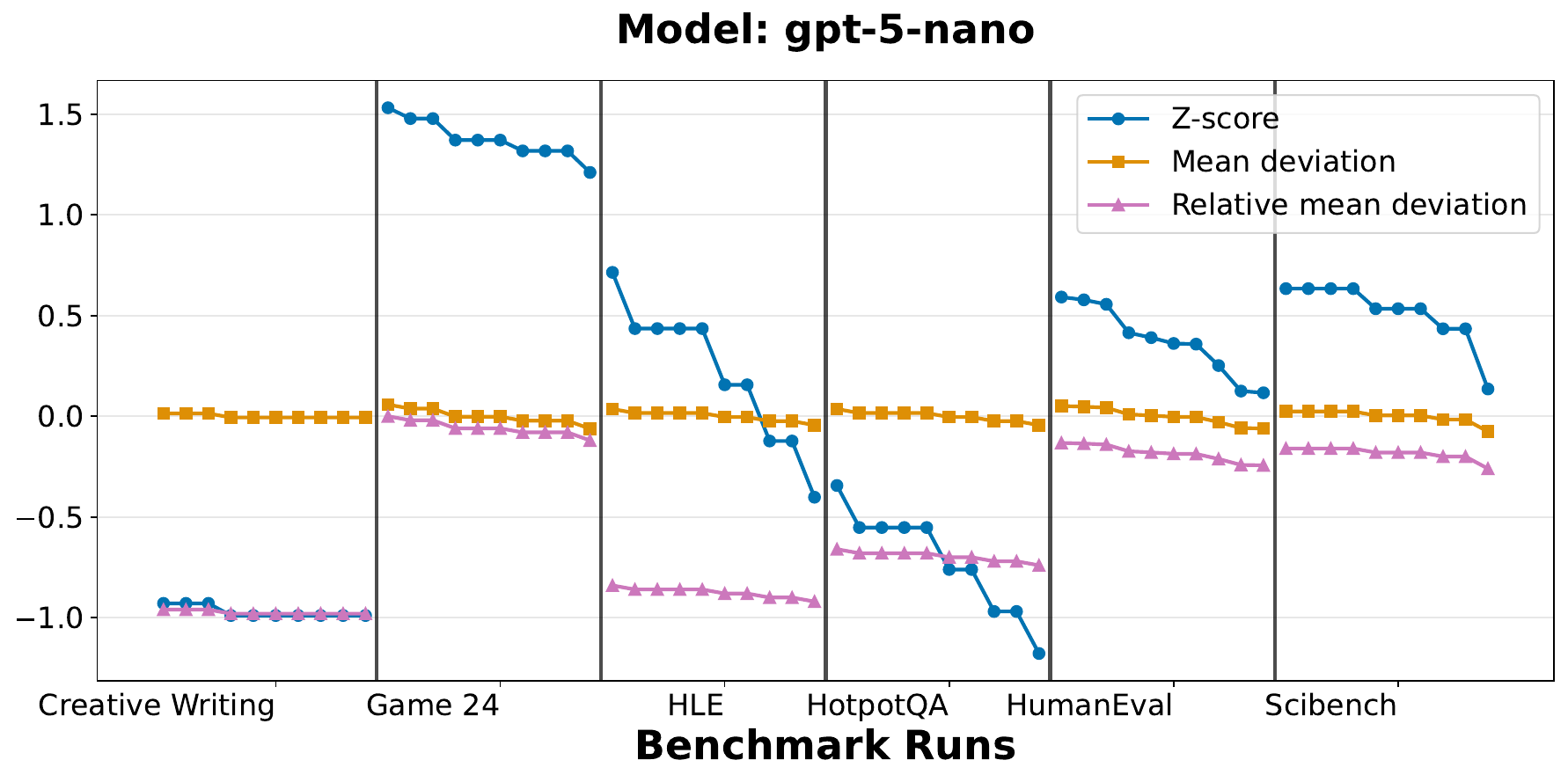}
    \caption{Stability trace, \textsc{GPT-5 Nano}.}
    \label{fig:diagnosis_gpt_5_nano}
\end{figure}

\subsection{Cross-Axis Details}
\label{appendix:cross_axis}

This subsection collects the numerical details and figures deferred from \Cref{sec:cross_axis}. ``Model axis'' refers to the experiment that varies models with a fixed strategy; ``strategy axis'' refers to the experiment that varies strategies with a fixed model (\textsc{GPT-4.1-nano}).

\xhdr{Range asymmetry}
On 30 runs, suite-level Score spans $0.628$ across models versus $0.360$ across strategies (a factor of $1.7\times$ wider on the model axis), while Rel~Error spans $0.587$ vs $0.315$ ($1.9\times$ wider on the strategy axis) and Global~Noise spans $1.414$ vs $0.917$ ($1.5\times$). Run~Noise spans $0.029$ vs $0.038$ ($1.3\times$).

\xhdr{Game24 character flip}
Game24 is the most discriminating benchmark on the model axis (score std $0.402$, range $0$--$0.959$) and the least on the strategy axis (std $0.070$, range $0.004$--$0.213$), a five-position rank shift in benchmark discrimination. Run~Noise is comparable in magnitude across axes (max $0.064$ on models, $0.052$ on strategies); the SNR ratio is $14.0$ on the model axis vs $2.8$ on the strategy axis ($5\times$ worse for strategies).

\begin{figure}[tbp]
\centering
\includegraphics[width=0.95\linewidth]{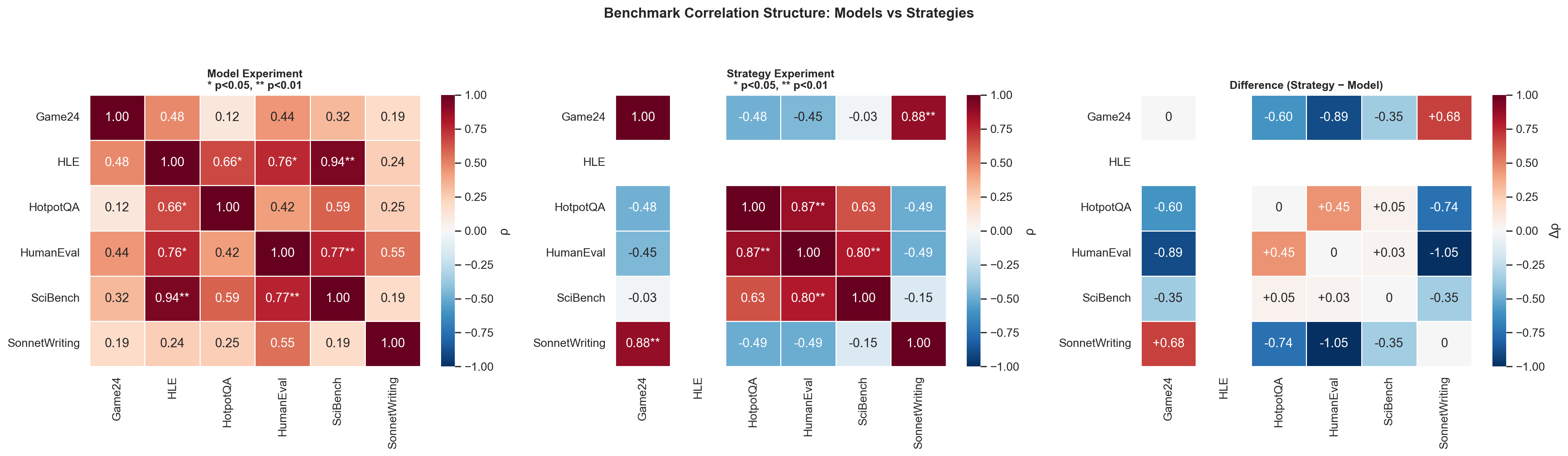}
\caption{\textbf{Pairwise benchmark correlation structures across axes.} Spearman $\rho$ between benchmark scores, computed across the model panel (left) and across the strategy panel (right).}
\label{fig:appx_cross_axis_corr}
\end{figure}

\xhdr{Benchmark correlation structures}
\Cref{fig:appx_cross_axis_corr} shows the two correlation matrices. On the model axis, the dominant pairs are HLE$\leftrightarrow$SciBench ($\rho{=}{+}0.94$), HumanEval$\leftrightarrow$SciBench (${+}0.77$), and HLE$\leftrightarrow$HumanEval (${+}0.76$): models cluster benchmarks by content domain. On the strategy axis, the dominant pairs are Game24$\leftrightarrow$SonnetWriting (${+}0.88$) and HotpotQA$\leftrightarrow$HumanEval (${+}0.87$): strategies cluster benchmarks by reasoning mode. Several pairs reverse sign across axes -- HumanEval$\leftrightarrow$SonnetWriting (${+}0.55 \to {-}0.49$), HotpotQA$\leftrightarrow$SonnetWriting (${+}0.25 \to {-}0.49$), Game24$\leftrightarrow$HotpotQA (${+}0.12 \to {-}0.48$).

\xhdr{Zero-score collapse rates}
On the model axis, $9$ of $60$ cells collapse ($15\%$): $3$ at exactly zero, $6$ near-zero ($<0.05$). On the strategy axis, $14$ of $53$ cells collapse ($26\%$): $5$ at zero, $9$ near-zero. Strategy-axis collapses concentrate on \totdfs{} (HumanEval, HotpotQA, SciBench), where peer strategies on the same underlying model reach $0.05$--$0.47$, indicating format mismatch rather than capability ceiling. Model-axis collapses concentrate on SonnetWriting (\eg \textsc{DeepSeek R1} scoring $0$ across all $10$ runs).

\begin{figure}[tbp]
\centering
\includegraphics[width=0.85\linewidth]{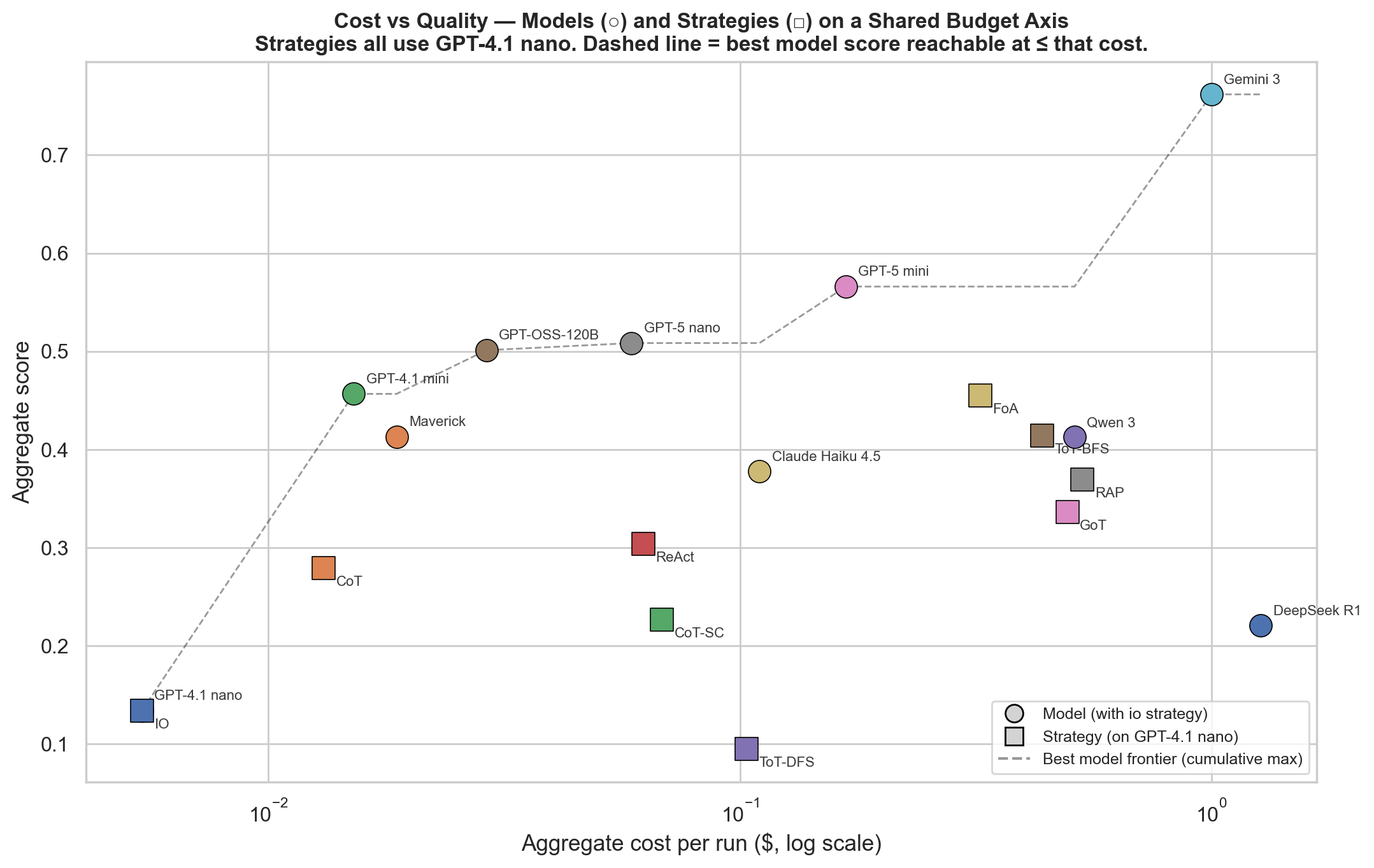}
\caption{\textbf{Mean cost vs mean quality on both axes.} Each marker is one suite-level mean. Models (blue) and strategies on \textsc{GPT-4.1-nano} (orange) share the baseline configuration $(\textsc{nano}, \textsc{io})$, so costs are directly comparable. The dashed line is the model frontier.}
\label{fig:appx_cross_axis_cost}
\end{figure}

\xhdr{Suite-level cost vs quality}
\Cref{fig:appx_cross_axis_cost} plots the joint cost-quality scatter. Cost spans $\$0.005$--$\$1.27$ across models ($236\times$) and $\$0.005$--$\$0.53$ across strategies on \textsc{GPT-4.1-nano} ($99\times$). \textsc{DeepSeek R1} is the most expensive model and the second-lowest in suite quality ($0.221$), while \totdfs{} ($\$0.103$) is more expensive than both \io{} ($\$0.005$) and \cot{} ($\$0.013$) yet scores below them. The dashed line is the \emph{model frontier} -- the highest model score reachable at or below each cost. Above ${\sim}\$0.01$ per run, the frontier dominates every strategy mean; only \cot{} (cost $\$0.013$, score $0.279$) sits above it, and only because no model in the panel is cheaper than \textsc{nano} (cost $\$0.005$, score $0.134$).

\xhdr{Within-cell cost--score correlations}
The most negative within-cell Spearman correlations between per-run cost and per-run score are, on the model axis: \textsc{GPT-5 Nano}$\,\times\,$Game24 $\rho{=}{-}0.76$ ($p{<}0.001$, $30$ runs); \textsc{DeepSeek R1}$\,\times\,$HotpotQA ${-}0.61$ ($p{=}0.061$, $10$ runs); \textsc{Llama-4 Maverick}$\,\times\,$HumanEval ${-}0.54$ ($p{=}0.014$, $20$ runs). On the strategy axis: \textsc{\totbfs{}}$\,\times\,$SonnetWriting ${-}0.91$, \rap{}$\,\times\,$HumanEval ${-}0.90$, \react{}$\,\times\,$HLE ${-}0.83$, \react{}$\,\times\,$HotpotQA ${-}0.70$, \react{}$\,\times\,$SciBench ${-}0.67$ (all $p{<}0.05$, $30$ runs). The strategy-axis correlations are stronger and more numerous, but the qualitative finding -- that on a substantial fraction of cells, more cost predicts \emph{worse} quality within a single (entity, benchmark) pair -- holds on both axes.

\xhdr{Joint cost-quality failure rates and threshold sensitivity}
\Tabref{tab:appx_joint_strategies_median} and \Tabref{tab:appx_joint_models_median} report suite-aggregated joint-failure rates under the cross-entity median quality/cost threshold. Rates reach $81.7\%$ for \textsc{DeepSeek R1}, $60.0\%$ for \textsc{Claude-Haiku 4.5}, and $45.0\%$ for \textsc{GPT-5 Nano} on the model axis, and $43.3\%$ for \got{}, $41.3\%$ for \rap{}, and $25.0\%$ for \totdfs{} on the strategy axis. At the opposite extreme, \textsc{Gemini-3 Flash} fails the cost threshold on every run yet rarely falls below the quality threshold (joint failure $1.7\%$), while \io{} and \cot{} are cheap enough that they never trip the cost threshold (joint failure $0\%$). \Tabref{tab:appx_joint_strategies_hard}, \Tabref{tab:appx_joint_models_hard}, \Tabref{tab:appx_joint_strategies_2sd}, and \Tabref{tab:appx_joint_models_2sd} repeat the analysis under two stricter threshold criteria (cross-entity hard reference values, and per-pair mean $\pm 2\,$SD); the qualitative pattern, including the cheap-immune / expensive-exposed asymmetry, is preserved across all three threshold criteria.

\begin{table*}[tbp]
\centering
\caption{\textbf{Joint Failure Rates: Strategies, Median Threshold.} Per benchmark, quality fails if below the pooled median; cost fails if above the pooled median. Joint Fail = fraction of runs where both quality and cost fail simultaneously. Rank 1 = lowest failure rate (best); ties resolved alphabetically.}
\label{tab:appx_joint_strategies_median}
\scriptsize
\resizebox{0.92\textwidth}{!}{%
\begin{tabular}{lcccccc}
\toprule
& \multicolumn{2}{c}{\textbf{Quality}} & \multicolumn{2}{c}{\textbf{Cost}} & \multicolumn{2}{c}{\textbf{Joint}} \\
\cmidrule(lr){2-3} \cmidrule(lr){4-5} \cmidrule(lr){6-7}
Strategy & Fail (\%) & Rank & Fail (\%) & Rank & Fail (\%) & Rank \\
\midrule
\io{}       & 83.9 & 9 & 0.0  & 1 & 0.0  & 1 \\
\cot{}      & 35.0 & 3 & 0.0  & 2 & 0.0  & 2 \\
\cotsc{}   & 58.9 & 7 & 20.6 & 3 & 11.1 & 3 \\
\react{}    & 45.6 & 5 & 37.8 & 4 & 22.2 & 5 \\
\totdfs{}  & 66.7 & 8 & 50.0 & 5 & 25.0 & 7 \\
\totbfs{}  & 23.3 & 2 & 98.9 & 9 & 23.3 & 6 \\
\got{}      & 50.0 & 6 & 82.2 & 6 & 43.3 & 9 \\
\rap{}      & 41.3 & 4 & 82.7 & 7 & 41.3 & 8 \\
\foa{}      & 12.8 & 1 & 83.3 & 8 & 12.8 & 4 \\
\bottomrule
\end{tabular}%
}
\end{table*}

\begin{table*}[tbp]
\centering
\caption{\textbf{Joint Failure Rates: Models, Median Threshold.} Per benchmark, quality fails if below the pooled median; cost fails if above the pooled median. Joint Fail = fraction of runs where both quality and cost fail simultaneously. Rank 1 = lowest failure rate (best).}
\label{tab:appx_joint_models_median}
\scriptsize
\resizebox{0.92\textwidth}{!}{%
\begin{tabular}{lcccccc}
\toprule
& \multicolumn{2}{c}{\textbf{Quality}} & \multicolumn{2}{c}{\textbf{Cost}} & \multicolumn{2}{c}{\textbf{Joint}} \\
\cmidrule(lr){2-3} \cmidrule(lr){4-5} \cmidrule(lr){6-7}
Model & Fail (\%) & Rank & Fail (\%) & Rank & Fail (\%) & Rank \\
\midrule
DeepSeek R1       & 81.7  & 9  & 100.0 & 7 & 81.7 & 10 \\
Claude Haiku 4.5  & 76.7  & 8  & 83.3  & 5 & 60.0 & 9 \\
GPT-4.1 Nano      & 100.0 & 10 & 0.0   & 1 & 0.0  & 1 \\
Qwen3 235B        & 35.0  & 4  & 100.0 & 7 & 35.0 & 7 \\
Llama-4 Maverick  & 53.3  & 6  & 0.0   & 1 & 0.0  & 1 \\
GPT-4.1 Mini      & 56.1  & 7  & 2.8   & 3 & 2.8  & 5 \\
GPT-OSS 120B      & 27.8  & 3  & 4.4   & 4 & 0.5  & 4 \\
GPT-5 Nano        & 49.4  & 5  & 76.1  & 4 & 45.0 & 8 \\
GPT-5 Mini        & 13.9  & 2  & 100.0 & 7 & 13.9 & 6 \\
Gemini 3 Flash    & 1.7   & 1  & 100.0 & 7 & 1.7  & 3 \\
\bottomrule
\end{tabular}%
}
\end{table*}

\begin{table*}[tbp]
\centering
\caption{\textbf{Joint Failure Rates: Strategies, Hard Threshold.} Per benchmark, quality fails if below the mean of the weakest strategy (\io{}); cost fails if above the mean of the most expensive strategy (\rap{}). Joint Fail = fraction of runs where both quality and cost fail simultaneously. Rank 1 = lowest failure rate (best); ties resolved alphabetically.}
\label{tab:appx_joint_strategies_hard}
\scriptsize
\resizebox{0.92\textwidth}{!}{%
\begin{tabular}{lcccccc}
\toprule
& \multicolumn{2}{c}{\textbf{Quality}} & \multicolumn{2}{c}{\textbf{Cost}} & \multicolumn{2}{c}{\textbf{Joint}} \\
\cmidrule(lr){2-3} \cmidrule(lr){4-5} \cmidrule(lr){6-7}
Strategy & Fail (\%) & Rank & Fail (\%) & Rank & Fail (\%) & Rank \\
\midrule
\io{}       & 56.7 & 8 & 0.0  & 1 & 0.0  & 1 \\
\cot{}      & 21.3 & 4 & 0.0  & 1 & 0.0  & 1 \\
\cotsc{}   & 31.3 & 5 & 20.0 & 5 & 0.0  & 1 \\
\react{}    & 40.0 & 6 & 0.0  & 1 & 0.0  & 1 \\
\totdfs{}  & 60.0 & 9 & 20.0 & 5 & 20.0 & 9 \\
\totbfs{}  & 0.7  & 2 & 42.0 & 7 & 0.0  & 1 \\
\got{}      & 1.3  & 3 & 60.0 & 9 & 0.0  & 1 \\
\rap{}      & 40.0 & 6 & 47.3 & 8 & 19.3 & 8 \\
\foa{}      & 0.0  & 1 & 40.0 & 6 & 0.0  & 1 \\
\bottomrule
\end{tabular}%
}
\end{table*}

\begin{table*}[tbp]
\centering
\caption{\textbf{Joint Failure Rates: Models, Hard Threshold.} Per benchmark, quality fails if below the mean of GPT-4.1 Nano (weakest); cost fails if above the mean of DeepSeek R1 (most expensive). Joint Fail = fraction of runs where both quality and cost fail simultaneously. Rank 1 = lowest failure rate (best); ties resolved alphabetically.}
\label{tab:appx_joint_models_hard}
\scriptsize
\resizebox{0.92\textwidth}{!}{%
\begin{tabular}{lcccccc}
\toprule
& \multicolumn{2}{c}{\textbf{Quality}} & \multicolumn{2}{c}{\textbf{Cost}} & \multicolumn{2}{c}{\textbf{Joint}} \\
\cmidrule(lr){2-3} \cmidrule(lr){4-5} \cmidrule(lr){6-7}
Model & Fail (\%) & Rank & Fail (\%) & Rank & Fail (\%) & Rank \\
\midrule
DeepSeek R1       & 49.4 & 9  & 45.9 & 10 & 20.3 & 10 \\
Claude Haiku 4.5  & 20.0 & 7  & 0.0  & 1  & 0.0  & 1 \\
GPT-4.1 Nano      & 63.3 & 10 & 0.0  & 1  & 0.0  & 1 \\
Qwen3 235B        & 31.7 & 8  & 0.0  & 1  & 0.0  & 1 \\
Llama-4 Maverick  & 16.7 & 5  & 0.0  & 1  & 0.0  & 1 \\
GPT-4.1 Mini      & 10.5 & 4  & 0.0  & 1  & 0.0  & 1 \\
GPT-OSS 120B      & 0.5  & 2  & 0.0  & 1  & 0.0  & 1 \\
GPT-5 Nano        & 16.7 & 5  & 0.0  & 1  & 0.0  & 1 \\
GPT-5 Mini        & 1.1  & 3  & 16.7 & 8  & 1.1  & 9 \\
Gemini 3 Flash    & 0.0  & 1  & 33.3 & 9  & 0.0  & 1 \\
\bottomrule
\end{tabular}%
}
\end{table*}

\begin{table*}[tbp]
\centering
\caption{\textbf{Joint Failure Rates: Strategies, Per-pair Mean $\pm 2\,$SD Threshold.} Per (strategy, benchmark) pair, quality fails if below its own mean $-\,2\,$SD; cost fails if above its own mean $+\,2\,$SD. Joint Fail = fraction of runs where both quality and cost fail simultaneously. Rank 1 = lowest failure rate (best); ties resolved alphabetically.}
\label{tab:appx_joint_strategies_2sd}
\scriptsize
\resizebox{0.92\textwidth}{!}{%
\begin{tabular}{lcccccc}
\toprule
& \multicolumn{2}{c}{\textbf{Quality}} & \multicolumn{2}{c}{\textbf{Cost}} & \multicolumn{2}{c}{\textbf{Joint}} \\
\cmidrule(lr){2-3} \cmidrule(lr){4-5} \cmidrule(lr){6-7}
Strategy & Fail (\%) & Rank & Fail (\%) & Rank & Fail (\%) & Rank \\
\midrule
\io{}       & 4.4  & 8 & 2.8 & 8 & 0.0 & 1 \\
\cot{}      & 1.1  & 2 & 2.2 & 5 & 0.0 & 1 \\
\cotsc{}   & 1.6  & 5 & 2.2 & 5 & 0.0 & 1 \\
\react{}    & 0.5  & 1 & 1.1 & 2 & 0.0 & 1 \\
\totdfs{}  & 16.7 & 9 & 2.8 & 8 & 0.5 & 7 \\
\totbfs{}  & 2.2  & 6 & 1.1 & 2 & 0.5 & 7 \\
\got{}      & 1.1  & 2 & 1.6 & 4 & 0.0 & 1 \\
\rap{}      & 2.0  & 4 & 3.3 & 7 & 0.7 & 9 \\
\foa{}      & 2.2  & 6 & 3.3 & 7 & 0.0 & 1 \\
\bottomrule
\end{tabular}%
}
\end{table*}

\begin{table*}[tbp]
\centering
\caption{\textbf{Joint Failure Rates: Models, Per-pair Mean $\pm 2\,$SD Threshold.} Per (model, benchmark) pair, quality fails if below its own mean $-\,2\,$SD; cost fails if above its own mean $+\,2\,$SD. Joint Fail = fraction of runs where both quality and cost fail simultaneously. Rank 1 = lowest failure rate (best); ties resolved alphabetically.}
\label{tab:appx_joint_models_2sd}
\scriptsize
\resizebox{0.92\textwidth}{!}{%
\begin{tabular}{lcccccc}
\toprule
& \multicolumn{2}{c}{\textbf{Quality}} & \multicolumn{2}{c}{\textbf{Cost}} & \multicolumn{2}{c}{\textbf{Joint}} \\
\cmidrule(lr){2-3} \cmidrule(lr){4-5} \cmidrule(lr){6-7}
Model & Fail (\%) & Rank & Fail (\%) & Rank & Fail (\%) & Rank \\
\midrule
DeepSeek R1       & 0.0 & 1  & 3.3 & 9  & 0.0 & 1 \\
Claude Haiku 4.5  & 0.0 & 1  & 1.7 & 3  & 0.0 & 1 \\
GPT-4.1 Nano      & 4.4 & 10 & 2.8 & 8  & 0.0 & 1 \\
Qwen3 235B        & 3.3 & 8  & 0.0 & 1  & 0.0 & 1 \\
Llama-4 Maverick  & 0.8 & 3  & 1.7 & 3  & 0.0 & 1 \\
GPT-4.1 Mini      & 2.2 & 6  & 2.2 & 5  & 0.0 & 1 \\
GPT-OSS 120B      & 2.8 & 7  & 2.2 & 5  & 0.0 & 1 \\
GPT-5 Nano        & 1.7 & 5  & 1.1 & 2  & 0.0 & 1 \\
GPT-5 Mini        & 1.1 & 4  & 3.9 & 10 & 0.0 & 1 \\
Gemini 3 Flash    & 3.5 & 9  & 2.4 & 7  & 0.5 & 10 \\
\bottomrule
\end{tabular}%
}
\end{table*}

\xhdr{Heavy-tailed cost and noise coupling}
Across the full $\$722.90$ of API spend, the top $10\%$ of runs account for $52.4\%$ of model-axis cost and $42.9\%$ of strategy-axis cost. Aggregate Score Run~Noise and Cost Run~Noise are uncorrelated across models (Spearman $\rho{=}{-}0.07$, $p{=}0.85$) but correlated across strategies ($\rho{=}{+}0.63$, $p{=}0.067$): on the model axis, score variance from token sampling and cost variance from generation length are independent stochastic sources, while on the strategy axis a search procedure that wanders stochastically through the same problem produces both a different answer and a different bill.

\end{document}